\documentclass[journal]{IEEEtran}
\usepackage{amssymb}
\usepackage{amsmath}
\usepackage{amsfonts}
\usepackage{cite}
\usepackage{graphics}
\usepackage{epsf}
\usepackage{float}
\usepackage[pdftex]{graphicx}
\usepackage[small]{caption2}
\usepackage{bm}
\usepackage{acronym}
\usepackage{multirow}
\usepackage{color}
\usepackage[normalem]{ulem}

\newcommand{\be}{\begin{equation}}
\newcommand{\ee}{\end{equation}}
\newcommand{\bea}{\begin{eqnarray}}
\newcommand{\eea}{\end{eqnarray}}
\newcommand{\beaa}{\begin{eqnarray*}}
\newcommand{\eeaa}{\end{eqnarray*}}

\acrodef{1D}[1D]{one-dimensional}
\acrodef{2D}[2D]{two-dimensional}
\acrodef{ML}[ML]{machine-learning}

\title{Generalized Tree-Based Wavelet Transform}

\begin{document}
\sloppy

\author{Idan Ram, Michael~Elad,~\IEEEmembership{Senior~Member,~IEEE}, and Israel~Cohen,~\IEEEmembership{Senior~Member,~IEEE}
\thanks{I. Ram and I. Cohen are with the Department of Electrical Engineering, Technion --
Israel Institute of Technology, Technion City, Haifa 32000, Israel. E-mail
addresses: idanram@tx.technion.ac.il (I. Ram), icohen@ee.technion.ac.il
(I. Cohen); tel.: +972-4-8294731; fax: +972-4-8295757.
M. Elad is with the Department of Computer Science, Technion --
Israel Institute of Technology, Technion City, Haifa 32000, Israel. E-mail
address: elad@cs.technion.ac.il

This research was partly supported by the ISF grant no. 1031/08 and by the European Community's FP7-FET program, SMALL project, under grant
agreement no. 225913.} }

\maketitle

\begin{abstract}

In this paper we propose a new wavelet transform applicable to functions defined on graphs, high dimensional data and networks.
The proposed method generalizes the Haar-like transform proposed in \cite{gavish2010mwot},
and it is defined via a hierarchical tree,
which is assumed to capture the geometry and structure of the input data.
It is applied to the data using a \textcolor{black}{modified version of the common 1D wavelet}
filtering and decimation scheme, which can employ different wavelet filters.
\textcolor{black}{In each level of this wavelet decomposition scheme, a permutation derived from the tree
is applied to the approximation coefficients, before they are filtered.}
We propose a tree construction method \textcolor{black}{that} results in \textcolor{black}{an} efficient representation of the input function in the transform domain.
We show that the proposed transform is more efficient than both the \ac{1D} and \ac{2D} separable wavelet transforms in representing images.
We also explore the application of the proposed transform to image denoising,
and show that combined with a subimage averaging scheme,
it achieves denoising results which are similar to \textcolor{black}{those} obtained with the K-SVD algorithm.

\end{abstract}


\section{Introduction}

Most traditional signal processing methods are designed for data defined on regular Euclidean grids.
Development of comparable methods capable of handling non-uniformly sampled signals,
data defined on graphs or ``point clouds'', is important and very much needed.
Many signal processing problems involve inference of an unknown
scalar target function defined on such data.
For example, function denoising involves estimating such a scalar function from its noisy version.
A different example is image inpainting which involves estimating missing pixels of a column \textcolor{black}{stacked} version of an image from its known pixels.
A major challenge in processing functions on topologically complicated data,
is to find efficient methods to represent and learn them.

Many signal processing techniques are based on transform methods, which represent the input data in a new basis, before analyzing or processing it.
One of the most successful types of transforms, which has been proven to be a very useful tool for signal and image processing, is wavelet analysis \cite{mallat2009wavelet}.
A major advantage of wavelet methods is their ability to simultaneously localize signal content in both space
and frequency.
This property allows them to compactly represent signals such as \ac{1D} steps or images with edges,
whose primary information content lies in localized singularities.
Moreover, wavelet methods represent such signals much more compactly than either the original domain
or transforms with global basis elements such as the Fourier transform.
\textcolor{black}{We aim at} extending the wavelet transform to irregular, non-Euclidean spaces,
\textcolor{black}{and thus obtain a} \textcolor{black}{transform} that
efficiently represent functions defined on such data.

Several extensions of the wavelet transform, operating on graphs and high dimensional data, have
already been proposed.
The wavelet transforms proposed in \textcolor{black}{\cite{murtagh2007haar},\cite{lee2008treelets} and \cite{chen2010multiscale} were applied to the data points themselves, rather than functions defined on the data.
Other methods took different approaches to construct wavelets applied to functions on the data.
Maggioni and Coifman \cite{coifman2006diffWave} and Hammond et al. \cite{hammond2010wavelets} proposed wavelets based on diffusion operators and the graph Laplacian \cite{chung1997spectral},\cite{von2007tutorial}, respectively.
Jansen et al. \cite{jansen2009multiscale} proposed three methods which were based on a variation of the lifting scheme \cite{sweldens1996lifting},\cite{sweldens1998lifting}.
Gavish et al. \cite{gavish2010mwot} assumed that the geometry and structure of the input data are captured in a hierarchical tree.
Then, given such a tree, they built a data adaptive Haar-like orthonormal
basis for the space of functions over the data set.
This basis, can be seen as a generalization of the one proposed in \cite{egiazarian2002tree} for binary trees.
Our proposed method generalizes the algorithm in \cite{gavish2010mwot}, and can also construct data adaptive orthonormal non-Haar wavelets,
given a data driven-hierarchical tree.}

\textcolor{black}{We note that the wavelet transforms proposed in \cite{mallat2009geometrical}, and \cite{plonka2009easy}, which are defined on images,
also share some similarities with our proposed algorithm.
These methods employ pairing or reordering of wavelet coefficients in the decomposition schemes in order to adapt to their input images.
In fact, the easy path wavelet transform proposed in \cite{plonka2009easy}, which only recently has come to our attention, employs a decomposition scheme which is very similar to ours. Constraining our algorithm to a regular data grid and using the same starting point and search neighborhood as the ones employed in \cite{plonka2009easy}, both algorithms essentially coincide.
Nevertheless, our approach is more general, as it tackles the more abstract problem of devising a transform for point clouds or high-dimensional graph-data, whereas \cite{plonka2009easy} concentrates on images.}

In this paper we introduce a generalized tree-based wavelet transform (GTBWT),
which is an extension of the transform introduced by Gavish et al \cite{gavish2010mwot}.
We first show \textcolor{black}{that} the transform of Gavish et al., when derived from a full binary tree,
\textcolor{black}{can be applied to a function over the data set using a modified version of} the common \ac{1D} Haar wavelet filtering and decimation scheme.
\textcolor{black}{In each level of this wavelet decomposition scheme,
a permutation derived from the tree is applied to the approximation coefficients, before they are filtered}.
Then we show how this scheme can be extended to work with different wavelet filters,
by modifying the way different levels of the tree are constructed.

\textcolor{black}{The construction of each coarse level of the tree involves finding a path which passes through the set of data points in the finer level.
The points order in this path defines the permutation applied to the approximation coefficients of the finer level in the wavelet decomposition scheme.
We propose a path constructed by starting from a random point, and then continue from each point to its nearest
neighbor according to some distance measure, visiting each point only once.
The corresponding permutation increases the regularity of the permuted approximation coefficients signal,
and therefore it is more efficiently (sparsely) represented using the wavelet transform.}

Next we show that the proposed scheme is more efficient
than both the common \ac{1D} and \ac{2D} separable wavelet transforms in representing images.
Finally, we explore the application of the proposed transform to image denoising,
and show that combined with a proposed subimage averaging scheme,
it achieves denoising results similar to the ones obtained with the K-SVD algorithm \cite{elad2006image}.

The paper is organized as follows:
In Section II we describe how the Haar-like basis introduced in \cite{gavish2010mwot} is \textcolor{black}{derived from} a full binary tree representation of the data.
We also describe how such a tree may be constructed.
In Section III we introduce the generalized tree-based wavelet transform.
We also explore the efficiency with which this transform represents an image.
In Section IV we explore the application of our proposed algorithm to image denoising.
We also describe the subimage averaging scheme and present some experimental results.
We summarize the paper in Section V.

\section{Tree-based Haar wavelets}

Let $\mathbf{X}=\left\{\mathbf{x}_1,...,\mathbf{x}_N\right\}$ be the dataset we wish to analyze,
where the samples $\mathbf{x}_i\textcolor{black}{\in \mathbb{R}^n}$ may be points in high dimension, or nodes in a weighted graph or network.
Also, let $f:\mathbf{X}\rightarrow \mathbb{R}$ be a scalar function defined on the dataset,
and let $V=\left\{f \left|f:\mathbf{X} \rightarrow \mathbb{R}  \right. \right\}$ be the space of all functions on the dataset.
Here we use the following inner product with the space $V$:
\begin{equation}
\langle f,g \rangle = \sum_{j=1}^{N}f(\mathbf{x}_j)g(\mathbf{x}_j).
\end{equation}
which is different from the one used by Gavish et al. \cite{gavish2010mwot}, since it does not contain a normalizing factor before the sum.

Gavish et al. assume that the geometry and structure of the data $\mathbf{X}$ are
captured by one or several hierarchical trees.
They do not insist on any specific construction method for these trees,
but only that they will be balanced \cite{gavish2010mwot}.
Given such a tree, they construct a multiscale wavelet-like orthonormal basis for the space $V$.
They start by showing that such a tree induces a multi-resolution analysis with an associated Haar-like wavelet.

Let $\ell=1,\ldots,L$ denote the level in the tree, with $\ell=1$ being the root and $\ell=L$ being the lowest level,
where each sample $\mathbf{x}_j$ is a single leaf.
Also,  let $V^\ell$ denote the space of functions constant on all folders (subtrees) at level $\ell$,
and let $\mathbf{1}_\mathbf{X}$ denote a constant function on $\mathbf{X}$ with the value 1.
Then $V^1=Span_{\mathbb{R}}\{\mathbf{1}_\mathbf{X}\}$, $V^L=V$ and by construction
\begin{equation}
V^1 \subset ... \subset V^\ell \subset V^{\ell+1} \subset ... \subset V^L=V.
\end{equation}
Now, let $W^\ell$ $(1\leq\ell<L)$ be the orthogonal complement of $V^\ell$ in $V^{\ell+1}$.
Then, the space of all functions $V$ can be decomposed as
\begin{equation}
V=V^L=\left[\bigoplus_{\ell=1}^{L-1}W^\ell \right]\bigoplus V^1.
\end{equation}

Before describing how the multiscale orthonormal basis is constructed given a hierarchical tree,
we first describe how such a tree can be constructed from the data.
Here we focus on the case of \textcolor{black}{complete} full binary trees and the corresponding orthonormal bases.
For the case of more general trees and Haar like bases, the reader may refer to \cite{gavish2010mwot}.
Let $\mathbf{c}_j^\ell$ denote the $j$-th point at level $\ell$ \textcolor{black}{of the tree},
where $\mathbf{c}_j^L=\mathbf{x}_j$,
and let $\mathcal{P}_\ell$ and $\mathbf{p}_\ell$ denote a set and a vector containing point \textcolor{black}{indices}, respectively.
Also, let \textcolor{black}{$w(\cdot,\cdot)$ be a distance measure in $\mathbb{R}^n$},
and let $\textcolor{black}{\mathbf{W}^\ell}$ be
\textcolor{black}{a distance} matrix associated with the $\ell$-th level of the tree,
where $\textcolor{black}{w^\ell_{i,j}=w(\mathbf{c}_i^\ell,\mathbf{c}_j^\ell)}$.
The \textcolor{black}{distance} function describes the first-order interaction between data-points,
and therefore it should be chosen so as to capture some notion of similarity between them,
which would be meaningful to the application at hand.
A \textcolor{black}{complete} full binary tree can be constructed from the data according to Algorithm 1.
\textcolor{black}{An example for such a tree is shown in Fig. \ref{Figure: binary tree}}.

\newcounter{alg}
\renewcommand{\thefigure}{\arabic{alg}}
\setcounter{alg}{1}
\begin{figure}[t]
     \begin{center}
        \begin{tabular}{|c|} \hline
        \begin{minipage}[h]{3.0in}
        \renewcommand{\baselinestretch}{1}
        \small \vspace{0.1in}
        \begin{list}{}{}
        \item \textsf {{\bf Task:} Construct a \textcolor{black}{complete} full binary tree from the data $\mathbf{X}$.}

        \item \textsf {{\bf Parameters:} We are given the points $\{\mathbf{x}_j\}_{j=1}^N$
        and the \textcolor{black}{distance} function $w$.}

        \item \textsf {{\bf Initialization:} Set $\mathbf{c}_j^L=\mathbf{x}_j$ as the tree leaves}.

        \item\textsf {{\bf Main Iteration:} Perform the following steps for $\ell=L,\ldots,2$:}
        \begin{itemize}

        \item \textsf {Construct \textcolor{black}{a distance} matrix $\textcolor{black}{\mathbf{W}^{\ell}}$,
        where $\textcolor{black}{w^\ell_{i,j}=w(\mathbf{c}_i^\ell,\mathbf{c}_j^\ell)}$.}

        \item \textsf {Set $\mathcal{P}_\ell=\emptyset$.}

        \item \textsf {Group the points in level $\ell$ in pairs by repeating $N/2^{1+L-\ell}$ times:}

        \begin{itemize}

            \item \textsf {Choose a random point $\mathbf{c}_{j_0}^\ell$, $j_0 \notin \mathcal{P}_\ell$,
            and update $\mathcal{P}_\ell=\mathcal{P}_\ell\cup\{j_0\}$.}

            \item \textsf{Pair $\mathbf{c}_{j_0}^\ell$ with the point $\mathbf{c}_{j_1}^\ell$,
            where $\textcolor{black}{j_1=\min_{j\notin \mathcal{P}_\ell} w^\ell_{j_0,j}}$.}

            \item \textsf{Update $\mathcal{P}_\ell=\mathcal{P}_\ell\cup\{j_1\}$.}

        \end{itemize}

        \item \textsf{Place in a vector $\mathbf{p}_\ell$ the reordered point \textcolor{black}{indices} of $\mathcal{P}_\ell$.}

        \item \textsf {Construct the coarse level $\ell - 1$ from the finer level $\ell$ by replacing each pair $\mathbf{c}_i^\ell$ and $\mathbf{c}_j^\ell$ with the \textcolor{black}{mean} point $\frac{1}{\sqrt{2}}[\mathbf{c}_i^\ell + \mathbf{c}_j^\ell]$.}

        \end{itemize}

        \item \textsf{{\bf Output:} The tree node points $\mathbf{c}_{j}^\ell$
        and the vectors $\mathbf{p}_\ell$ containing the points order in each tree level.} \vspace{0.05in}
        \end{list}
        \end{minipage}
        \\\hline
        \end{tabular}
        \\ \vspace{0.1in}
        \renewcommand{\figurename}{Algorithm}
        \caption{\textcolor{black}{Complete} full binary tree construction from the data $\mathbf{X}$.
        \label{Haar_tree_construction}}
     \end{center}
     \linespread{1.4}
\end{figure}
\addtocounter{figure}{-1}
\addtocounter{alg}{1}
\renewcommand{\thefigure}{\arabic{figure}}

\begin{figure}[t]
\centering
\includegraphics[scale=0.38]{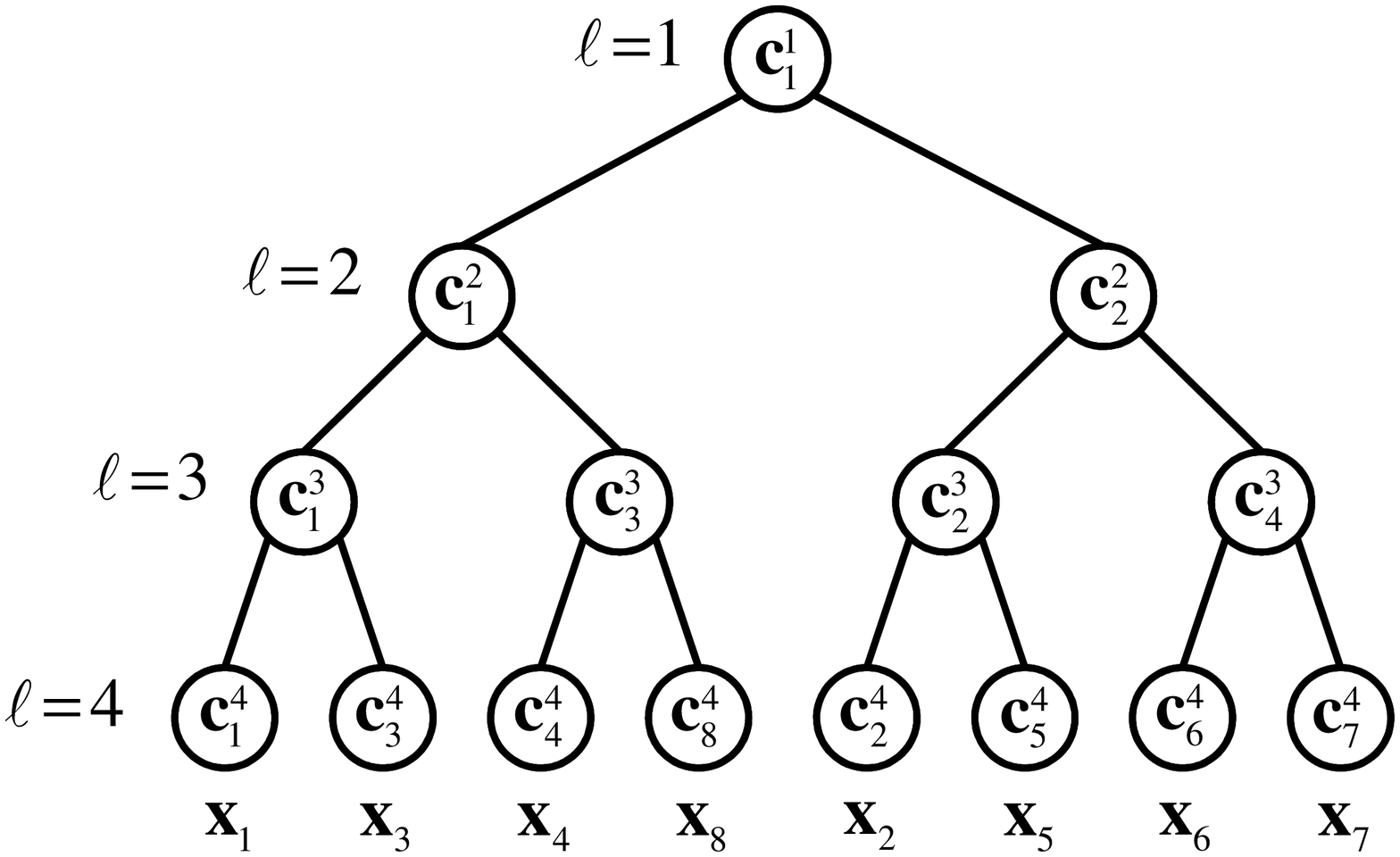}
\caption{An illustration of a complete full binary tree.}
\label{Figure: binary tree}
\end{figure}

\noindent

\textcolor{black}{In the case of a full binary tree, the Haar-like basis constructed from the tree is essentially the standard Haar basis which we denote $\{\psi_j\}_{j=1}^N$.
The adaptivity of the transform is manifested in the fact that this basis is used to represent \emph{a permuted} version of the signal $f$,
which is more efficiently represented by the Haar basis than $f$ itself.
The permutation is derived from the tree, and is dependent on the data $\mathbf{X}$.
Let $\mathbf{p}$ denote a vector of length $N$, which contains the indices of the points $\mathbf{x}_j$,
in the order determined by the lowest level of the fully constructed tree.
For example, for the tree in Fig. \ref{Figure: binary tree}, $\mathbf{p}=[1,3,4,8,2,5,6,7]^T$.
Also let $f^p$ be the signal $f$ permuted according to the vector $\mathbf{p}$.
The wavelet coefficients can be calculated by the inner products $\left<f^p,\psi_j\right>$,
or by applying the 1D wavelet filtering and decimation scheme with the Haar wavelet filters on $f^p$.
Similarly, the inverse transform is calculated by applying the inverse Haar transform on the wavelet coefficients,
and reordering the produced vector so as to cancel the index ordering in $\mathbf{p}$.}
We hereafter term the scheme described above as tree-based Haar wavelet transform,
and we show next that it can be extended to operate with general wavelet filters.

\section{Generalized tree-based wavelets}
\label{gen_tree_const}
\subsection{Generalized tree construction and transform}

The above-described building process of the tree can be presented a little differently.
\textcolor{black}{In every level $\ell$ of the tree,} we first construct
\textcolor{black}{a distance matrix $\mathbf{W}^{\ell}$} using the \textcolor{black}{mean} points
$\mathbf{c}_j^\ell$ and the \textcolor{black}{distance} function $w$.
Then we group in pairs the points $\mathbf{c}_j^\ell$,
according to the weights in $\textcolor{black}{\mathbf{W}^\ell}$,
as described in Algorithm 1.
Next we place the pairs \textcolor{black}{of column vectors} one after the other in a matrix $\mathbf{C}_\ell^p$ of size \textcolor{black}{$n \times N/2^{L-\ell}$},
and keep the \textcolor{black}{indices} of the points in their new order in a vector $\mathbf{p}_\ell$ \textcolor{black}{of length $N/2^{L-\ell}$}.
\textcolor{black}{For example in level $\ell=3$ of the tree in Fig. \ref{Figure: binary tree},
$\mathbf{C}_3^p=[\mathbf{c}_1^3,\mathbf{c}_3^3,\mathbf{c}_2^3,\mathbf{c}_4^3]$ and $\mathbf{p}_3=[1,3,2,4]^T$}.

Now let $\bar{\mathbf{h}}=\frac{1}{\sqrt{2}}\left[1,1\right]^T$ and $\bar{\mathbf{g}}=\frac{1}{\sqrt{2}}\left[{-1},1\right]^T$ be the Haar wavelet decomposition filters,
and let $\mathbf{h}=\frac{1}{\sqrt{2}}\left[1,1\right]^T$ and $\mathbf{g}=\frac{1}{\sqrt{2}}\left[1,{-1}\right]^T$ be the Haar wavelet reconstruction filters.
We notice that replacing each pair by its \textcolor{black}{mean} point can be done by filtering \textcolor{black}{the rows}
of $\mathbf{C}_\ell^p$ with the low pass filter $\bar{\mathbf{h}}^T$,
followed by decimation of the columns of the outcome by a factor of $2$.
\textcolor{black}{For example, the points $\mathbf{c}_2^1$ and $\mathbf{c}_2^2$ in level $\ell=2$ of the tree in Fig. \ref{Figure: binary tree}
are obtained by filtering the rows of $\mathbf{C}_3^p$ described above with the filter $\bar{\mathbf{h}}^T$,
and keeping the first and third columns of the produced matrix.}
Effectively this means that the approximation coefficients corresponding to a single-level Haar decomposition are calculated for each row of the matrix $\mathbf{C}_\ell^p$.

\textcolor{black}{Next} let $\mathbf{f}=[f(\mathbf{x}_1),\ldots,f(\mathbf{x}_N)]^T$, and let $\mathbf{a}_\ell$ and $\mathbf{d}_\ell$ denote the approximation and detail coefficient vectors, respectively,
received for $\mathbf{f}$ at level $\ell$, where $\mathbf{a}_L=\mathbf{f}$.
Also, let $P_\ell$ denote a linear operator that reorders a vector according to the \textcolor{black}{indices} in $\mathbf{p}_\ell$.
Then, applying the Haar transform derived from the tree to $\mathbf{f}$ can be carried out according to the decomposition algorithm in Algorithm 2.
Fig. \ref{Figure: wavelet decomposition} describes two single-level decomposition steps carried out according to Algorithm 2.
\textcolor{black}{We note that here the adaptivity of the transform is related to the permutations applied to the
coefficients $\mathbf{a}_\ell$ in every level of the tree. In fact,}
without the operator $P_\ell$ applied in each level,
the decomposition scheme of Algorithm 2 reduces to that of the common \ac{1D} orthogonal  wavelet transform.
Also, since permutation of points is a unitary transform, the described transform remains unitary.

\renewcommand{\thefigure}{\arabic{alg}}
\begin{figure}[t]
     \begin{center}
        \begin{tabular}{|c|} \hline
        \begin{minipage}[h]{3.0in}
        \renewcommand{\baselinestretch}{1}
        \small \vspace{0.1in}
        \begin{list}{}{}
        \item \textsf {{\bf Task:} Apply $L-1$ \textcolor{black}{levels of} tree-based wavelet decomposition to the signal $\mathbf{f}$.}

        \item \textsf {{\bf Parameters:} We are given the signal $\mathbf{f}$, the vectors $\{\mathbf{p}_\ell\}_{\ell=2}^L$,
        and the filters $\bar{\mathbf{h}}$ and $\bar{\mathbf{g}}$ .}

        \item \textsf {{\bf Initialization:} Set $\mathbf{a}_L=\mathbf{f}$}.

        \item\textsf {{\bf Main Iteration:} Perform the following steps for $\ell=L,\ldots,2$:}
        \begin{itemize}

        \item \textsf {Construct the operator $P_\ell$ that reorders its input vector according to the \textcolor{black}{indices} in $\mathbf{p}_\ell$.}

        \item \textsf {Apply $P_\ell$ to $\mathbf{a}_\ell$ and receive $\mathbf{a}_\ell^p$.}

        \item \textsf {Filter $\mathbf{a}_\ell^p$ with $\bar{\mathbf{h}}$ and decimate the result by $2$ to receive $\mathbf{a}_{\ell-1}$.}

        \item \textsf {Filter $\mathbf{a}_\ell^p$ with $\bar{\mathbf{g}}$ and decimate the result by $2$ to receive $\mathbf{d}_{\ell-1}$.}

        \end{itemize}

        \item \textsf{{\bf Output:} The approximation coefficients $\mathbf{a}_1$ and detail coefficients $\{\mathbf{d}_\ell\}_{\ell=1}^{L-1}$
        corresponding to $\mathbf{f}$.} \vspace{0.05in}
        \end{list}
        \end{minipage}
        \\\hline
        \end{tabular}
        \\ \vspace{0.1in}
        \renewcommand{\figurename}{Algorithm}
        \caption{Tree-based wavelet decomposition algorithm.
        \label{wavedec_algorithm}}
     \end{center}
     \linespread{1.4}
\end{figure}
\addtocounter{figure}{-1}
\addtocounter{alg}{1}
\renewcommand{\thefigure}{\arabic{figure}}

\begin{figure}[t]
\centering
\includegraphics[scale=0.7]{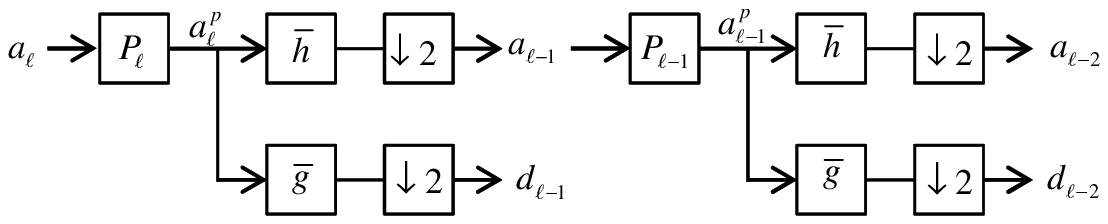}
\caption{Tree-based wavelet decomposition scheme}
\label{Figure: wavelet decomposition}
\end{figure}

Finally, let the linear operator $P_\ell^{-1}$ reorder a vector so as to cancel the ordering done by $P_\ell$.
Then the inverse transform is carried out using the reconstruction algorithm in Algorithm 3.
Fig. \ref{Figure: wavelet reconstruction} describes two single-level reconstruction steps carried out according to Algorithm 3.

\renewcommand{\thefigure}{\arabic{alg}}
\begin{figure}[t]
     \begin{center}
        \begin{tabular}{|c|} \hline
        \begin{minipage}[h]{3.0in}
        \renewcommand{\baselinestretch}{1}
        \small \vspace{0.1in}
        \begin{list}{}{}
        \item \textsf {{\bf Task:} Reconstruct the signal $\mathbf{f}$ based on a multilevel tree-based wavelet decomposition.}

        \item \textsf {{\bf Parameters:} We are given the approximation and detail coefficients
        $\mathbf{a}_1$ and $\{\mathbf{d}_\ell\}_{\ell=1}^{L-1}$,
        the vectors $\{\mathbf{p}_\ell\}_{\ell=2}^L$,
        and the filters $\mathbf{h}$ and $\mathbf{g}$ .}

        \item\textsf {{\bf Main Iteration:} Perform the following steps for $\ell=1,\ldots,L-1$:}
        \begin{itemize}

        \item \textsf {Interpolate $\mathbf{a}_\ell$ by a factor of $2$ and filter the result with $\mathbf{h}$.}

        \item \textsf {Interpolate $\mathbf{d}_\ell$ by a factor of $2$ and filter the result with $\mathbf{g}$.}

        \item \textsf {Sum the results of the two previous steps to receive $\mathbf{a}_{\ell+1}^p$.}

        \item \textsf {Construct the operator $P_{\ell+1}^{-1}$ that reorders its input vector so as to cancel the index ordering in $\mathbf{p}_{\ell+1}$.}

        \item \textsf {Apply $P_{\ell+1}^{-1}$ to $\mathbf{a}_{\ell+1}^p$ and receive $\mathbf{a}_{\ell+1}$.}

        \end{itemize}

        \item \textsf{{\bf Output:} The reconstructed signal $\mathbf{f}$.} \vspace{0.05in}
        \end{list}
        \end{minipage}
        \\\hline
        \end{tabular}
        \\ \vspace{0.1in}
        \renewcommand{\figurename}{Algorithm}
        \caption{Tree-based wavelet reconstruction algorithm.
        \label{waverec_algorithm}}
     \end{center}
     \linespread{1.4}
\end{figure}
\addtocounter{figure}{-1}
\addtocounter{alg}{1}
\renewcommand{\thefigure}{\arabic{figure}}

\begin{figure}[t]
\centering
\includegraphics[scale=0.7]{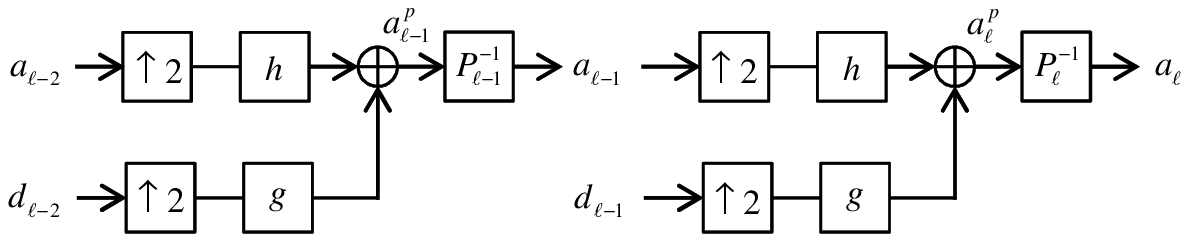}
\caption{Tree-based wavelet reconstruction scheme}
\label{Figure: wavelet reconstruction}
\end{figure}

\textcolor{black}{We next wish to extend the scheme described above to work with general wavelet filters.
This requires modifying the tree construction by replacing the Haar filters by different wavelet filters
and changing the manner in which the points $\mathbf{c}_j^\ell$ are ordered in each level of the tree.
The ordering procedure used in Algorithm 1 fits only the case when Haar wavelets are used,
and therefore needs to be replaced.}

\begin{figure}[t]
\centering
\includegraphics[scale=0.38]{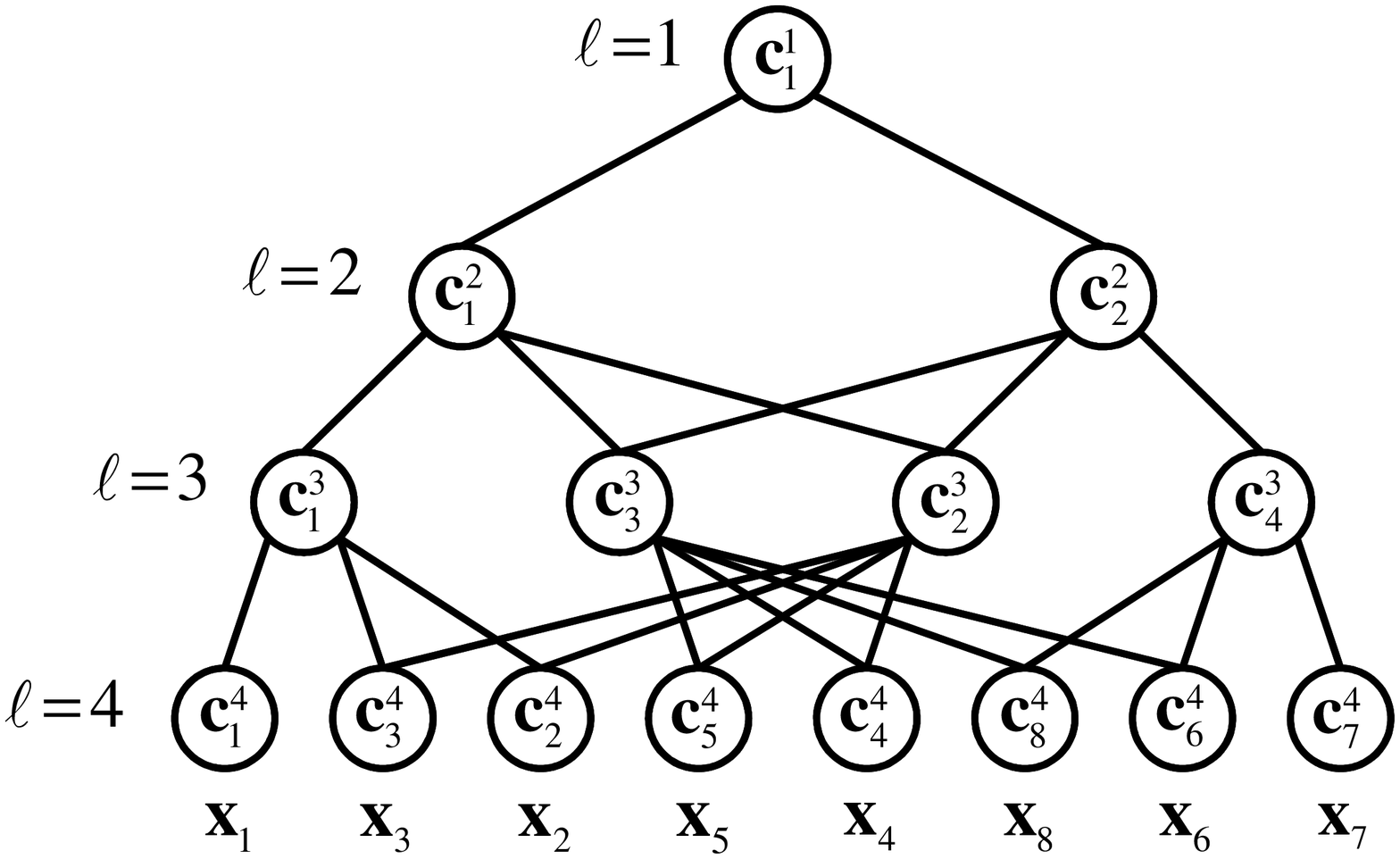}
\caption{An illustration of a "generalized" tree.}
\label{Figure: multipartiteGraph}
\end{figure}

\textcolor{black}{We note that when filters other than Haar are used in the tree construction scheme described above,
each point in the coarse level $\ell - 1$ is calculated as a weighted mean of more than two points from the finer level $\ell$,
where the coefficients in $\bar{\mathbf{h}}$ serve as the weights.
Therefore, the resultant graph is no longer a tree but rather a rooted $L$-partite graph,
which is a graph that contains $L$ disjoint sets of vertices so that no two vertices within the same set are adjacent.
As the wavelet scheme described above was originally designed with the Haar wavelet filters,
and in order to avoid cumbersome distinction between trees and $L$-partite graphs,
with a small abuse of terminology we will hereafter refer to the latter also as trees.
An example of such a "generalized" tree, is shown in Fig. \ref{Figure: multipartiteGraph}.}

\textcolor{black}{The wavelet decomposition and reconstruction schemes corresponding to each generalized tree are those described in Algorithms 2 and 3,
with the necessary change of wavelet filters type and index vectors $\mathbf{p_\ell}$ to the ones used in the construction of the graph.
We next propose a method to order the points in each level of the generalized trees.}

\renewcommand{\thefigure}{\arabic{alg}}
\begin{figure}[t]
     \begin{center}
        \begin{tabular}{|c|} \hline
        \begin{minipage}[h]{3.0in}
        \renewcommand{\baselinestretch}{1}
        \small \vspace{0.1in}
        \begin{list}{}{}
        \item \textsf {{\bf Task:} Construct of a "generalized" tree from the data $\mathbf{X}$.}

        \item \textsf {{\bf Parameters:} We are given the points $\{\mathbf{x}_j\}_{j=1}^N$
        and the weight function $w$ .}

        \item \textsf {{\bf Initialization:} Set $\mathbf{c}_j^L=\mathbf{x}_j$} as the tree leaves.

        \item\textsf {{\bf Main Iteration:} Perform the following steps for $\ell=L,\ldots,2$:}

        \begin{itemize}

        \item \textsf {Construct \textcolor{black}{a distance} matrix $\textcolor{black}{\mathbf{W}^{\ell}}$,
        where $\textcolor{black}{w^\ell_{i,j}=w(\mathbf{c}_i^\ell,\mathbf{c}_j^\ell)}$.}

        \item \textsf {Choose a random point $\mathbf{c}_{j_0}^\ell$ and set $\mathcal{P}_\ell=\{j_0\}$.}

        \item \textsf {Reorder the points $\mathbf{c}_j^\ell$ so that they will form a smooth path by repeating $N/2^{L-\ell}-1$ times:}

        \begin{itemize}

            \item \textsf{Set $\textcolor{black}{j_1=\min_{j\notin \mathcal{P}_\ell} w^\ell_{j_0,j}}$ and update $\mathcal{P}_\ell=\mathcal{P}_\ell\cup\{j_1\}$.}

            \item \textsf {Set $j_0=j_1$.}

        \end{itemize}

        \item \textsf{Place in a vector $\mathbf{p}_\ell$ the reordered point \textcolor{black}{indices} of $\mathcal{P}_\ell$.}

        \item \textsf {Order the points $\mathbf{c}_j^\ell$ according to the \textcolor{black}{indices} in $\mathbf{p}_\ell$ and place them
        in a matrix $\mathbf{C}_\ell^p$.}

        \item \textsf {Obtain the points $\mathbf{c}_j^{\ell-1}$ by:}

        \begin{itemize}

            \item \textsf {Apply the filter $\bar{\mathbf{h}}^T$ to the matrix $\mathbf{C}_\ell^p$.}

            \item \textsf {Decimate the columns of the outcome by a factor of 2.}

        \end{itemize}

        \end{itemize}

        \item \textsf{{\bf Output:} The tree node points $\mathbf{c}_{j}^\ell$
        and the vectors $\mathbf{p}_\ell$ containing the points order in each tree level.} \vspace{0.05in}
        \end{list}
        \end{minipage}
        \\\hline
        \end{tabular}
        \\ \vspace{0.1in}
        \renewcommand{\figurename}{Algorithm}
        \caption{Construction of a "generalized" tree from the data $\mathbf{X}$.
        \label{Generalized_tree_construction}}
     \end{center}
     \linespread{1.4}
\end{figure}
\addtocounter{figure}{-1}
\addtocounter{alg}{1}
\renewcommand{\thefigure}{\arabic{figure}}

\subsection{Smoothing $\mathbf{a}_\ell$}

\textcolor{black}{We wish to order the points in each a level of a tree in a manner which results in an efficient representation of the input signal by the tree-based wavelets.}
More specifically, we want the transformed signal to contain a small number of large coefficients, i.e. to be sparse.
The wavelet transform is known to produce a small number of large coefficients when it is applied to piecewise regular signals \cite{mallat2009wavelet}.
Thus, we would like the operator $P_\ell$, applied to $\mathbf{a}_\ell$, to produce a signal which is as regular as possible.
When the signal $\mathbf{f}$ is known,
the optimal solution would be to apply a simple {\em sort} operation on the corresponding coefficients $\mathbf{a}_\ell$,
obtained in each level.
However, since we are interested in the case where $\mathbf{f}$ is not necessarily known (such as in the case where $\mathbf{f}$ is noisy, or has missing values),
we would try to find a suboptimal ordering operation in each level $\ell$, using the feature points $\mathbf{c}_j^\ell$.
We assume that under the distance measure $w(\cdot,\cdot)$,
proximity between the two points $\mathbf{c}_i^\ell$ and $\mathbf{c}_j^\ell$ suggests proximity between the coefficients $a_\ell(i)$ and $a_\ell(j)$.
Thus, we would try to reorder the points $\mathbf{c}_j^\ell$ so that they will form a smooth path,
hoping that the corresponding reordered \ac{1D} signal $\mathbf{a}_\ell^p$ will also be smooth.

The ``smoothness'' of a \ac{1D} signal $\mathbf{y}$ \textcolor{black}{of length $N$} can be measured using its total variation
\begin{equation}
\textcolor{black}{\|\mathbf{y}\|_V=\sum_{j=2}^{N} |y(j)-y(j-1)|}.
\end{equation}
By analogy, we measure the "smoothness" of the path $\mathbf{C}_\ell^p$ by the measure
\begin{equation}
C_{\ell,V}^p=\sum_{j=2}^{N} w(\mathbf{c}_j^\ell,\mathbf{c}_{j-1}^\ell).
\end{equation}
We notice that smoothing the path $\mathbf{C}_\ell^p$ comes down to finding the shortest path that passes through the set of points $\mathbf{c}_j^\ell$, visiting each point only once.
This can be regarded as an instance of the traveling salesman problem \cite{cormen2001introduction},
which can become very computationally exhaustive for large sets of points.
A simple approximate solution is to start from a random point, and then continue from each point to its nearest neighbor, not visiting any point twice.
As it turns out, ordering the points $\mathbf{c}_j^\ell$ in this manner improves the performance of the resultant transform.

\textcolor{black}{A} generalized tree construction,
\textcolor{black}{which employs the proposed point ordering method} is summarized in Algorithm 4.
\textcolor{black}{Fig. \ref{Figure: multipartiteGraph} shows an example of "generalized" tree which may be obtained with Algorithm 4
using a filter $\bar{\mathbf{h}}$ of length 4 and disregarding boundary issues in the different levels.}
We term the filtering schemes described in Algorithms 2 and 3
combined with the tree construction \textcolor{black}{described} in Algorithm 4
generalized tree-based wavelet transform (GTBWT).

\begin{figure*}[t]
\centering
\begin{tabular}{cc}
\includegraphics[scale=0.4]{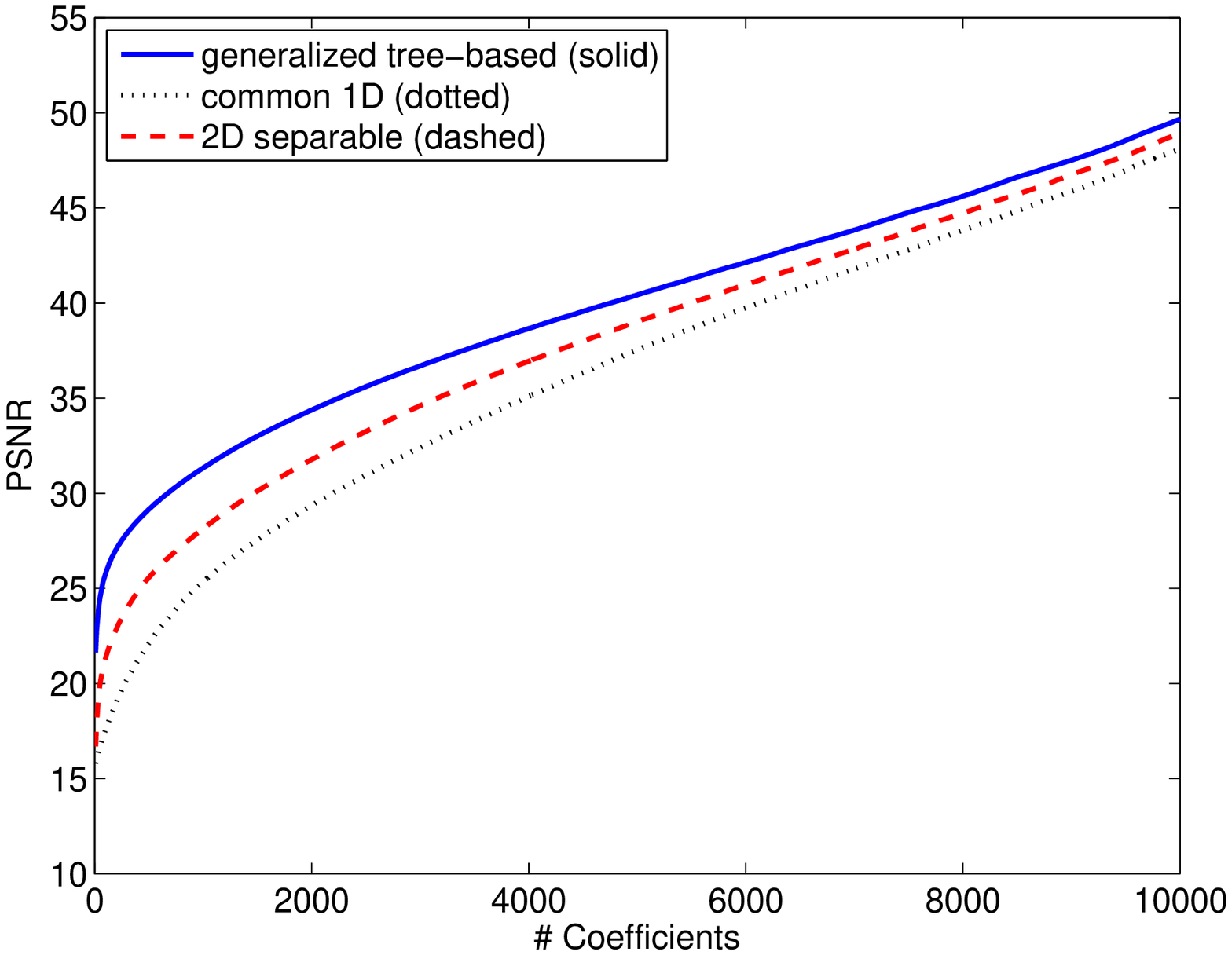} & \includegraphics[scale=0.4]{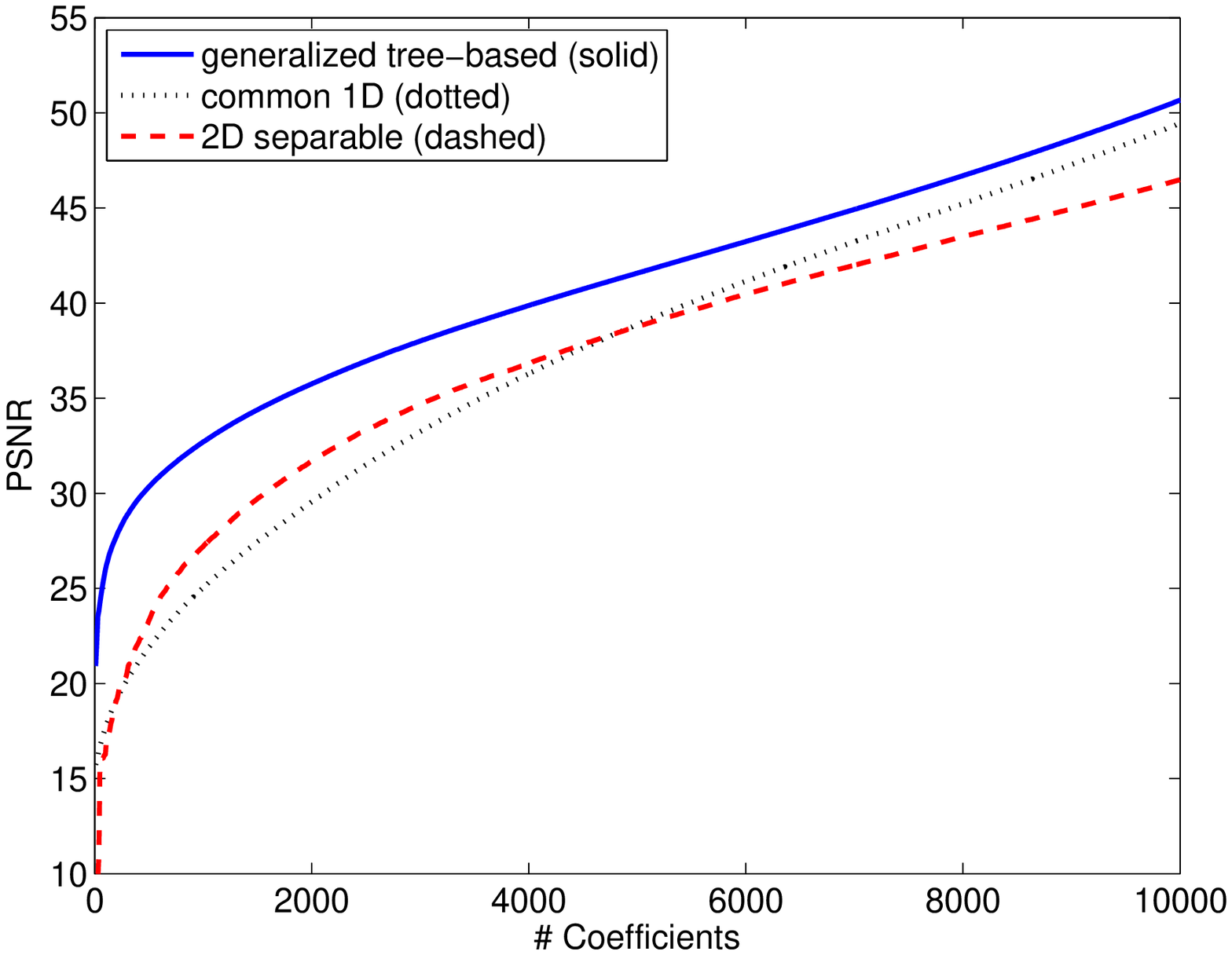}\\
{\small (a)} & {\small (b)}\\
\includegraphics[scale=0.4]{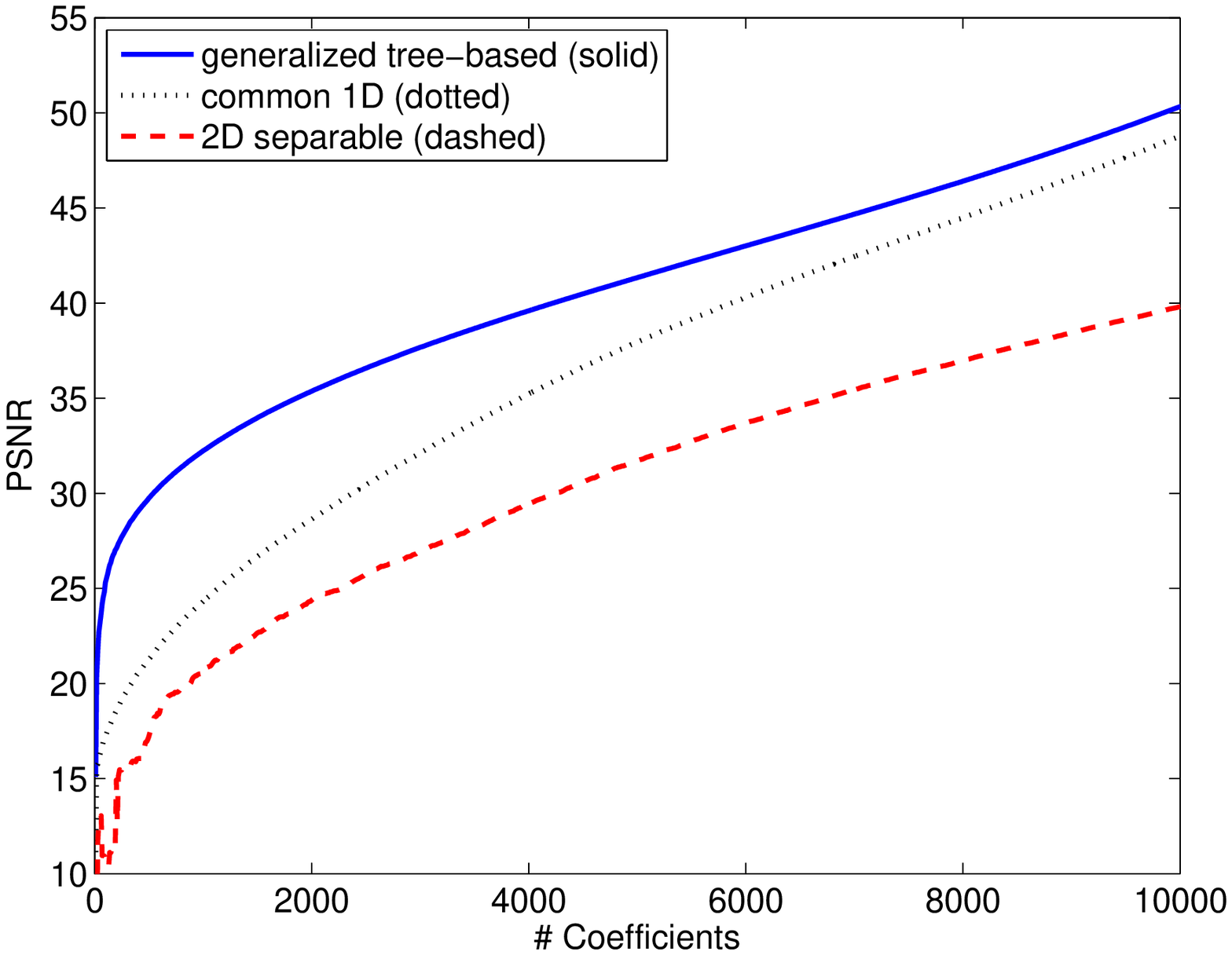} & \includegraphics[scale=0.4]{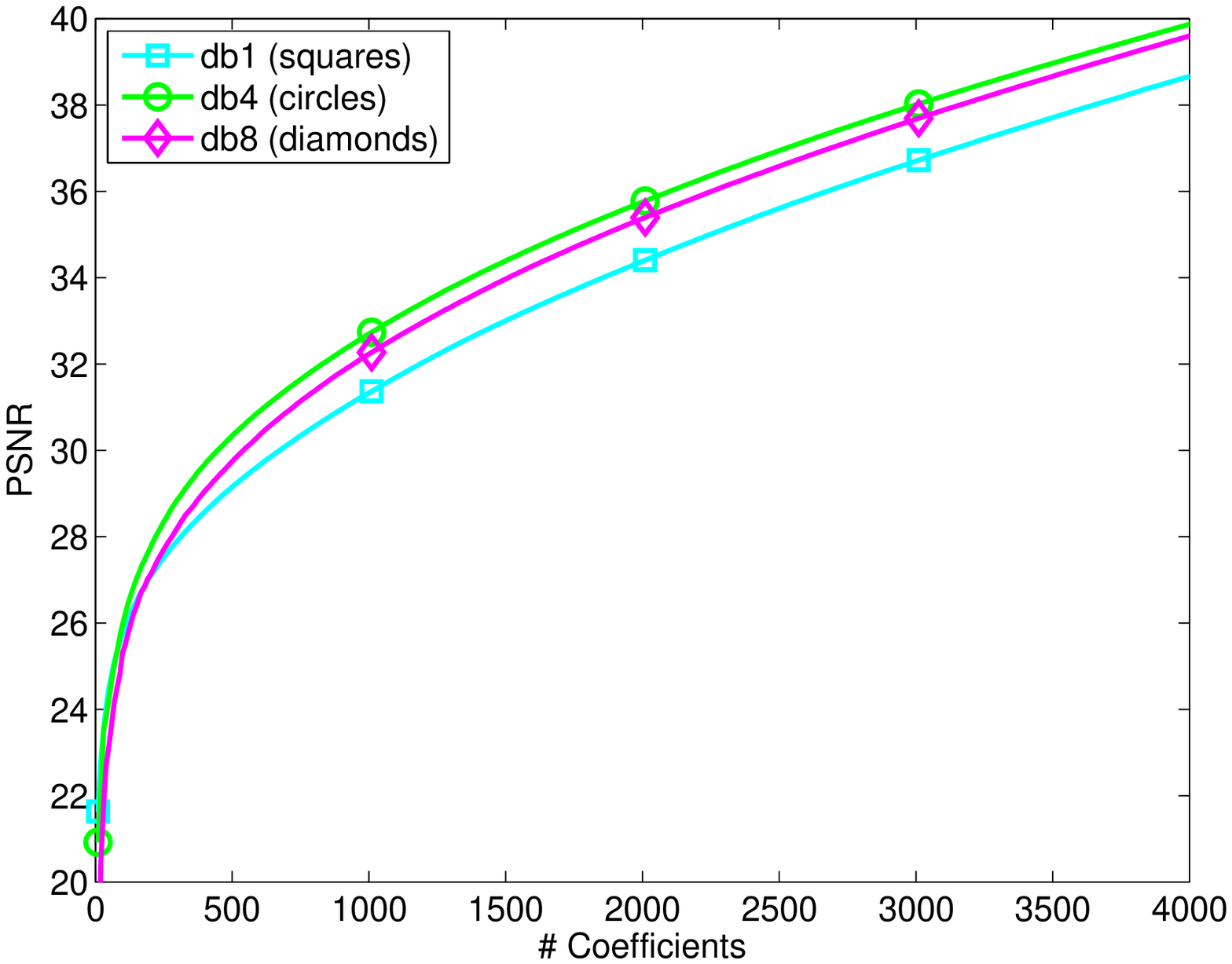}\\
{\small (c)} & {\small (d)} \\
\end{tabular}
\caption{$m$-term approximation results (PSNR in dB) for the generalized tree-based, common \ac{1D} and \ac{2D} separable wavelet transforms, obtained with different wavelet filters: (a) \textcolor{black}{Daubechies} 1 (Haar). (b) \textcolor{black}{Daubechies} 4. (c) \textcolor{black}{Daubechies} 8.
(d) Comparison between the generalized tree-based wavelet results obtained with the different filters. }
\label{Figure: smooth m term}
\end{figure*}

\textcolor{black}{An interesting question is whether the wavelet scheme described above represents the signal $\mathbf{f}$ more efficiently than the common \ac{1D} and \ac{2D} separable wavelet transforms.
Here, we measure efficiency by the $m$-term approximation error, i.e. the error obtained when representing a signal with $m$ non-zero transform coefficients.}

\textcolor{black}{Before relating to this question, we first describe how the wavelet scheme described above can be applied to images.
Let $\mathbf{F}$ be a grayscale image of size $\sqrt{N} \times \sqrt{N}$ and let $\mathbf{f}$ be its column stacked representation,
i.e. $\mathbf{f}$ is a vector of length $N$ containing individual pixel intensities.
Then we first need to extract the feature points $\mathbf{x}_i$ from the image, which will be later used to construct the tree.
Let $f_i$ be the $i$-th sample in $\mathbf{f}$, then we choose the point $\mathbf{x}_i$ associated with it as the $9 \times 9$ patch
around the location of $f_i$ in the image $\mathbf{F}$.
We next construct several trees, each with a different wavelet filter, according to the scheme described above.
We choose the weight function $w$ to be the squared Euclidean distance, i.e. the $(i,j)$ element in
$\mathbf{W}^\ell$ is}
\be
\textcolor{black}{w^\ell_{i,j}=\|\mathbf{c}_i^\ell-\mathbf{c}_j^\ell\|^2}.
\ee

\begin{figure*}[t]
\centering
\begin{tabular}{cccccc}
\multirow{2}{*}{\includegraphics[scale=0.28]{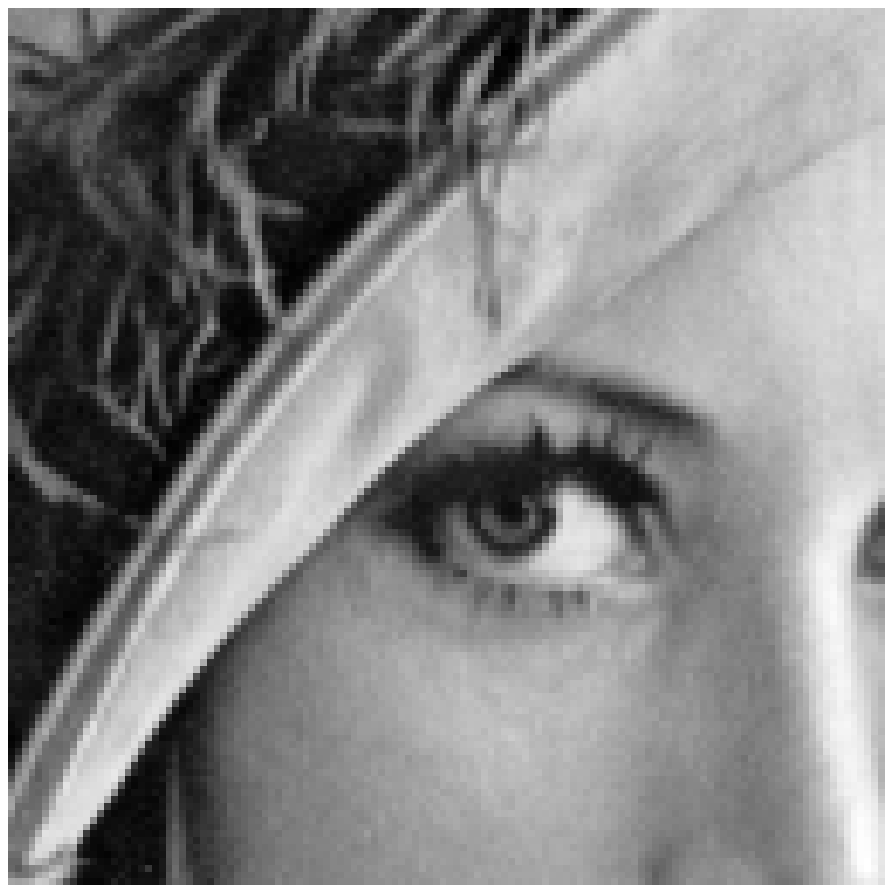}} & \includegraphics[scale=0.28]{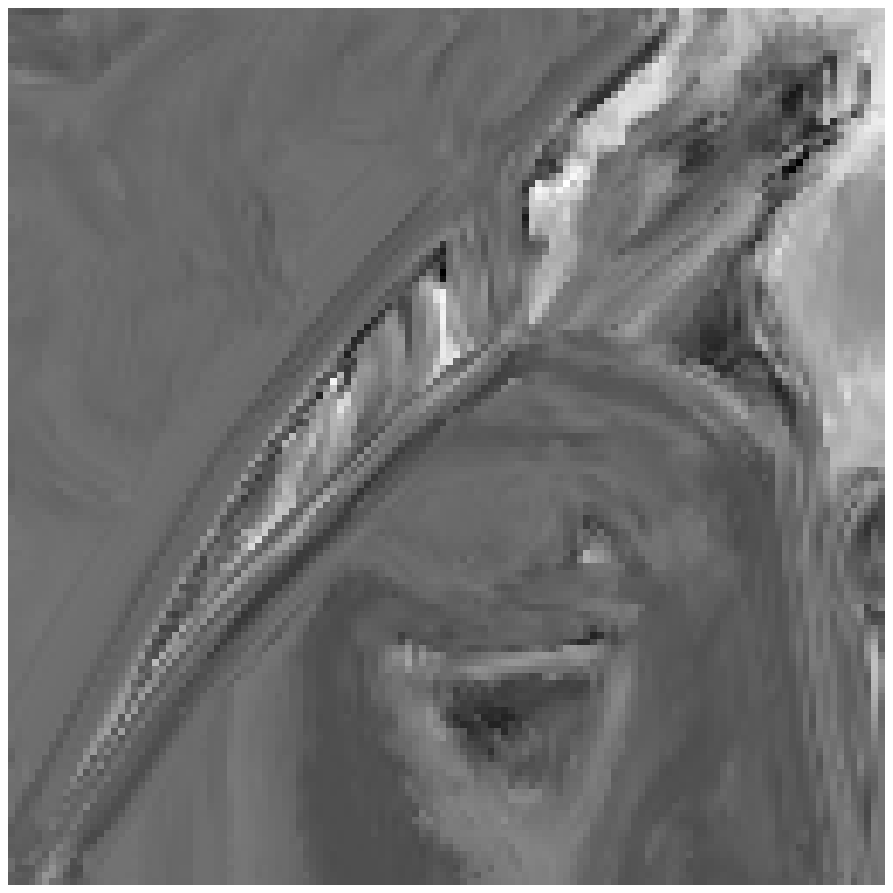} &
\includegraphics[scale=0.28]{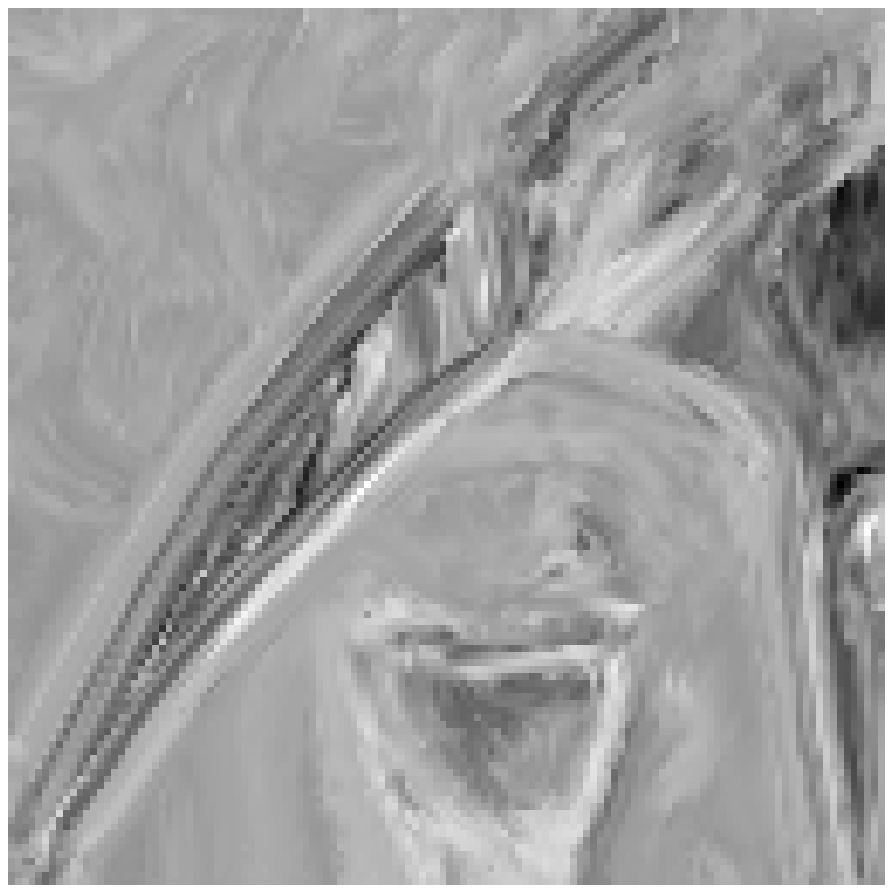} & \includegraphics[scale=0.28]{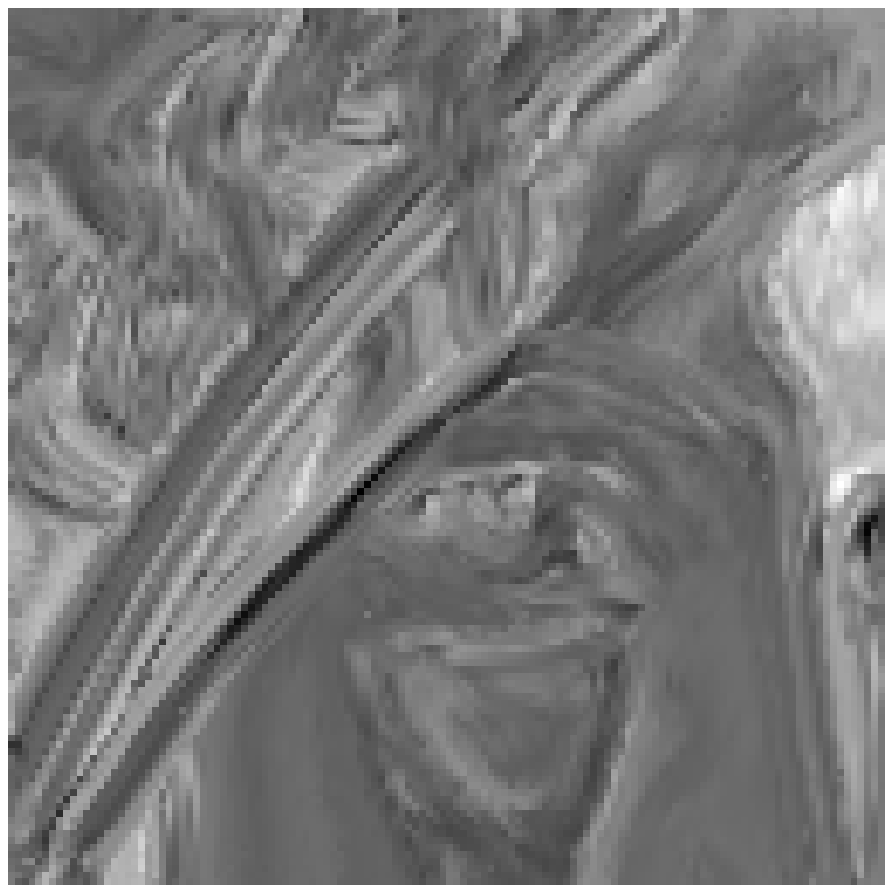} &
\includegraphics[scale=0.28]{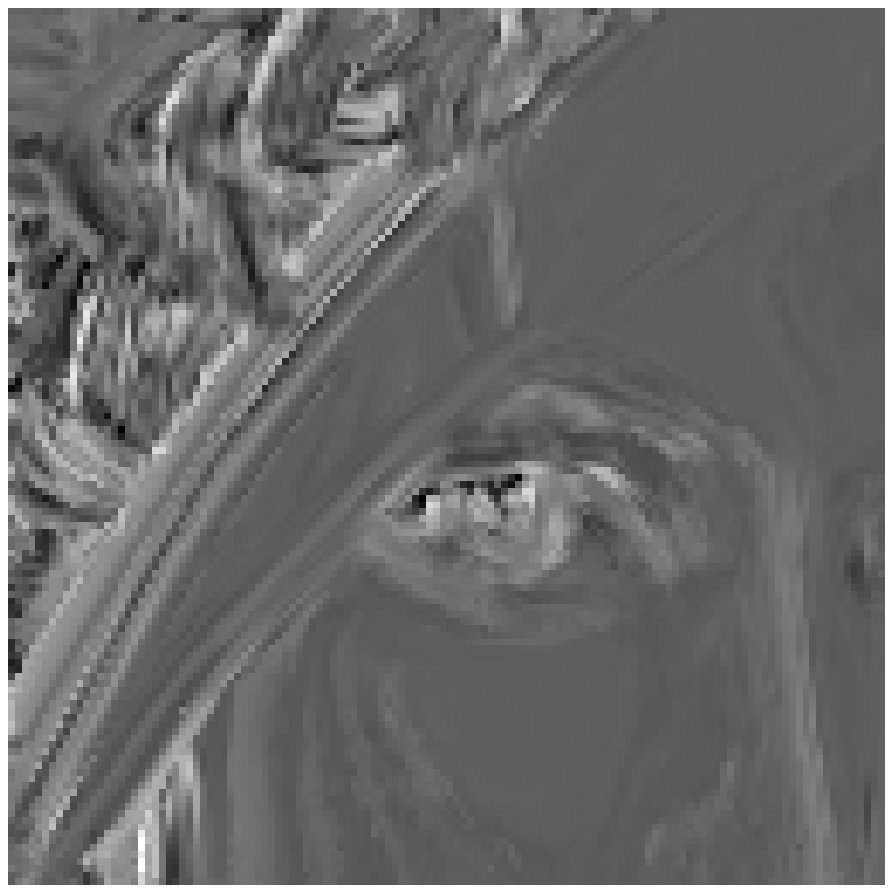} & \includegraphics[scale=0.28]{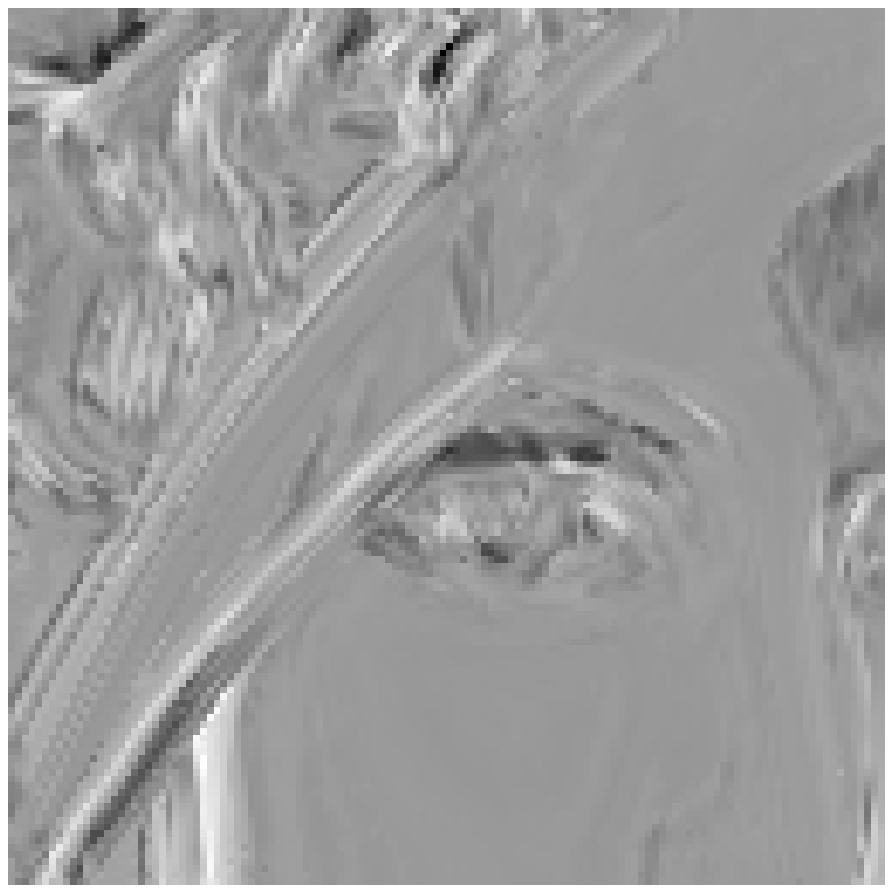}\\
& \includegraphics[scale=0.28]{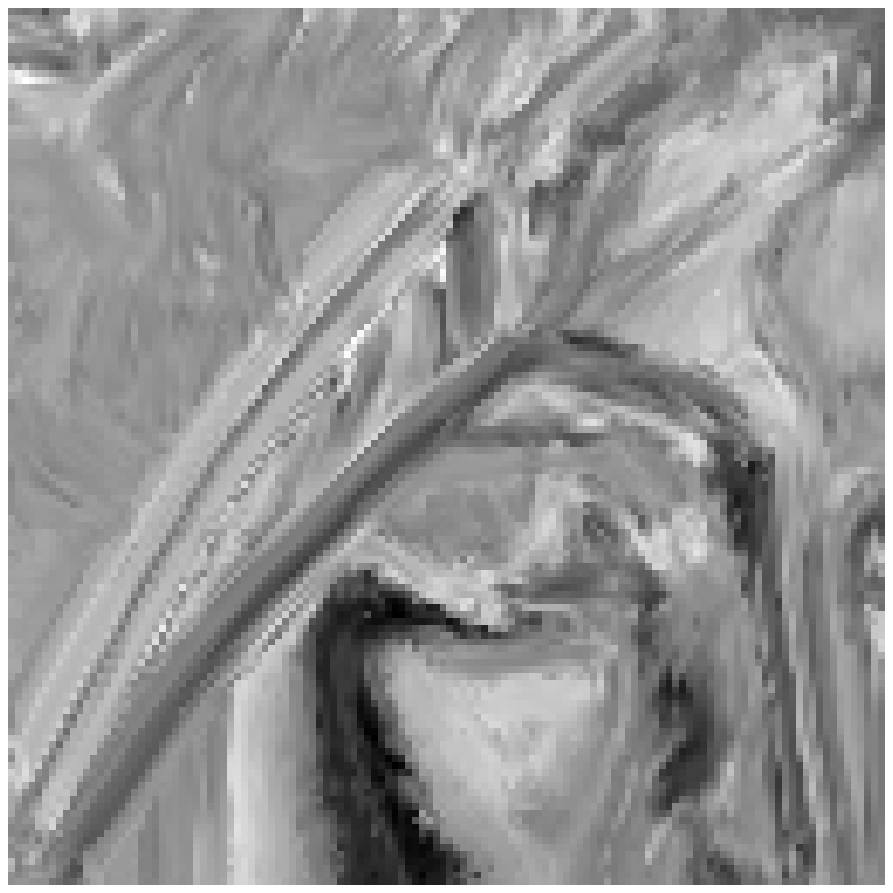} &
\includegraphics[scale=0.28]{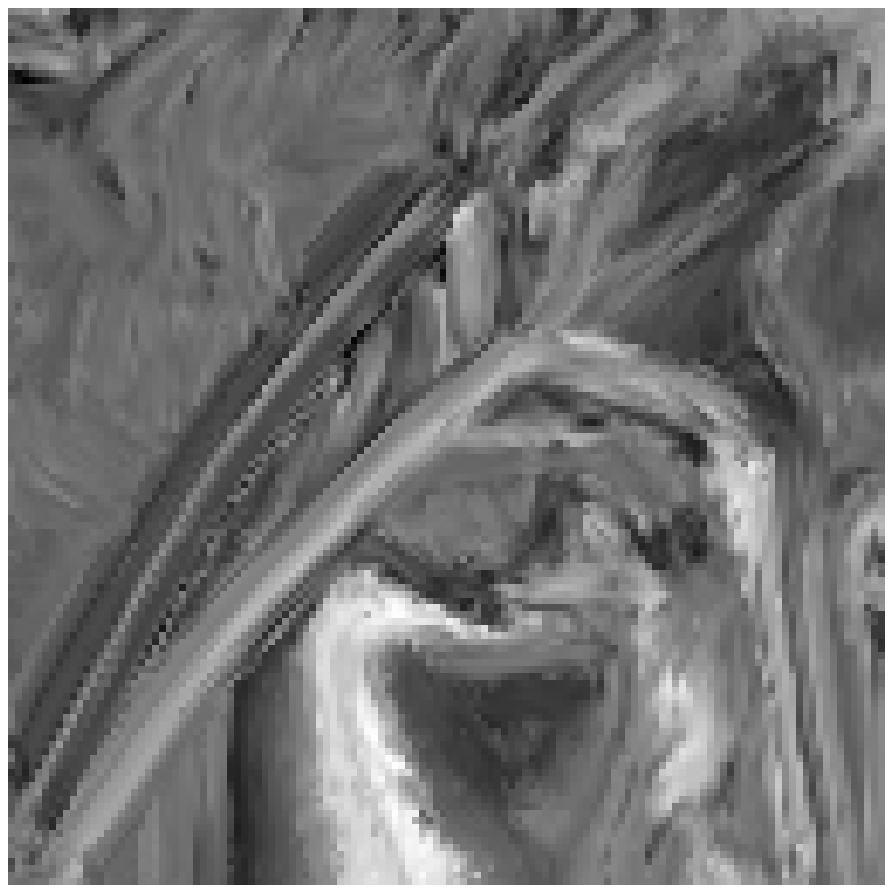} & \includegraphics[scale=0.28]{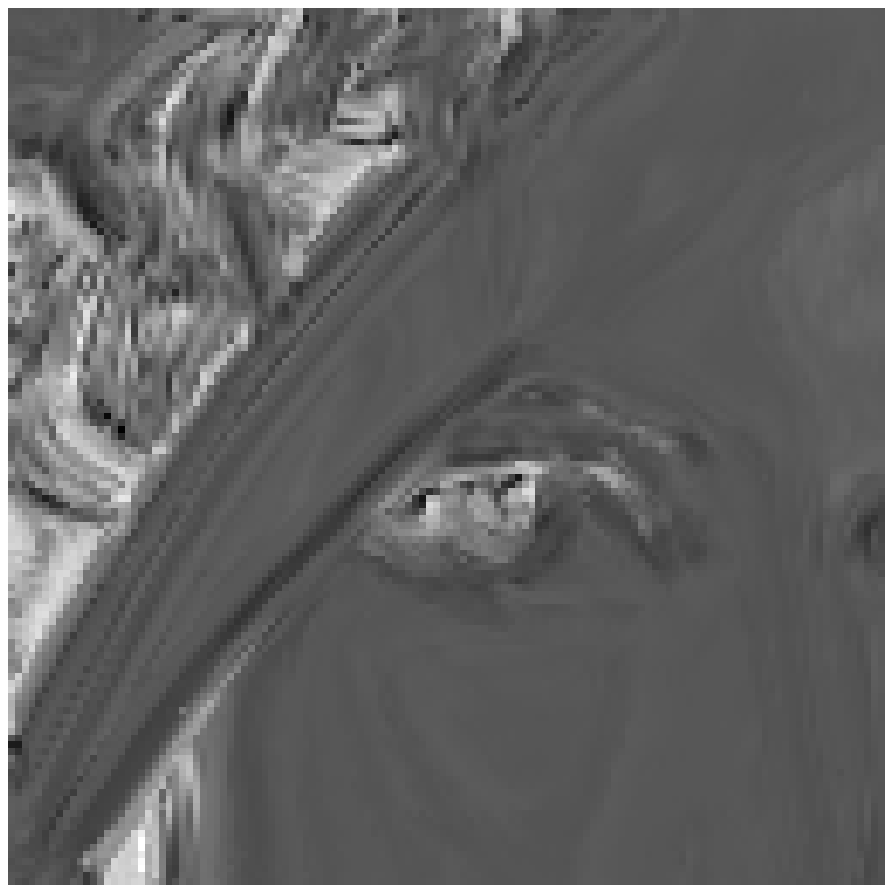} &
\includegraphics[scale=0.28]{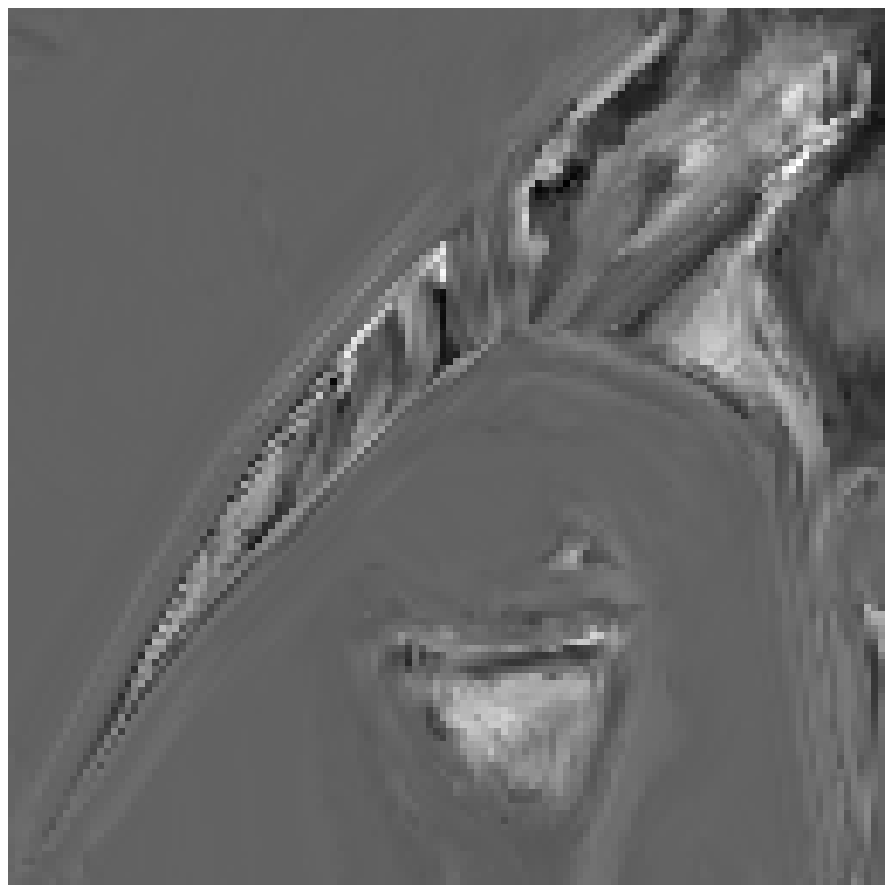} & \includegraphics[scale=0.28]{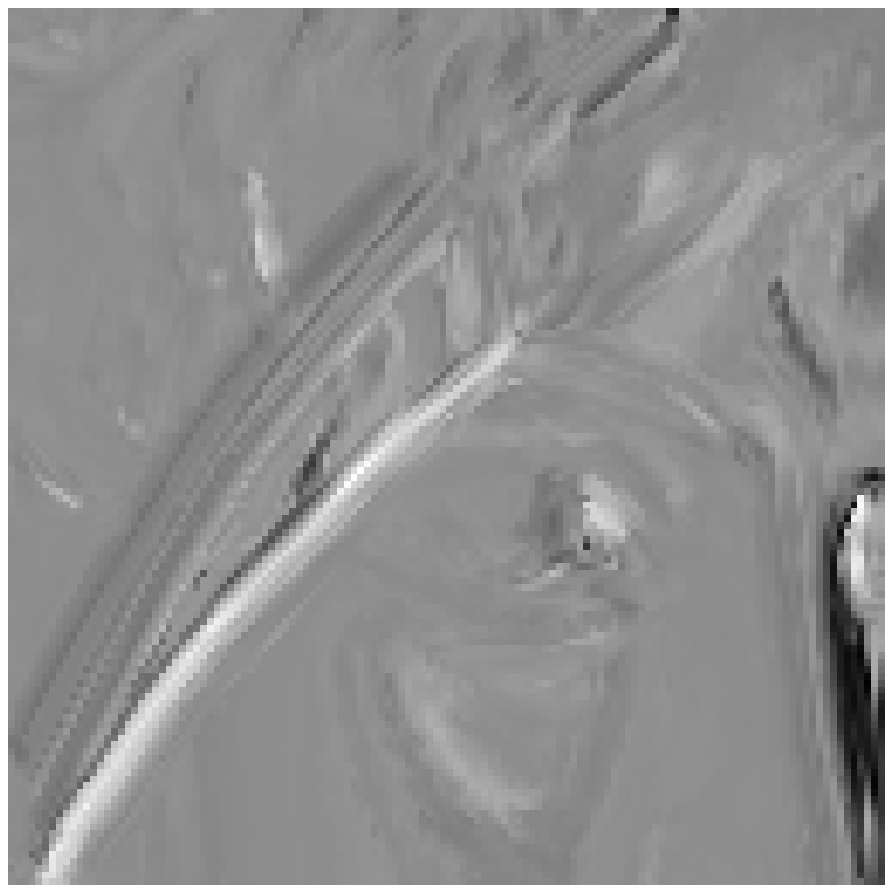}\\
{\small (a)} & {\small (b)} & {\small (c)} & {\small (d)} & {\small (e)} & {\small (f)}\\
\includegraphics[scale=0.28]{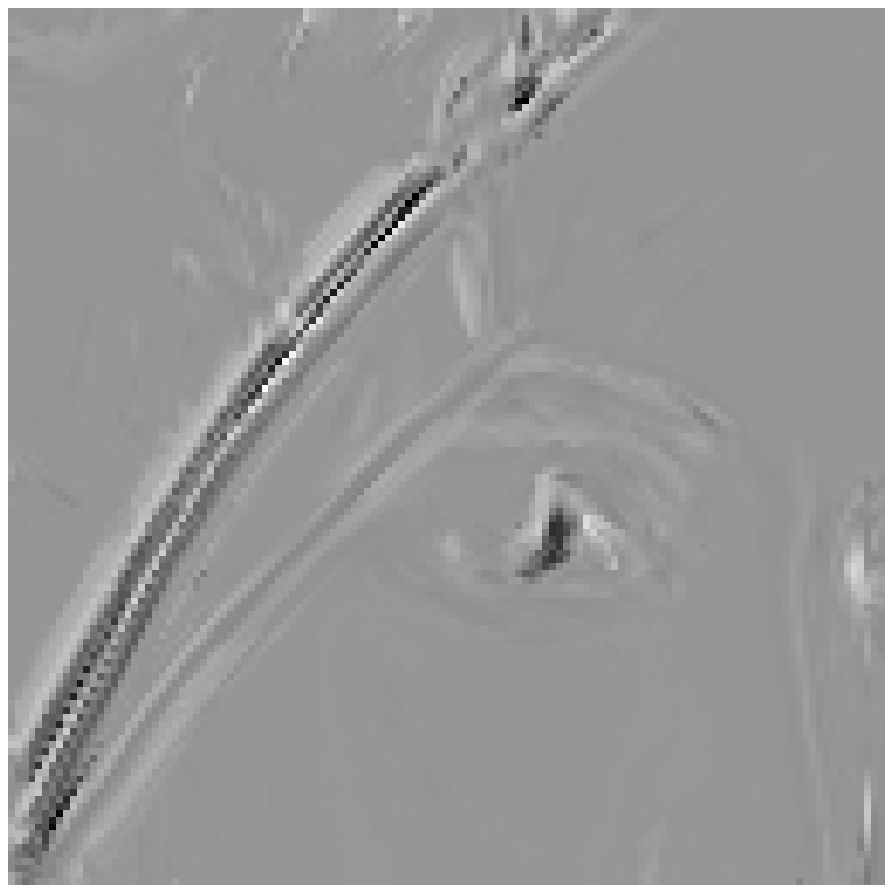} & \includegraphics[scale=0.28]{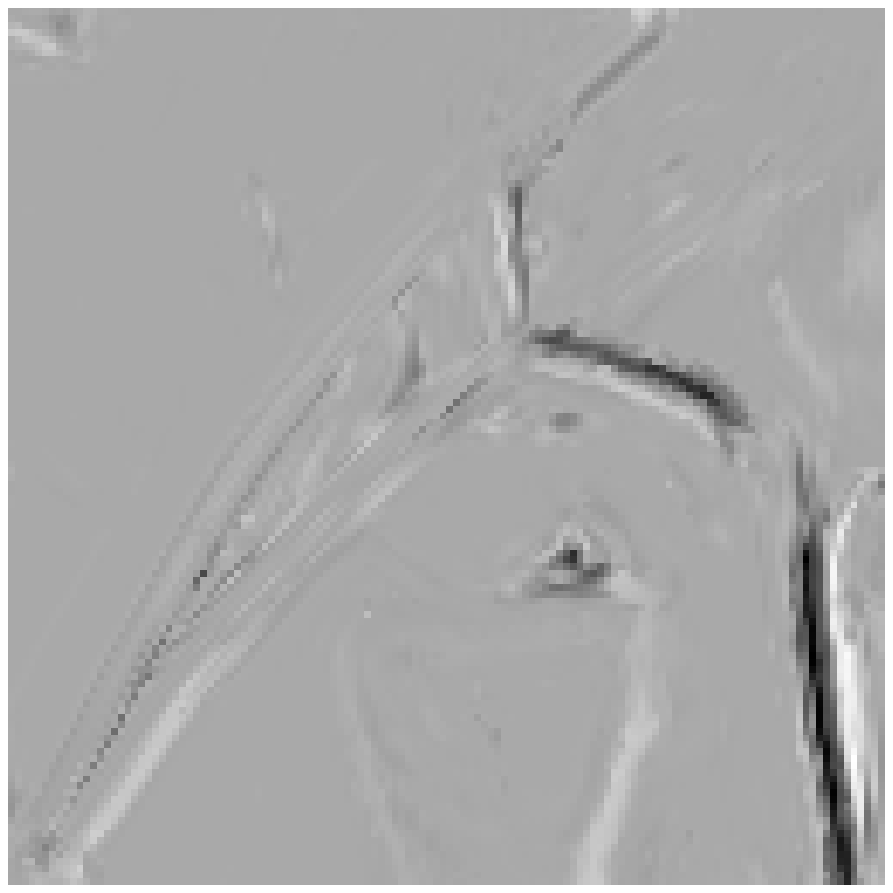} &
\includegraphics[scale=0.28]{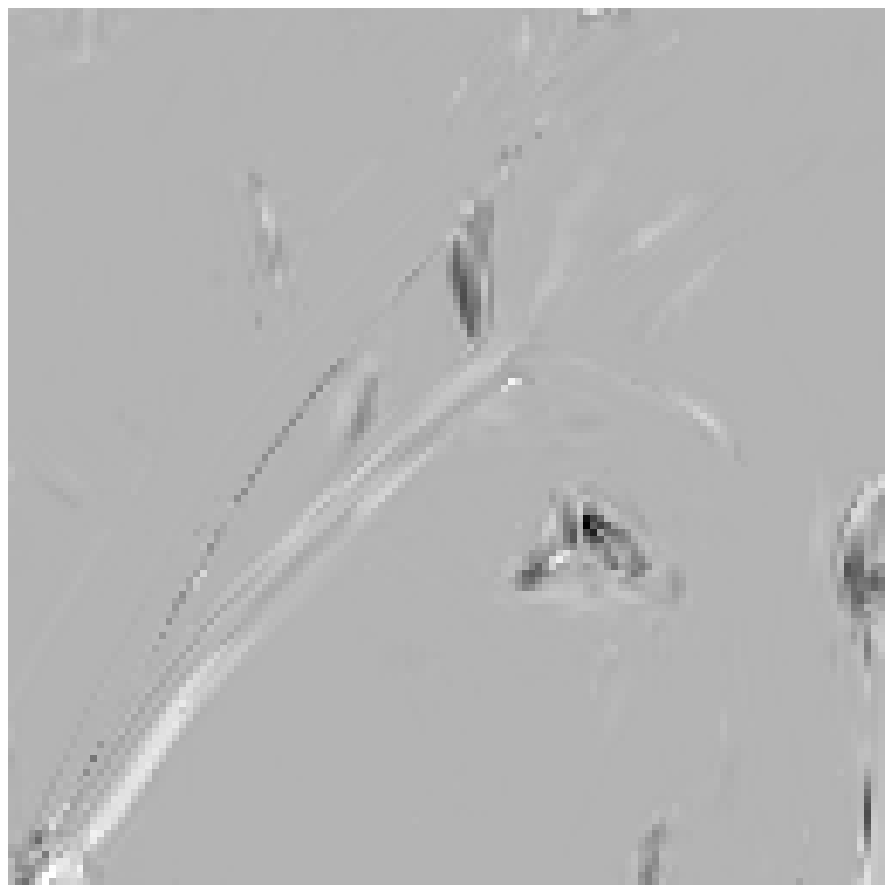} & \includegraphics[scale=0.28]{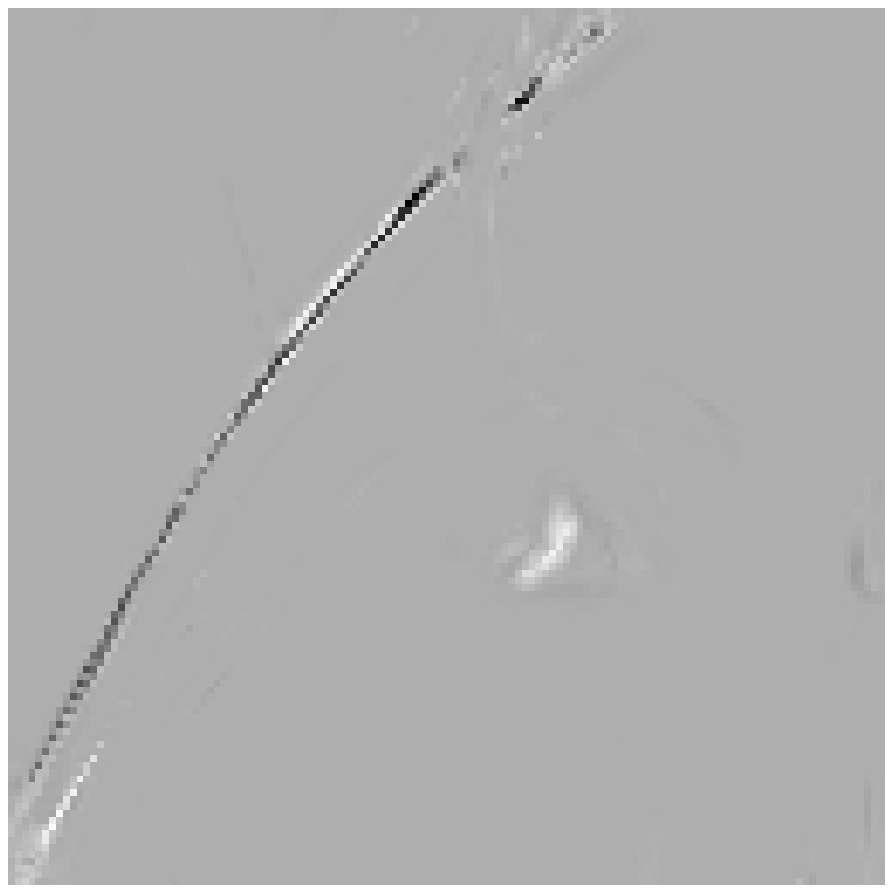} &
\includegraphics[scale=0.28]{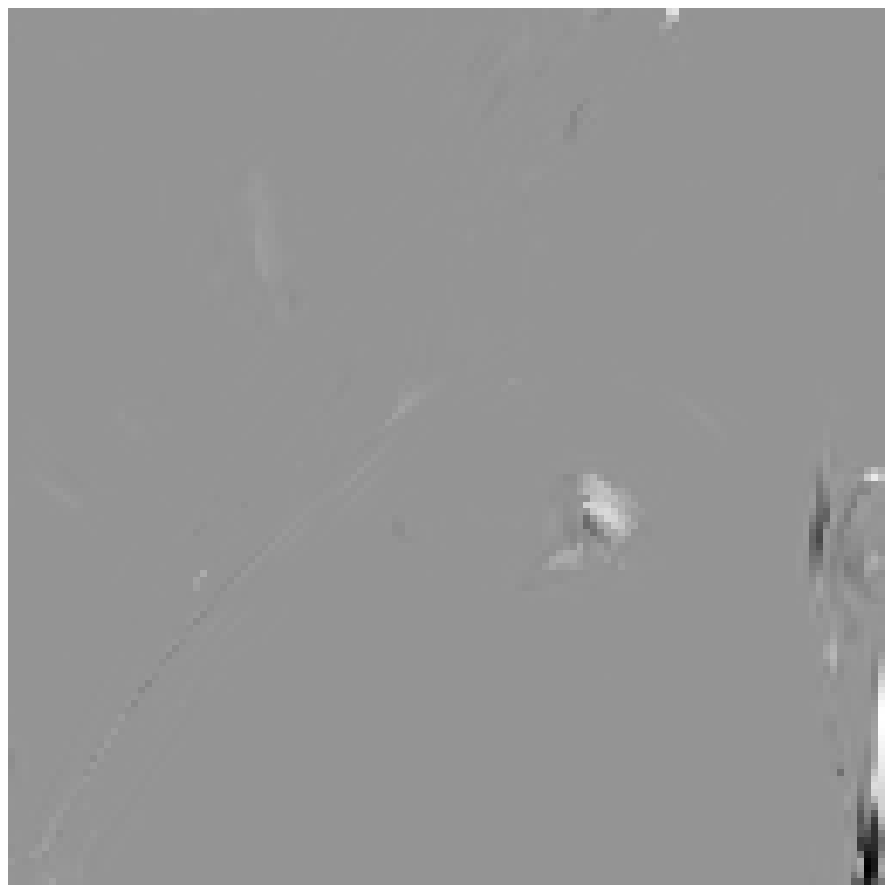} & \includegraphics[scale=0.28]{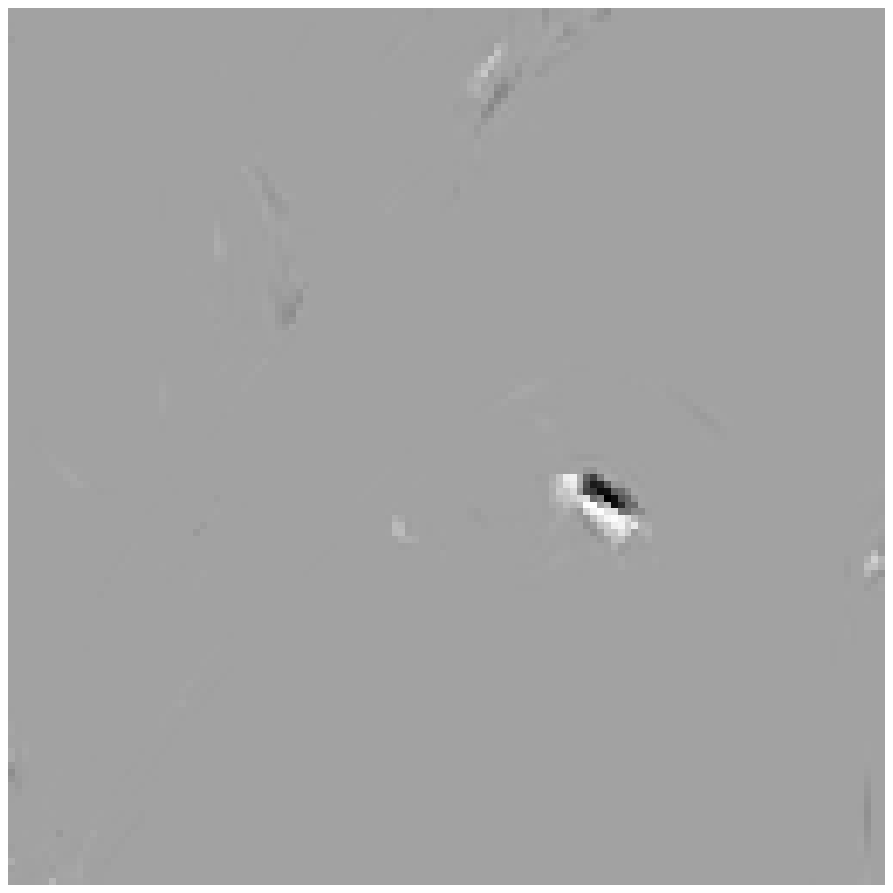}\\
\includegraphics[scale=0.28]{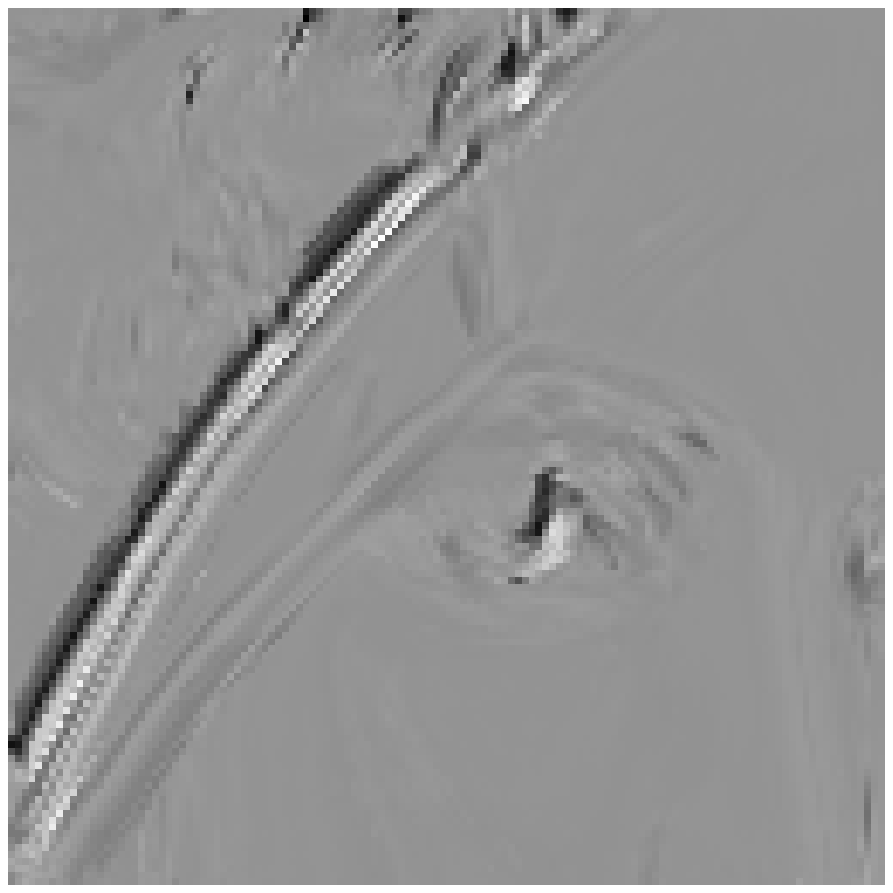} & \includegraphics[scale=0.28]{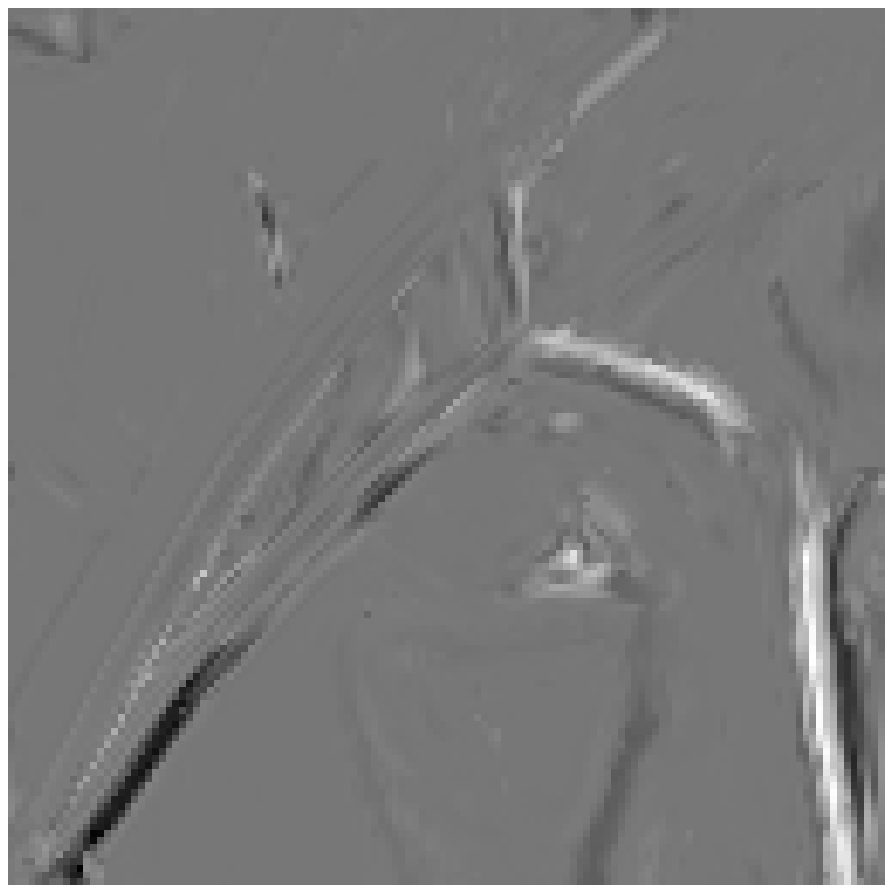} &
\includegraphics[scale=0.28]{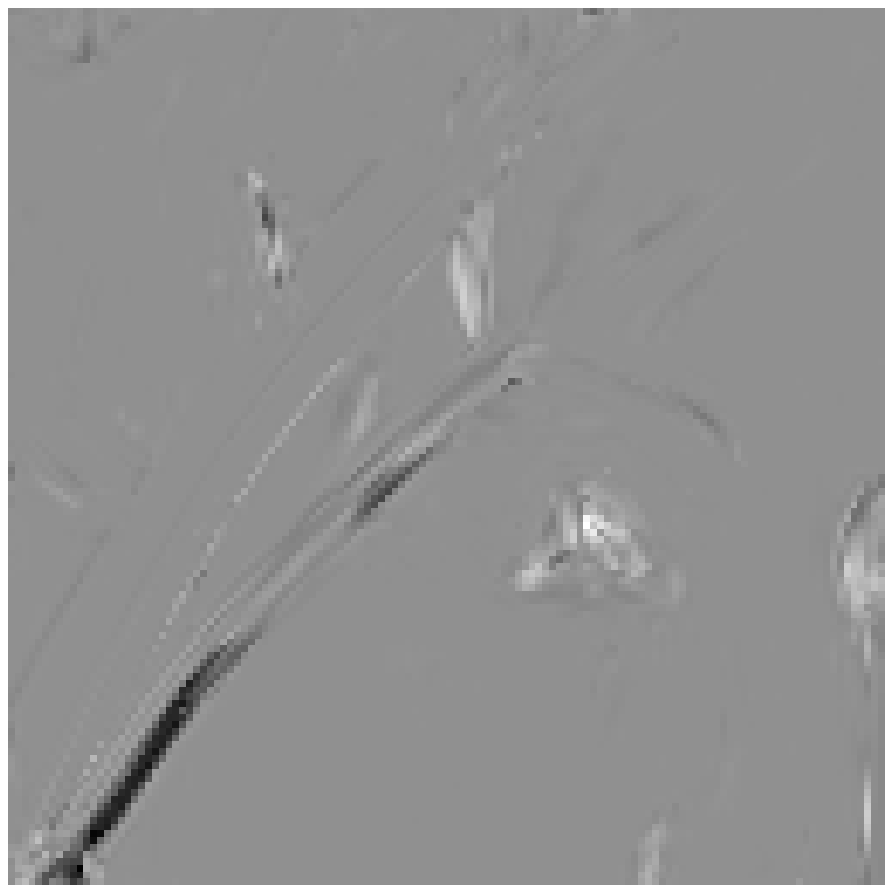} & \includegraphics[scale=0.28]{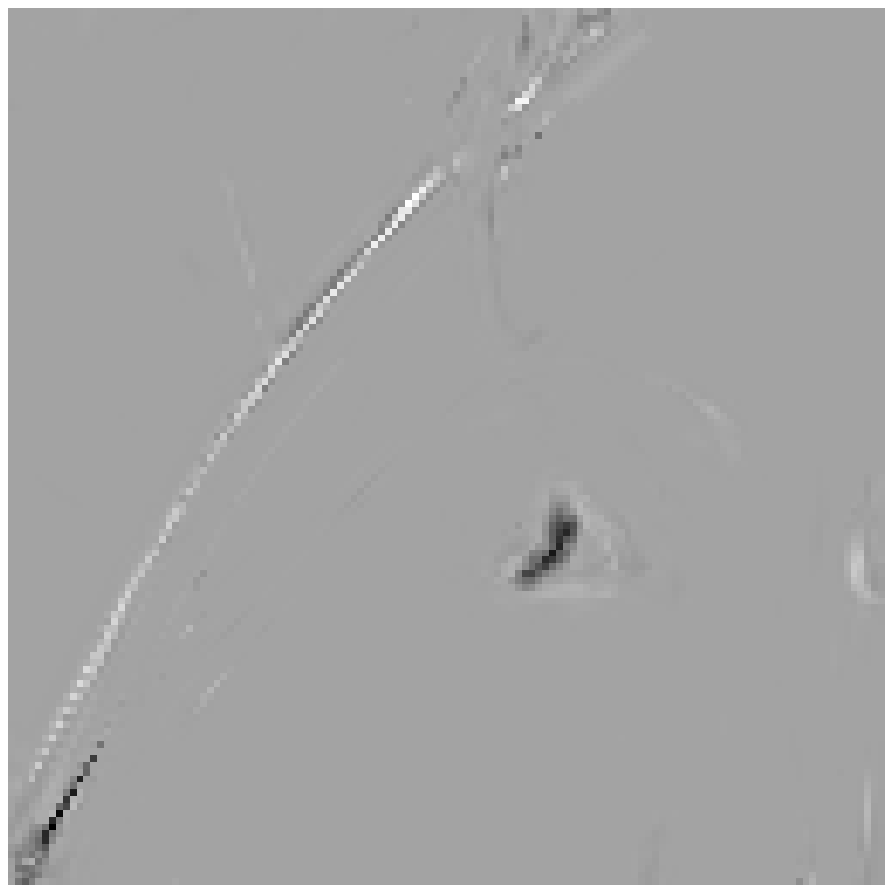} &
\includegraphics[scale=0.28]{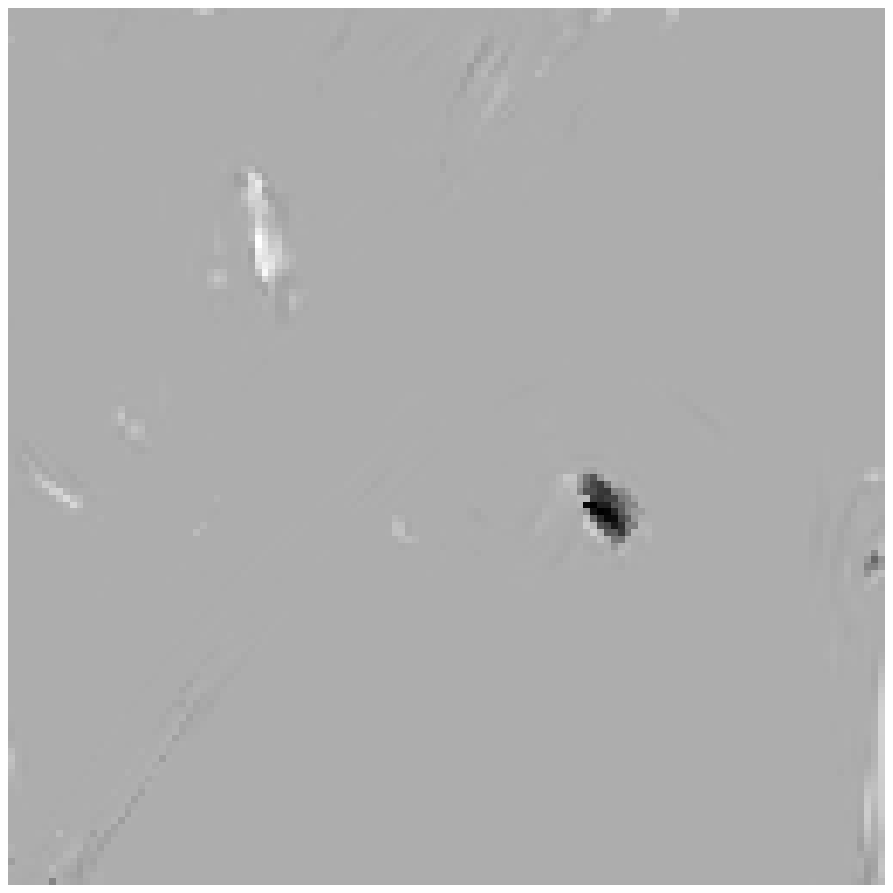} & \includegraphics[scale=0.28]{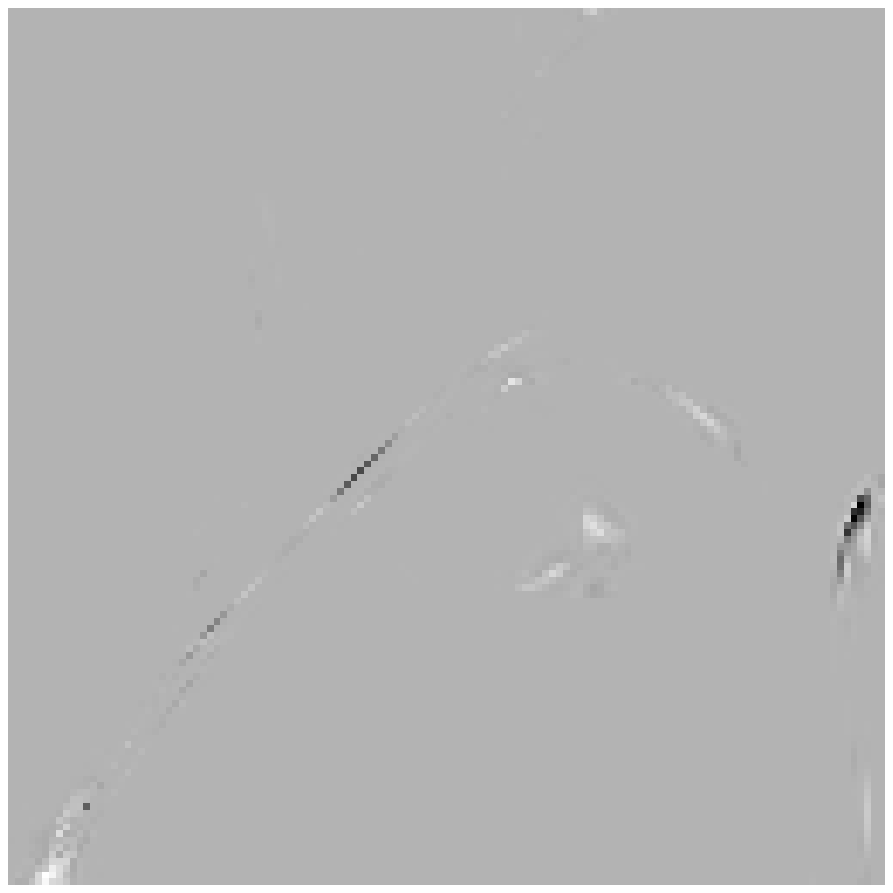}\\
{\small (g)} & {\small (h)} & {\small (i)} & {\small (j)} & {\small (k)} & {\small (l)}\\
\includegraphics[scale=0.28]{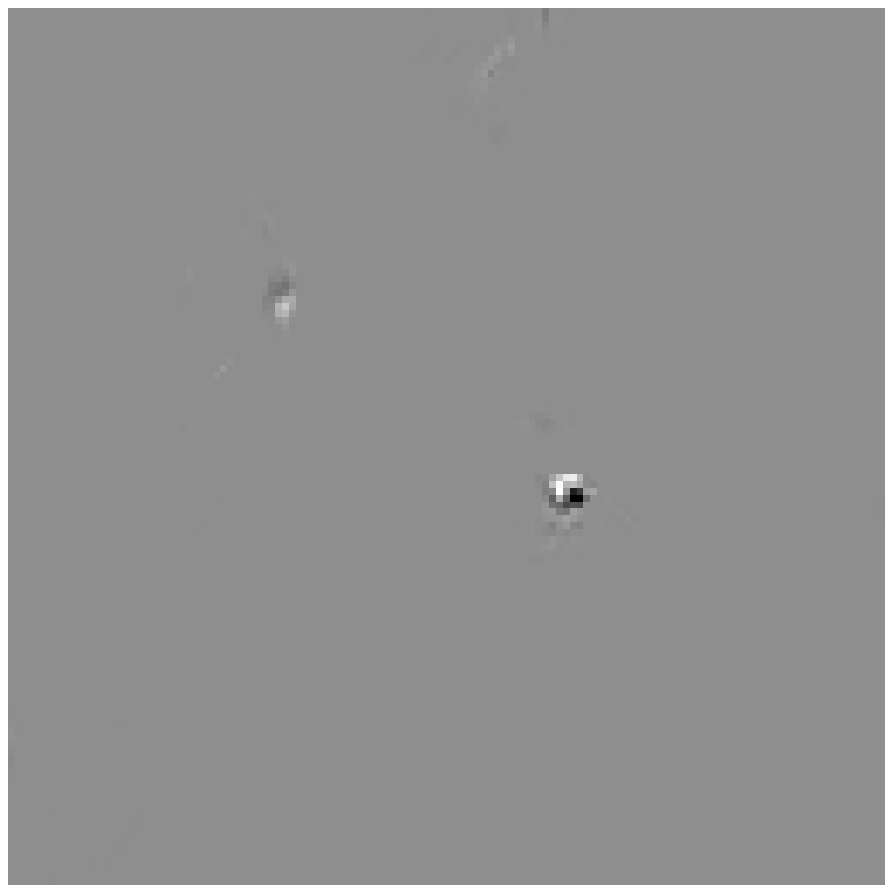} & \includegraphics[scale=0.28]{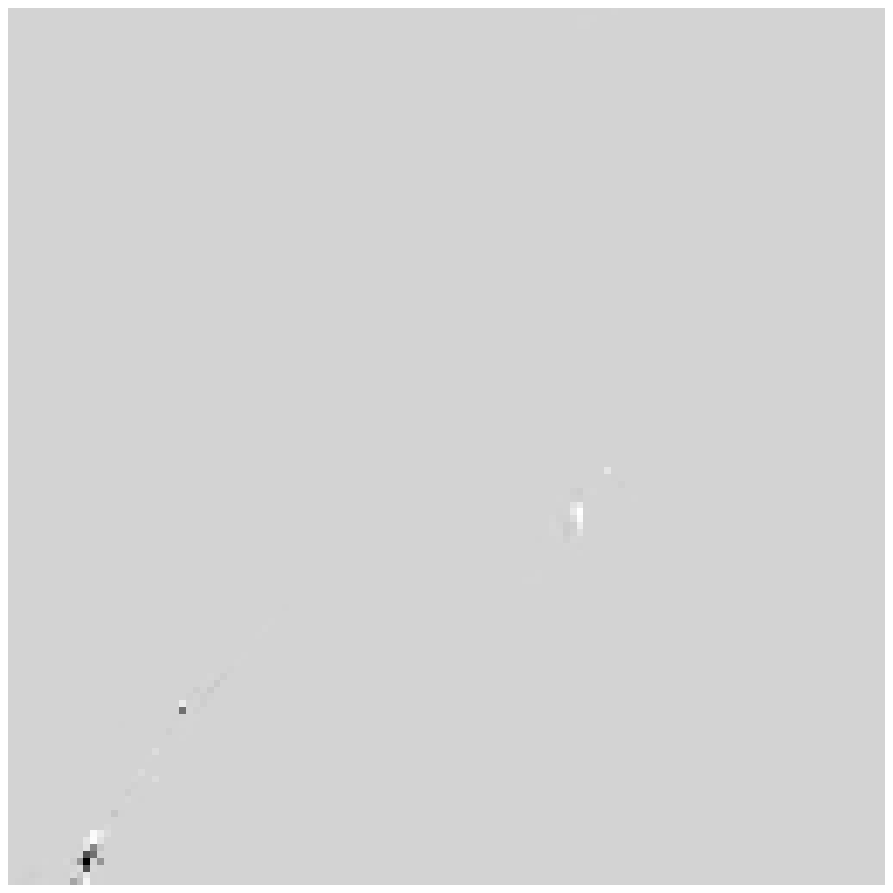} &
\includegraphics[scale=0.28]{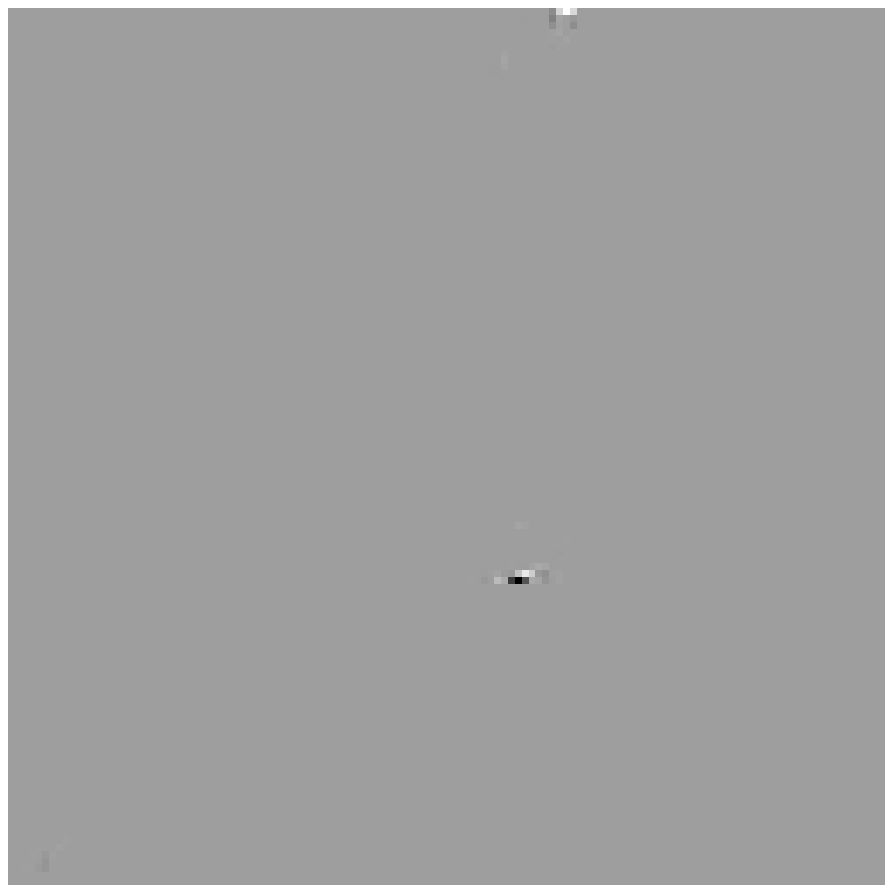} & \includegraphics[scale=0.28]{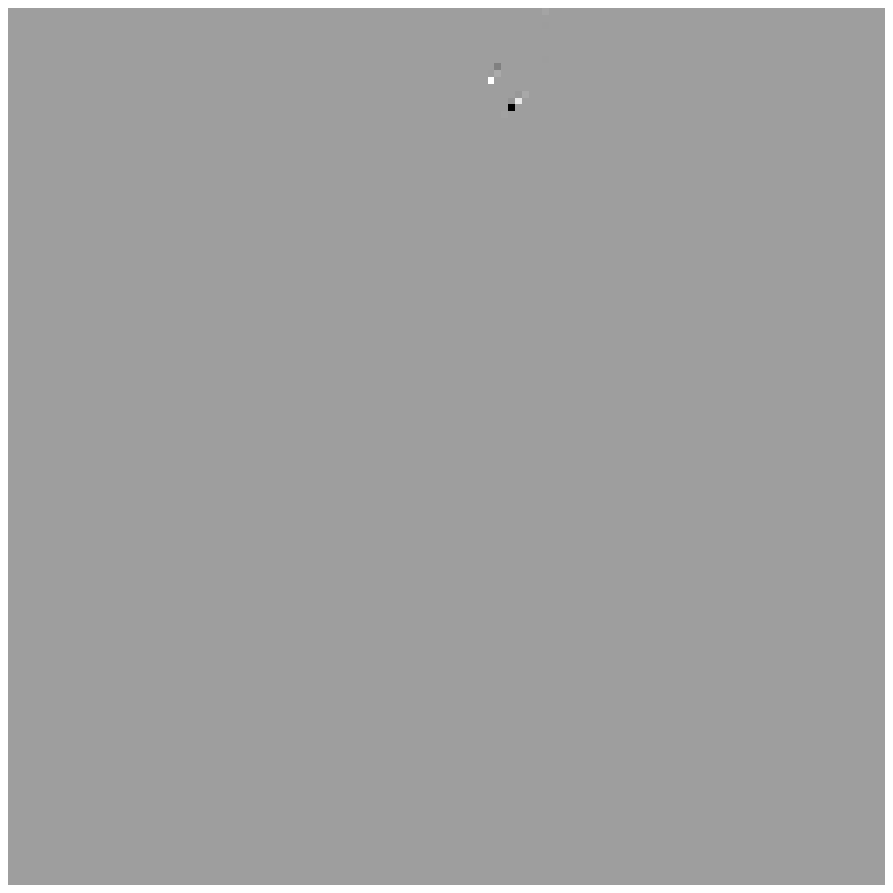} & &\\
\includegraphics[scale=0.28]{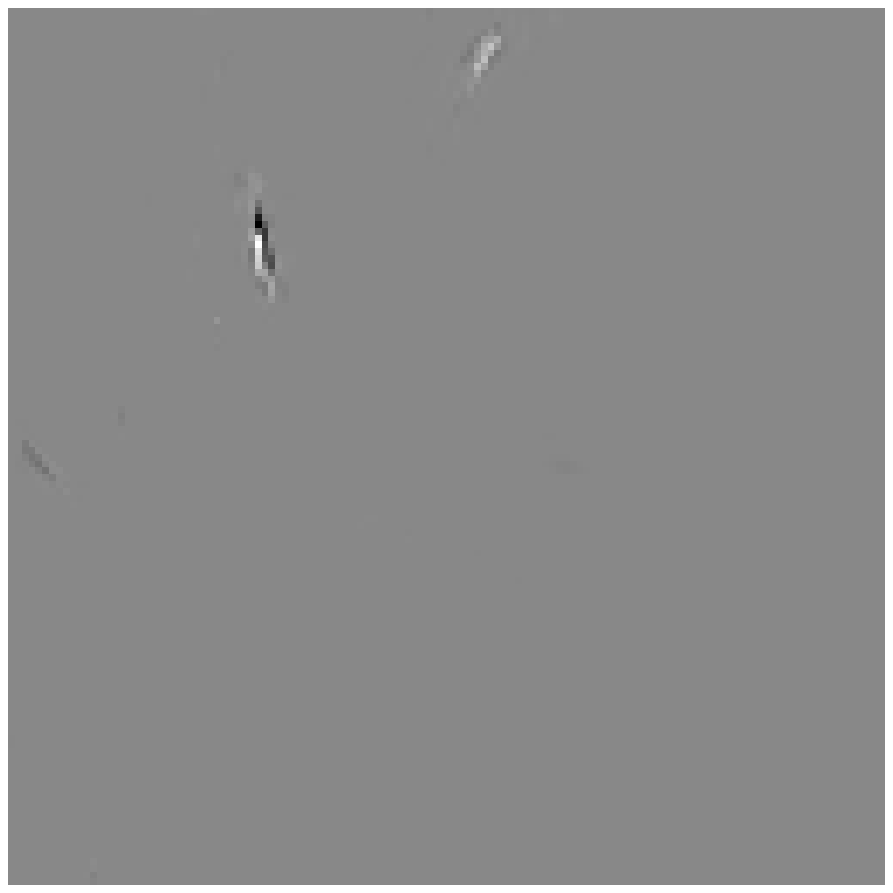} & \includegraphics[scale=0.28]{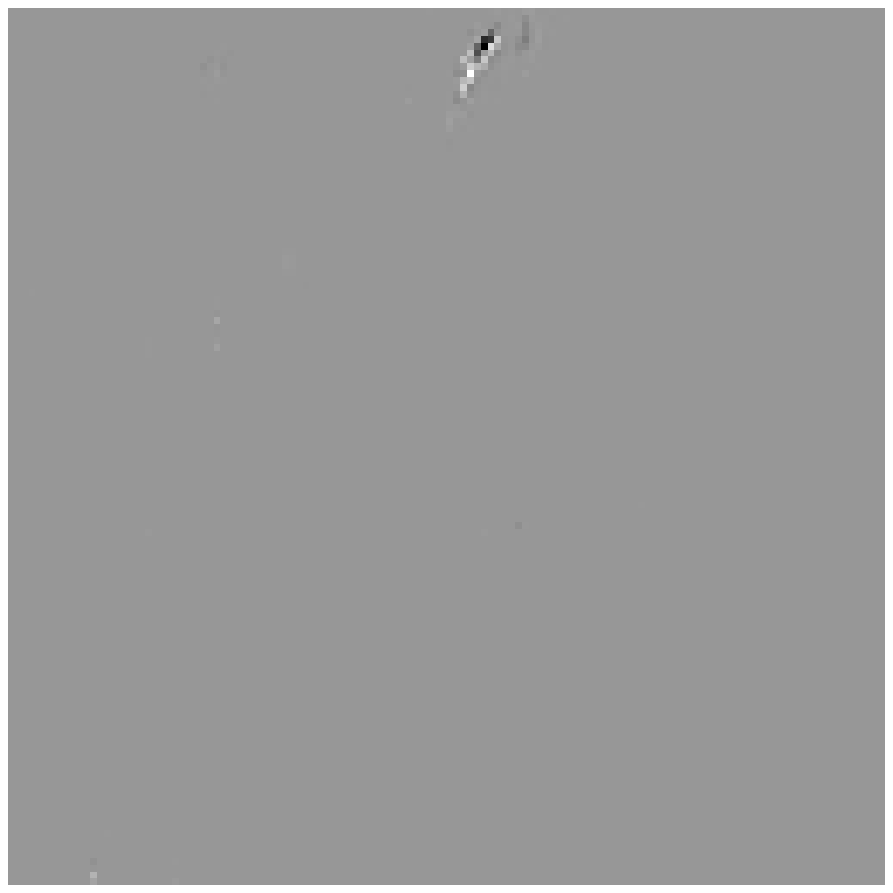} &
\includegraphics[scale=0.28]{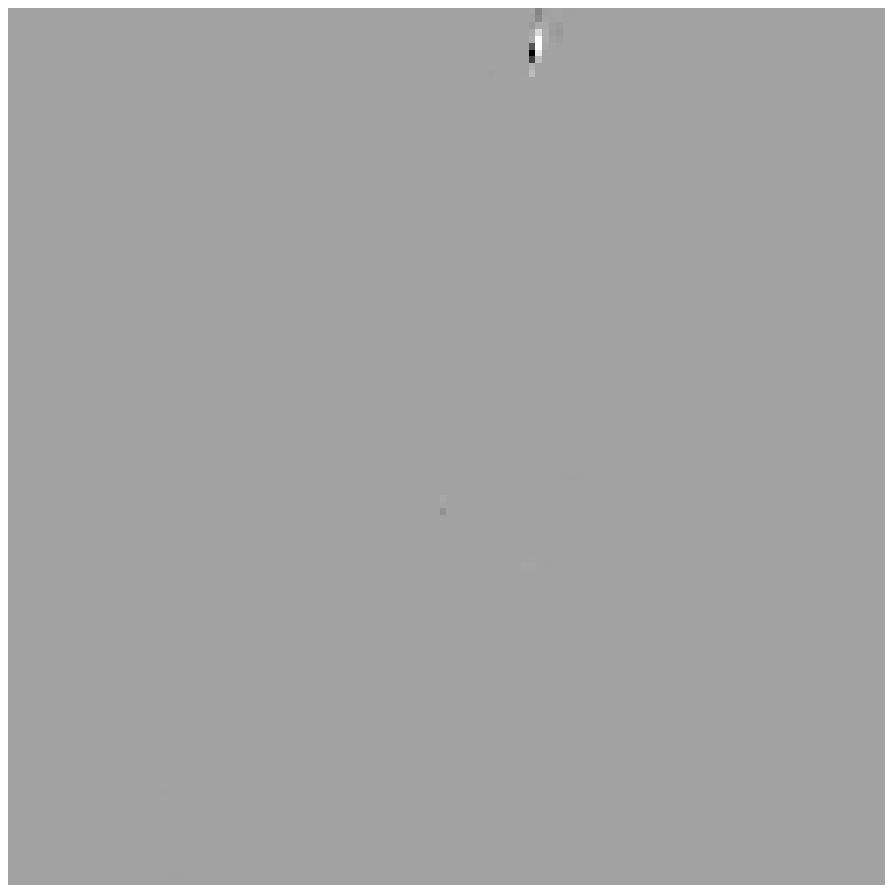} & \includegraphics[scale=0.28]{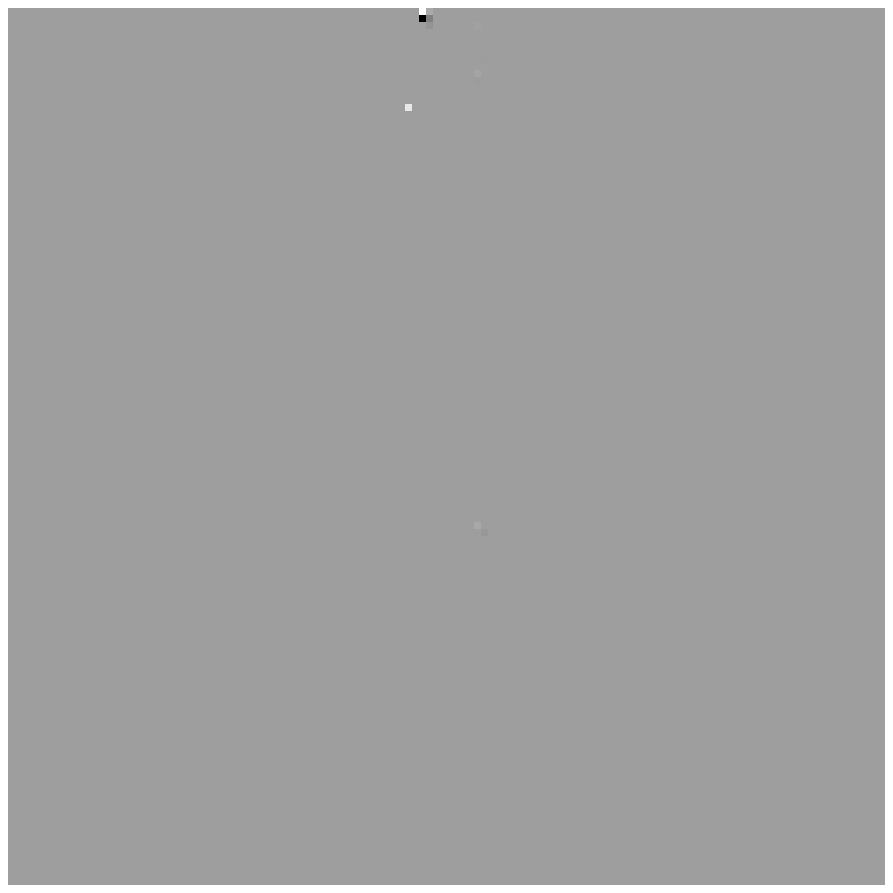} & &\\
{\small (m)} & {\small (n)} & {\small (o)} & {\small (p)} & & \\

\end{tabular}
\caption{Generalized tree-based wavelet basis elements derived from an image: (a) the original image. (b) scaling functions ($\ell$=1).
(c) wavelets ($\ell$=1). (d) wavelets ($\ell$=2). (e) wavelets ($\ell$=3). (f) wavelets ($\ell$=4). (g) wavelets ($\ell$=5).
(h) wavelets ($\ell$=6). (i) wavelets ($\ell$=7). (j) wavelets ($\ell$=8). (k) wavelets ($\ell$=9). (l) wavelets ($\ell$=10).
(m) wavelets ($\ell$=11). (n) wavelets ($\ell$=12). (o) wavelets ($\ell$=13). (p) wavelets ($\ell$=14). }
\label{Figure: wavelets}
\end{figure*}

\textcolor{black}{We use the transforms corresponding to these trees to obtain $m$-term approximations of a column \textcolor{black}{stacked} version of the $128 \times 128$ image shown in Fig. \ref{Figure: wavelets}(a) (the center of the Lena image).
The approximations, shown in Fig. \ref{Figure: smooth m term}, were carried out by keeping the highest coefficients (in absolute value) in the different transform domains.
We compare these results between themselves, and to the $m$-term approximations obtained with the common \ac{1D} and the \ac{2D} separable wavelet transforms (the latter applied to the original image) corresponding to the same wavelet filters.}
Fig. \ref{Figure: smooth m term} shows $m$-term approximation results (PSNR) obtained for the image in Fig. \ref{Figure: wavelets}(a) using different wavelet filters.
It can be seen that the generalized tree-based wavelet transform outperforms the two other transforms for all the wavelet filters that have been used.
It can also be seen that the PSNR gap in the first thousands of coefficients increases with number of vanishing moments of the wavelet filters.
Further, Figure \ref{Figure: smooth m term}(d) shows that the generalized tree-based wavelet results obtained with the db4 and db8 filters are close, and better than the ones obtained with the Haar filter.

Before concluding this section, we are interested to see how the basis elements of the proposed wavelet transform look like.
Each basis element \textcolor{black}{is} obtained by setting the corresponding transform coefficient to 1 and all the other coefficients to zero, and applying the inverse transform.
\textcolor{black}{Unlike the case of common 1D wavelet bases, the series of data derived permutations applied to the reconstructed coefficients in the different stages of the inverse transform, result in basis functions which adapt themselves to the input signal $f$.}

\begin{figure*}[t]
\centering
\begin{tabular}{ccccccc}
\multirow{2}{*}{\includegraphics[scale=0.3]{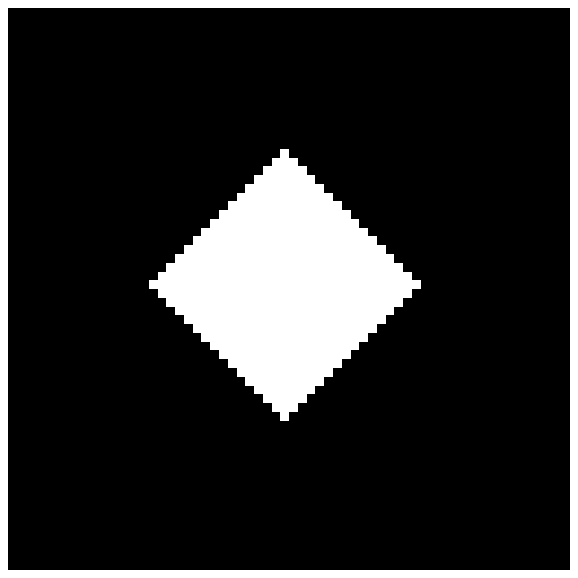}} &
\includegraphics[scale=0.32]{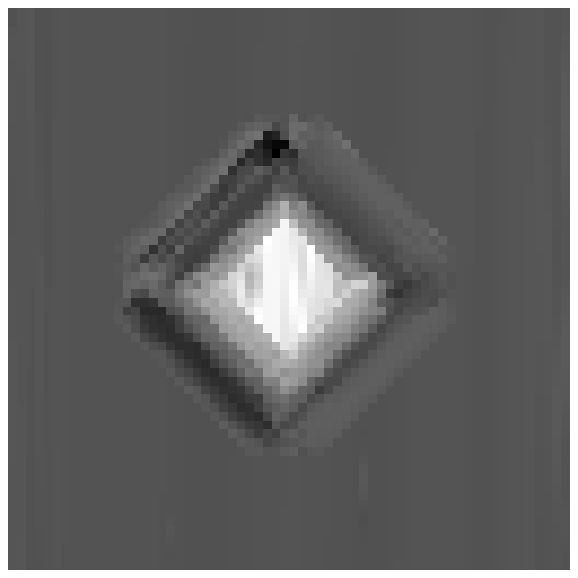} & \includegraphics[scale=0.32]{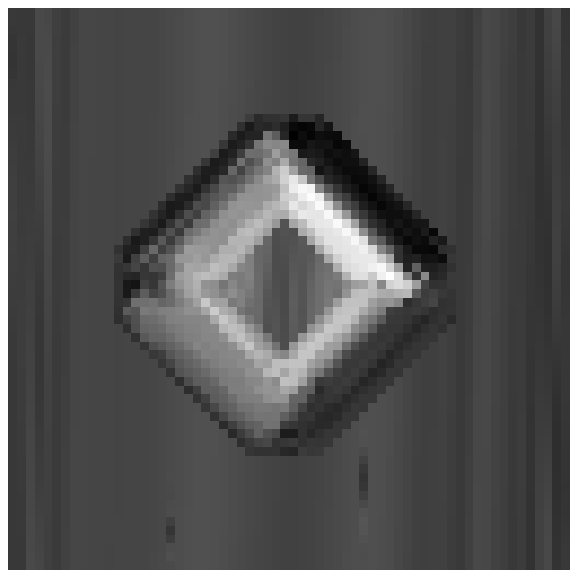} & \includegraphics[scale=0.32]{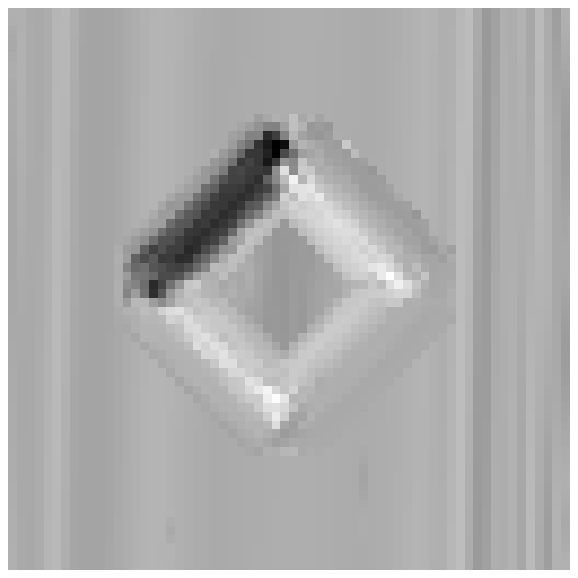} & \includegraphics[scale=0.32]{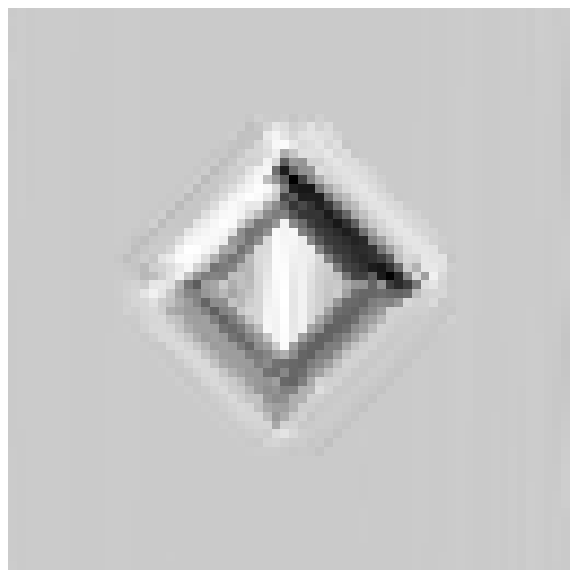} &
\includegraphics[scale=0.32]{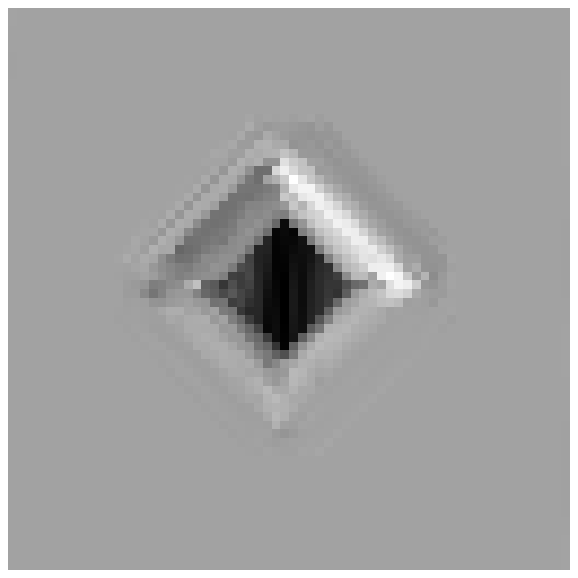} & \includegraphics[scale=0.32]{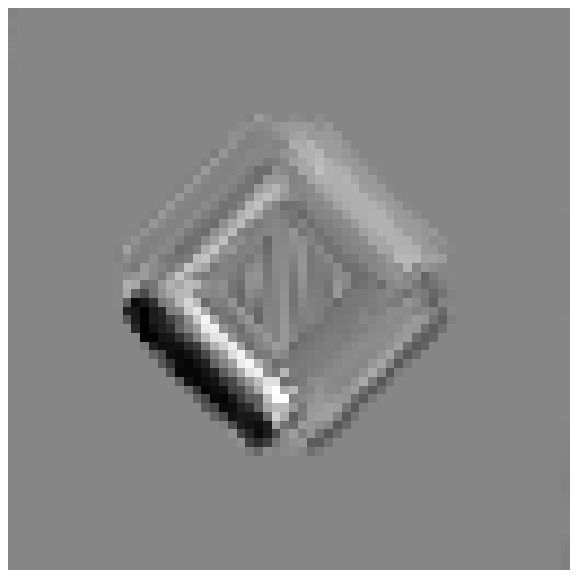}\\
&
\includegraphics[scale=0.32]{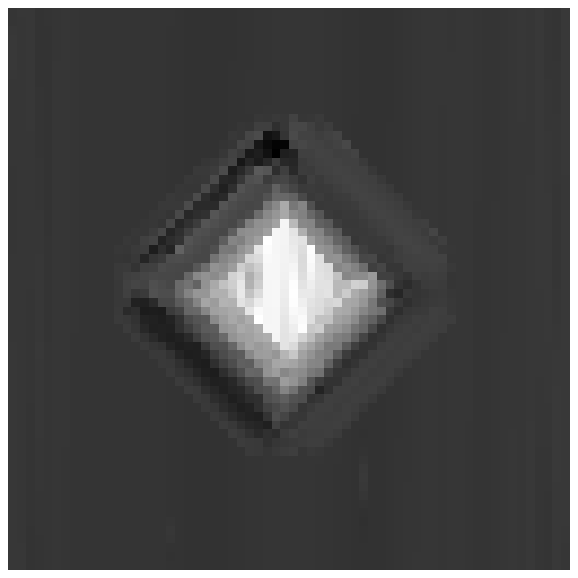} & \includegraphics[scale=0.32]{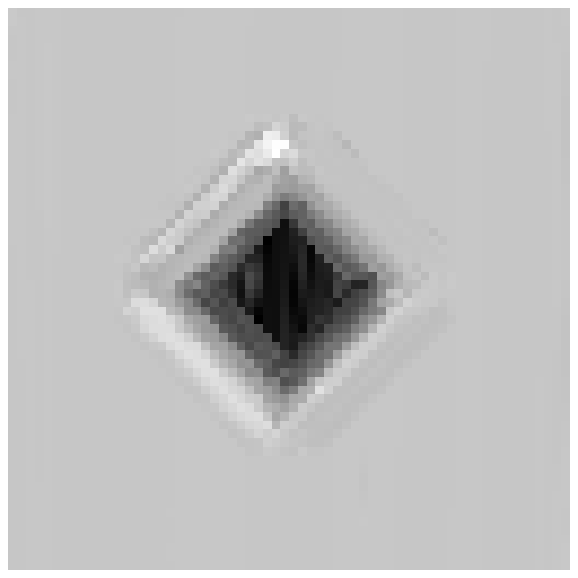} & \includegraphics[scale=0.32]{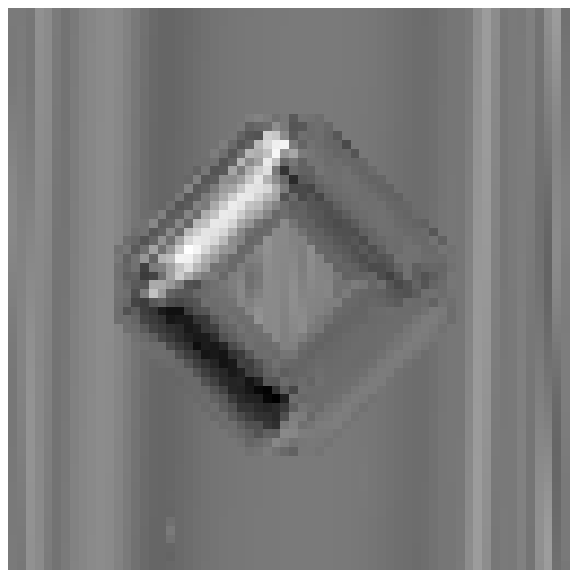} & \includegraphics[scale=0.32]{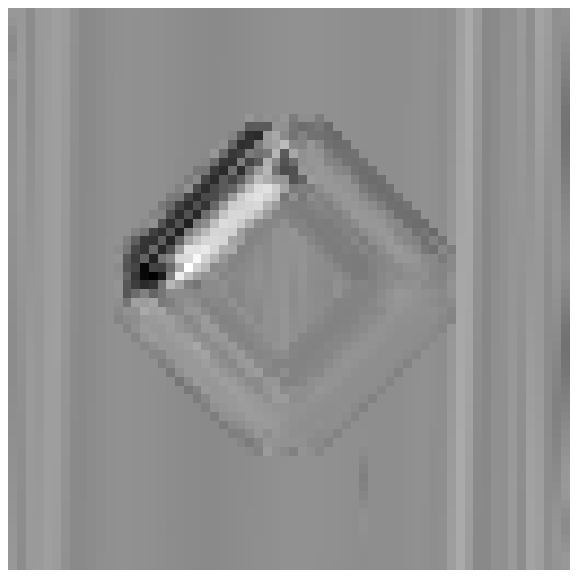} &
\includegraphics[scale=0.32]{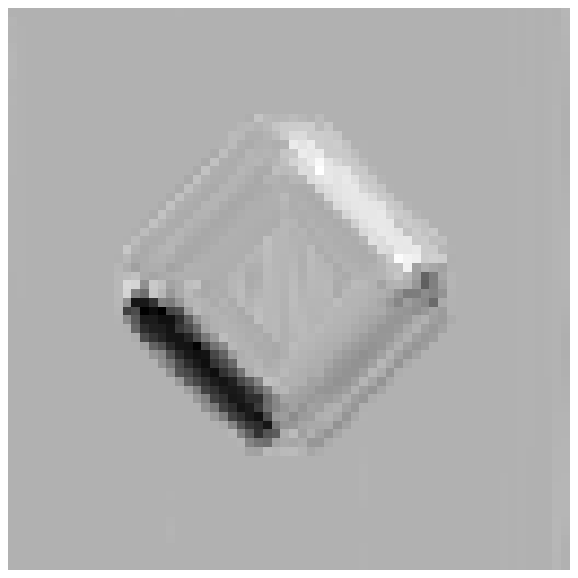} & \includegraphics[scale=0.32]{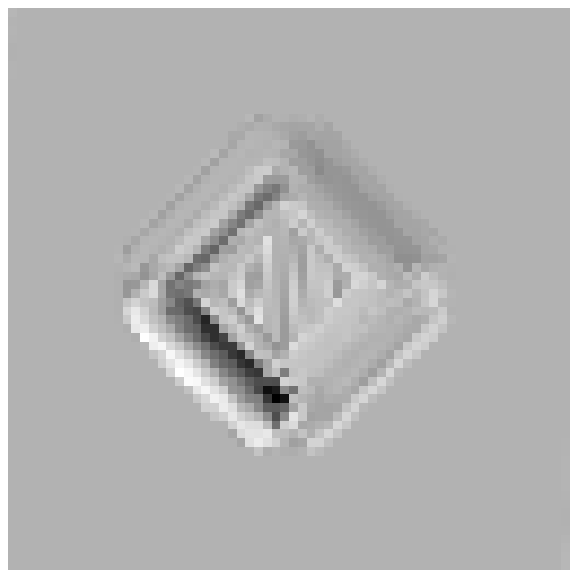}\\
{\small (a)} & {\small (b)} & {\small (c)} & {\small (d)} & {\small (e)} & {\small (f)} & {\small (g)} \\
\includegraphics[scale=0.32]{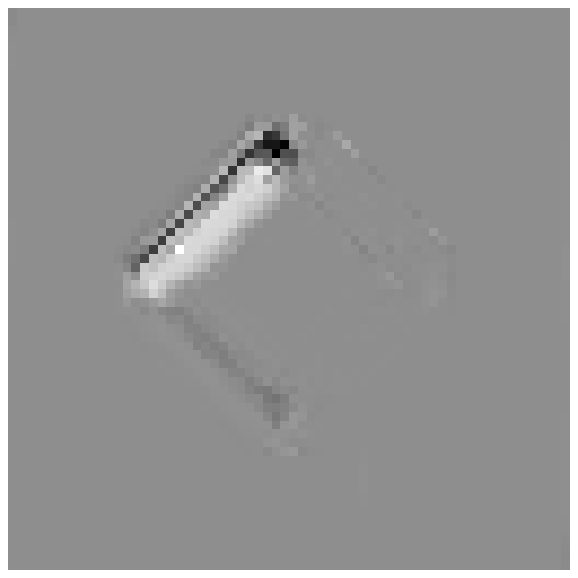} & \includegraphics[scale=0.32]{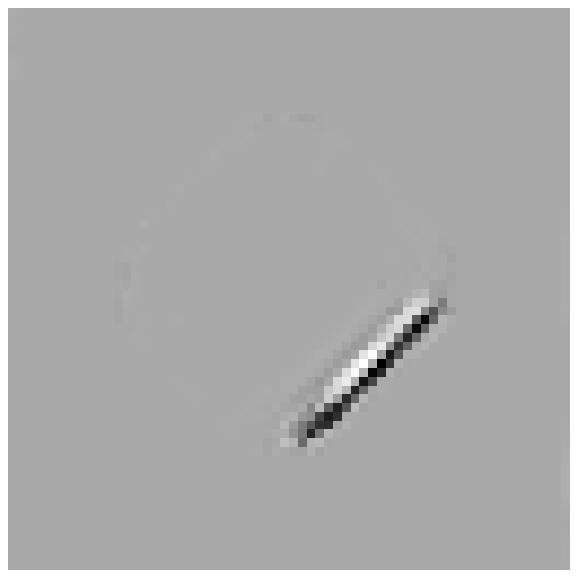} &
\includegraphics[scale=0.32]{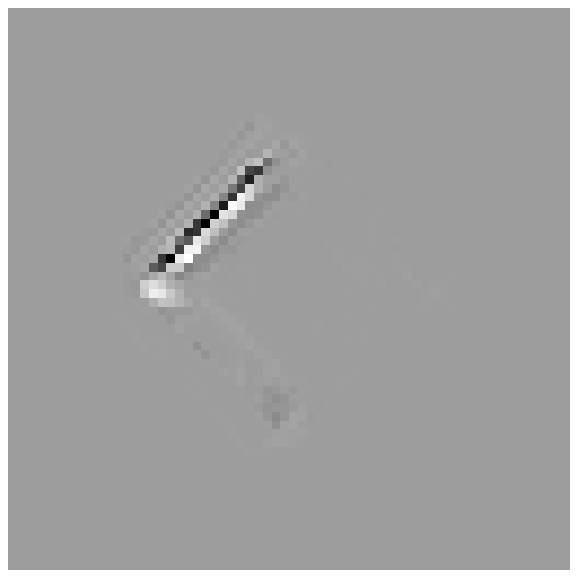} & \includegraphics[scale=0.32]{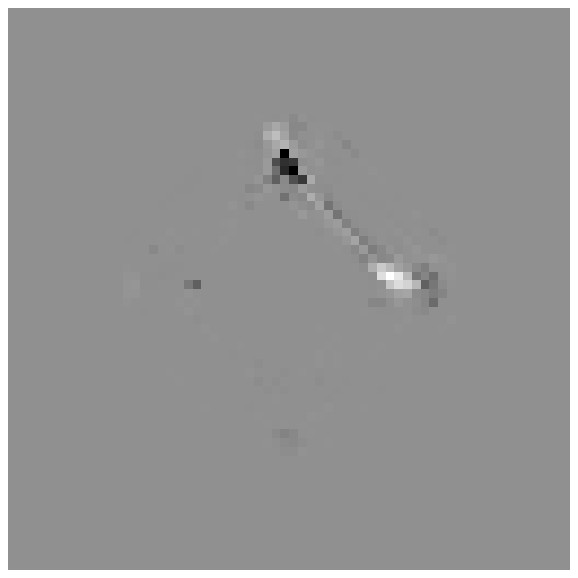} &
\includegraphics[scale=0.32]{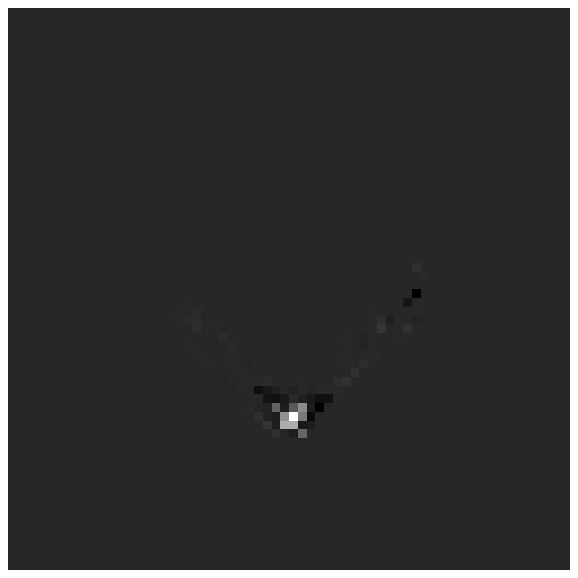} & \includegraphics[scale=0.32]{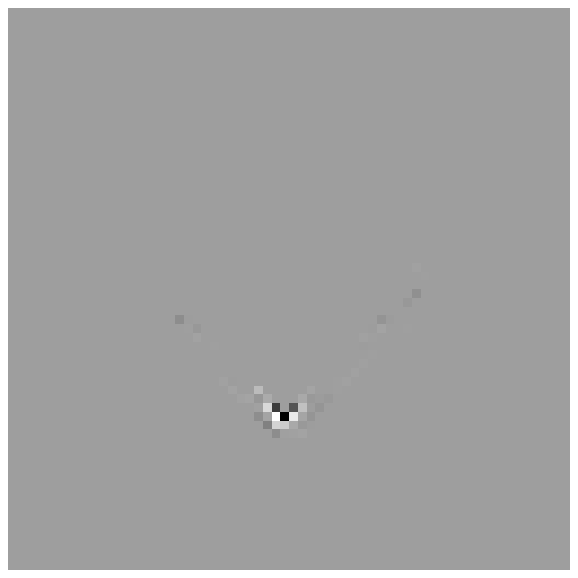} &
\includegraphics[scale=0.32]{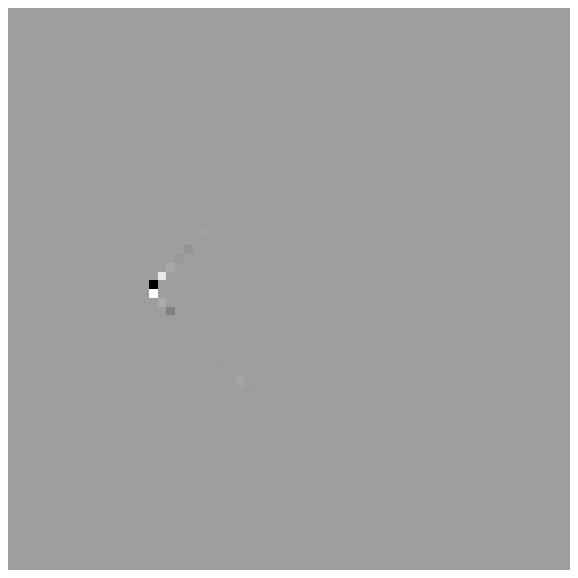}\\
\includegraphics[scale=0.32]{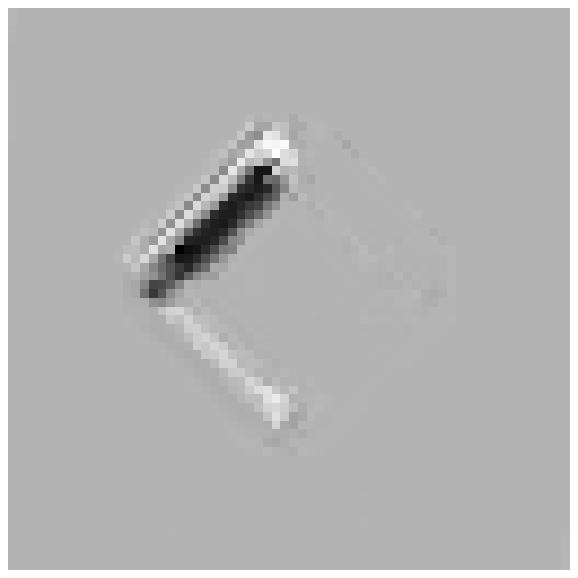} & \includegraphics[scale=0.32]{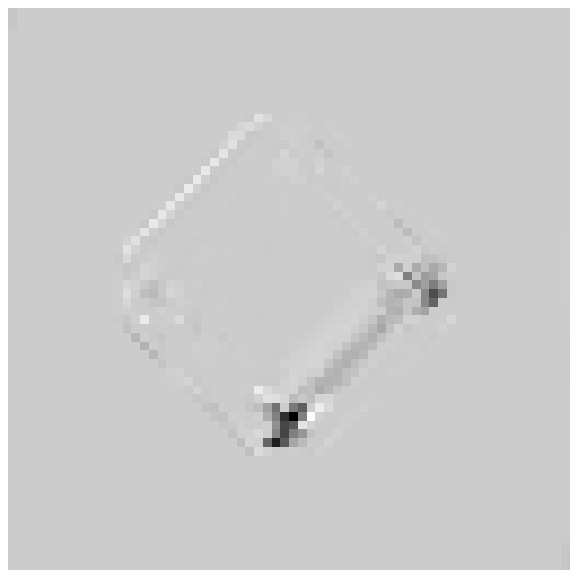} &
\includegraphics[scale=0.32]{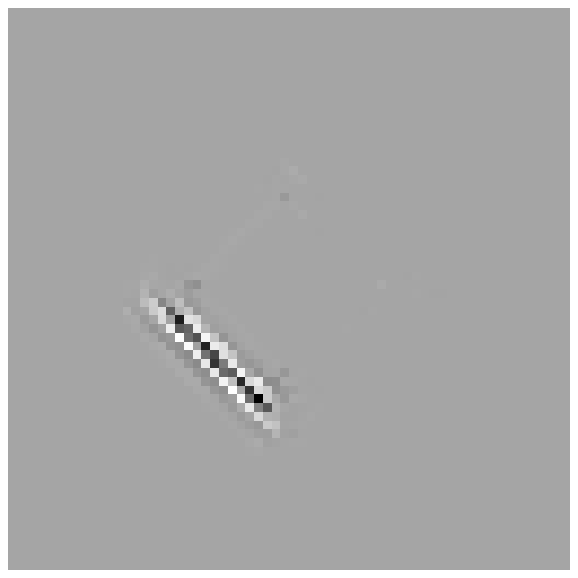} & \includegraphics[scale=0.32]{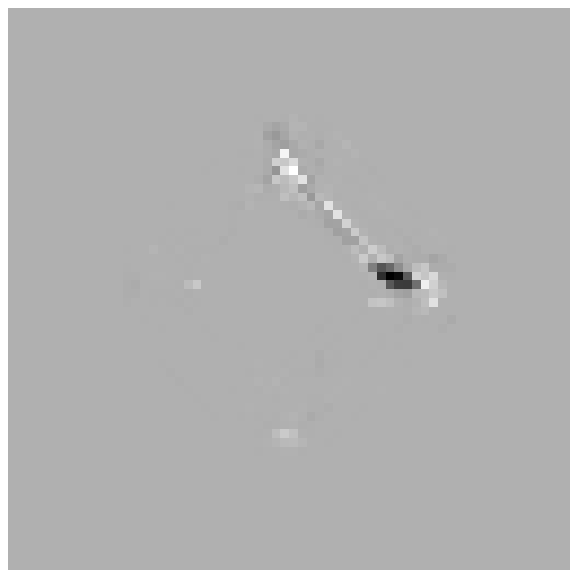} &
\includegraphics[scale=0.32]{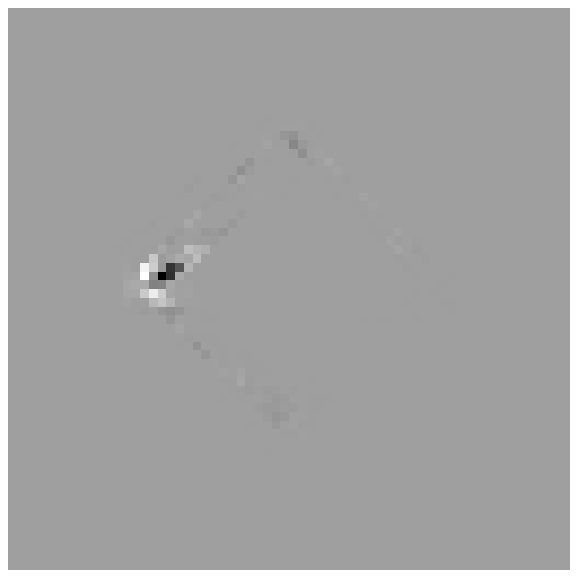} & \includegraphics[scale=0.32]{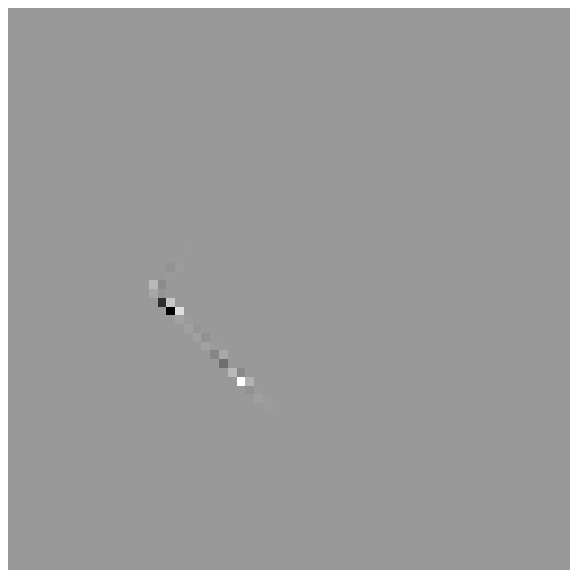} &
\includegraphics[scale=0.32]{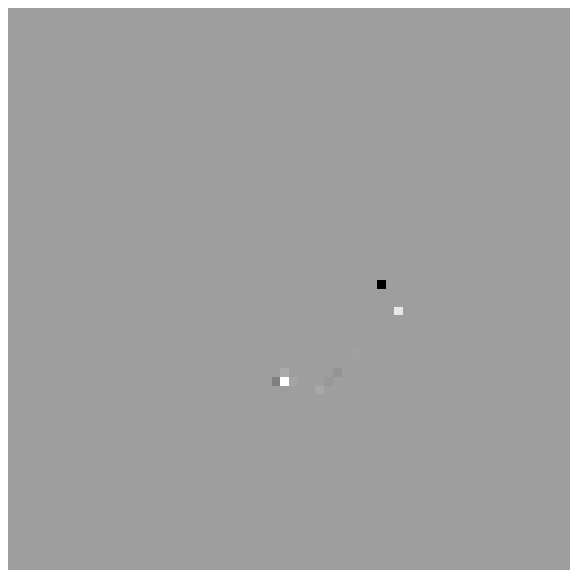}\\
{\small (h)} & {\small (i)} & {\small (j)} & {\small (k)} & {\small (l)} & {\small (m)} & {\small (n)} \\
\end{tabular}
\caption{Generalized tree-based wavelet basis elements derived from a synthetic image: (a) the original image. (b) scaling functions ($\ell$=1).
(c) wavelets ($\ell$=1). (d) wavelets ($\ell$=2). (e) wavelets ($\ell$=3). (f) wavelets ($\ell$=4). (g) wavelets ($\ell$=5).
(h) wavelet ($\ell$=6). (i) wavelet ($\ell$=7). (j) wavelet ($\ell$=8). (k) wavelets ($\ell$=9). (l) wavelets ($\ell$=10).
(m) wavelets ($\ell$=11). (n) wavelets ($\ell$=12).}
\label{Figure: synthetic wavelets}
\end{figure*}

As a first example, we examine the basis elements obtained for the synthetic image of size $64 \times 64$, shown in Fig \ref{Figure: synthetic wavelets}(a).
The image contains in its middle a square rotated in an angle of 45 degrees.
\textcolor{black}{We visualize the basis elements as images by reshaping them to the size of the input image.}
Figs \ref{Figure: synthetic wavelets}(b)-(n) show some of the basis elements corresponding to the synthetic image,
and obtained with the Symmlet 8 filter.
The figures show two scaling functions from level $\ell = 1$, and two wavelet basis functions from each level $\ell = 1,\ldots,12$,
all corresponding to the two largest coefficients in the same level.
We note that the reason we have more than one scaling function and one wavelet basis element in coarsest level $\ell=1$ is related to our implementation of the transform.
We use symmetric padding in the signal boundaries before applying it the wavelet filters,
which slightly increases the number of coefficients (and corresponding basis elements) obtained in each level of the tree.
It can be seen how the scaling functions and wavelets in the low levels of the tree ($\ell=1,\ldots,4$) represent the low frequency information in the image,
i.e. smooth versions of the rectangle.
It can also be seen that the wavelet functions represent finer edges as the level of the tree increases,
and that they successfully manage to capture edges which are not aligned with the vertical and horizontal axes.

We next examine the wavelet basis elements corresponding to the image in Fig. \ref{Figure: wavelets}(a),
obtained with the Symmlet 8 filter.
Figs. \ref{Figure: wavelets}(b)-(p) show two scaling functions from level $\ell = 1$, and two wavelet basis functions from each level $\ell = 1,\ldots,14$,
all corresponding to the two largest coefficients in the same level.
It can be seen how the basis functions adapt to the shapes in the images.
Here again the scaling functions and wavelets in the low levels of the tree ($\ell=1,\ldots,4$) represent the low frequency information in the image,
and the wavelet functions represent finer edges in the image as the level of the tree increases.
We next present the application of the generalized tree-based wavelet transform to image denoising.

\section{Image Denoising \textcolor{black}{using} the Generalized tree-Based Wavelet Transform}

\subsection{The Basics}

Let $\mathbf{F}$ be an image of size $\sqrt{N}\times\sqrt{N}$, and let $\tilde{\mathbf{F}}$ be its noisy version:
\begin{equation}
\tilde{\mathbf{F}}=\mathbf{F}+\mathbf{Z}.
\end{equation}
$\mathbf{Z}$ is a matrix of size $\sqrt{N}\times\sqrt{N}$ which denotes an additive white Gaussian noise independent of $\mathbf{F}$ with zero mean and
variance $\sigma^2$.
Also, let $\mathbf{f}$ and $\tilde{\mathbf{f}}$ be the column \textcolor{black}{stacked} representations of $\mathbf{F}$ and $\tilde{\mathbf{F}}$, respectively.
Our goal is to reconstruct $\mathbf{f}$ from $\tilde{\mathbf{f}}$ using the generalized tree-based wavelet transform.
To this end, we first extract the feature points $\mathbf{x}_i$ from $\tilde{\mathbf{F}}$ similarly to the way they were extracted from $\mathbf{F}$ in the previous section. Let $\tilde{f}_i$ be the $i$-th sample in $\tilde{\mathbf{f}}$, then here we choose the point $\mathbf{x}_i$ associated with it as the $9 \times 9$ patch
around the location of $\tilde{f}_i$ in the image $\tilde{\mathbf{F}}$.
We note that since different features are used for the clear and noisy images, the corresponding trees derived from these images will also be different.
Nevertheless, it was shown in \cite{buades2006review},\cite{szlam2008regularization},\cite{singer2009diffusion}
that the distance between the noisy patches $\mathbf{x}_i$ and $\mathbf{x}_j$
is a good predictor for the similarity between the clear versions of their middle pixels $f_i$ and $f_j$.
This means that the tree construction is roughly robust to noise,
and that the obtained tree will capture the geometry and structure of the clear image quite well.
We will further discuss the noise robustness of the algorithm in Section \ref{noise robustenss}.

We next construct the tree according to the scheme described in Algorithm 4,
where the dissimilarity between the points $\mathbf{c}_i^\ell$ and $\mathbf{c}_j^\ell$ in level $\ell$ is again measured by the squared Euclidean distance between them,
which can be found in the $(i,j)$ location in $\textcolor{black}{\mathbf{W}^\ell}$.

Similarly to Gavish et al. \cite{gavish2010mwot},
we use \textcolor{black}{an approach which resembles the} ``cycle spinning'' method \cite{coifman1995translation}
in order to \textcolor{black}{smooth the} image denoising \textcolor{black}{outcome}.
This means that we randomly construct $10$ different trees,
utilize the transforms corresponding to each of them to denoise $\tilde{\mathbf{f}}$,
and average the produced images.
Each level of a random tree is constructed by choosing the first point at random,
and then continue from each point $\mathbf{x}_{j_0}$ to its nearest neighbor $\mathbf{x}_{j_1}$ with a probability
$p_1\varpropto \exp(-\|\mathbf{x}_{j_0}-\mathbf{x}_{j_1}\|^2 / \epsilon)$,
or to its second nearest neighbor \textcolor{black}{$\mathbf{x}_{j_2}$} with a probability
$p_2\varpropto \exp(-\|\mathbf{x}_{j_0}-\mathbf{x}_{j_2}\|^2 / \epsilon)$, where here we set $\epsilon=0.1$.

The denoising itself is performed by applying the proposed wavelet transform
(derived from the tree of $\tilde{\mathbf{f}}$) on $\tilde{\mathbf{f}}$,
using hard thersholding on the transform coefficients, and computing the inverse transform.

In order to assess the performance of the proposed denoising algorithm we first apply it,
using different wavelet filters, to a noisy version of the image shown in Fig. \ref{Figure: wavelets}(a).
For each of the transforms we perform $5$ experiments for different realizations of noise with standard deviation $\sigma=25$ and average the results
-- these averages are given in Table \ref{Table: lena_face}.
\textcolor{black}{We note that the denoising thresholds were manually found to produce good denoising results,
as the theoretical wavelet threshold $\sigma \sqrt{2\log_e{N}}$ led to poorer results, with PSNR values which were lower by about 0.6 dB.}
It can be seen that better results are obtained with the Symmlet wavelets, and that generally better results are obtained with wavelets with a high number of vanishing moments. It can also be seen that all the transforms require about $320$ coefficients to represent the image
(for each of the 10 results produced by the different trees) which is about $2$ percents of the number of coefficients required in the original space.

We next apply the proposed scheme with the Symmlet 8 wavelet to noisy versions of the images Lena and Barbara,
with noise standard deviation $\sigma=25$ and PSNR of $20.17$ dB.
We note that this time we use patches of size $13 \times 13$ and perform only one experiment for each image, for one realization of noise.
The clear, noisy and recovered images can be seen in Fig. \ref{Figure: full Images}.
For comparison reasons, we also apply to the two images the denoising scheme of Elad et al. \cite{elad2006image},
which utilize the K-SVD algorithm \cite{aharon2006k}.
We chose to compare our results to the ones obtained by this scheme, as it also employs an efficient representation of the image content,
and is one of many state-of-the-art image denoising algorithms \cite{mairal2008sparse},\cite{mairal2010non}, which are based on sparse and redundant representations \cite{bruckstein2009sparse}.
However, we note that this scheme is based on efficient representations of the image patches,
while our scheme is based on efficient representation of the whole image.
The PSNR of the results obtained with the proposed scheme and the K-SVD algorithm are shown in Table \ref{Table: full_images_PSNR}.
It can be seen that the results obtained by our algorithm are inferior compared to the ones obtained with the K-SVD.
We next try to improve the results produced by our proposed scheme by adding an averaging element into it,
which is also a variation on the ``cycle spinning'' method.

\subsection{Subimage averaging}

Let $\mathbf{X}_p$ be an $\textcolor{black}{n \times (\sqrt{N}-\sqrt{n}+1)^2}$ matrix,
containing column \textcolor{black}{stacked} versions of all the
$\textcolor{black}{\sqrt{n} \times \sqrt{n}}$ patches $\mathbf{x}_i$ inside the image.
When we built a tree for the image, we assumed that each patch is associated only with its middle pixel.
Therefore the tree was associated with the signal composed of the middle points in the patches,
which is the middle row of $\mathbf{X}_p$, and the transform was applied to this signal.
However, we can alternatively choose to associate all the patches with a pixel located in a different position,
for example the top right pixel in each patch.
Since the whole patches are used in the construction of the tree,
effectively this means that the tree can be associated with any one of the $N_p$ signals located in the rows of $\mathbf{X}_p$.
These signals are the column \textcolor{black}{stacked} versions of the $\textcolor{black}{n}$ subimages of size
$\textcolor{black}{(\sqrt{N}-\sqrt{n}+1) \times (\sqrt{N}-\sqrt{n}+1)}$,
whose top right pixel reside in the top right
$\textcolor{black}{\sqrt{n} \times \sqrt{n}}$ patch in the image.
We next apply the generalized tree-based wavelet transform denoising scheme
to each of these \textcolor{black}{$n$} signals.
We then plug each subimage into its original place in the image, and average the different values obtained for each pixel,
similarly to the way an image is reconstructed from its patches in the K-SVD image denoising scheme \cite{elad2006image}.
We hereafter refer to scheme described above as Subimage Averaging (SA).

\textcolor{black}{The subimage averaging scheme can also be viewed a little differently.
The tree also defines a wavelet transform on the data points $\mathbf{x}_j$,
where the approximation coefficient vectors in level $\ell$ are the points $\mathbf{c}_j^\ell$.
We denote detail coefficients vectors in level $\ell$ as $\mathbf{q}_j^\ell$.
These vectors can be calculated by applying the filter $\bar{\mathbf{g}}^T$ to the matrix $\mathbf{C}_{\ell+1}^p$
and decimating the columns of the outcome by a factor of 2.
A matrix $\mathbf{Q}$ containing $\mathbf{c}_1^1$ and all the detail coefficient vectors $\{\mathbf{q}_j^\ell\}_{\ell=1}^{L-1}$ can also be obtained by applying the generalized tree-based wavelet transform to each row of the matrix $\mathbf{X}_p$.
Similarly, the inverse transform can be performed on $\mathbf{Q}$ by applying each row of this matrix the generalized tree--based wavelet inverse transform.
Therefore the subimage averaging scheme can be viewed as applying the transform derived from the tree directly on the image pathes,
performing hard thresholding on them, applying the inverse transform to the coefficient vectors,
and reconstructing the image from the clean patches.}

The results obtained with the generalized tree-based wavelet transform, combined with the subimage averaging procedure,
for the noisy version of the image in Fig. \ref{Figure: wavelets}(a) are shown in Table \ref{Table: lena_face}.
It can be seen that this procedure increases the PSNR of the results by about $0.85$ dB.
We note that since we use $9 \times 9$ patches, each of the 10 images averaged in the cycle spinning procedure is reconstructed from 81 subimages,
and it can be seen that \textcolor{black}{on average} about 360 transform coefficients are required to represent each one of these subimages.
Thus the number of transform coefficients required to represent each subimage is slightly higher than the one required to represent an image when the subimage averaging is not used, but is still much lower than the number required in the original space.

\begin{figure*}[t]
\centering
\begin{tabular}{cccc}
\includegraphics[scale=0.3]{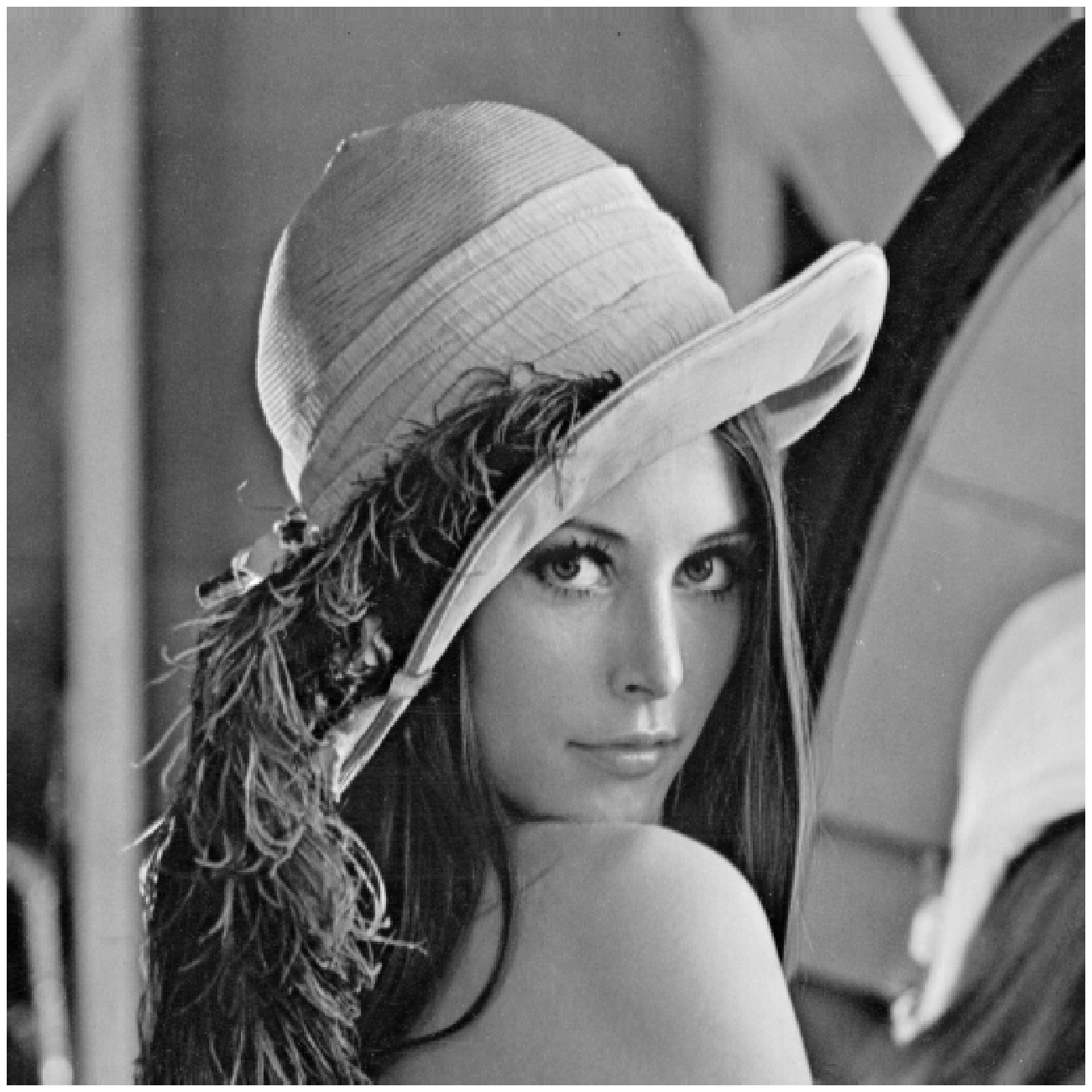} & \includegraphics[scale=0.3]{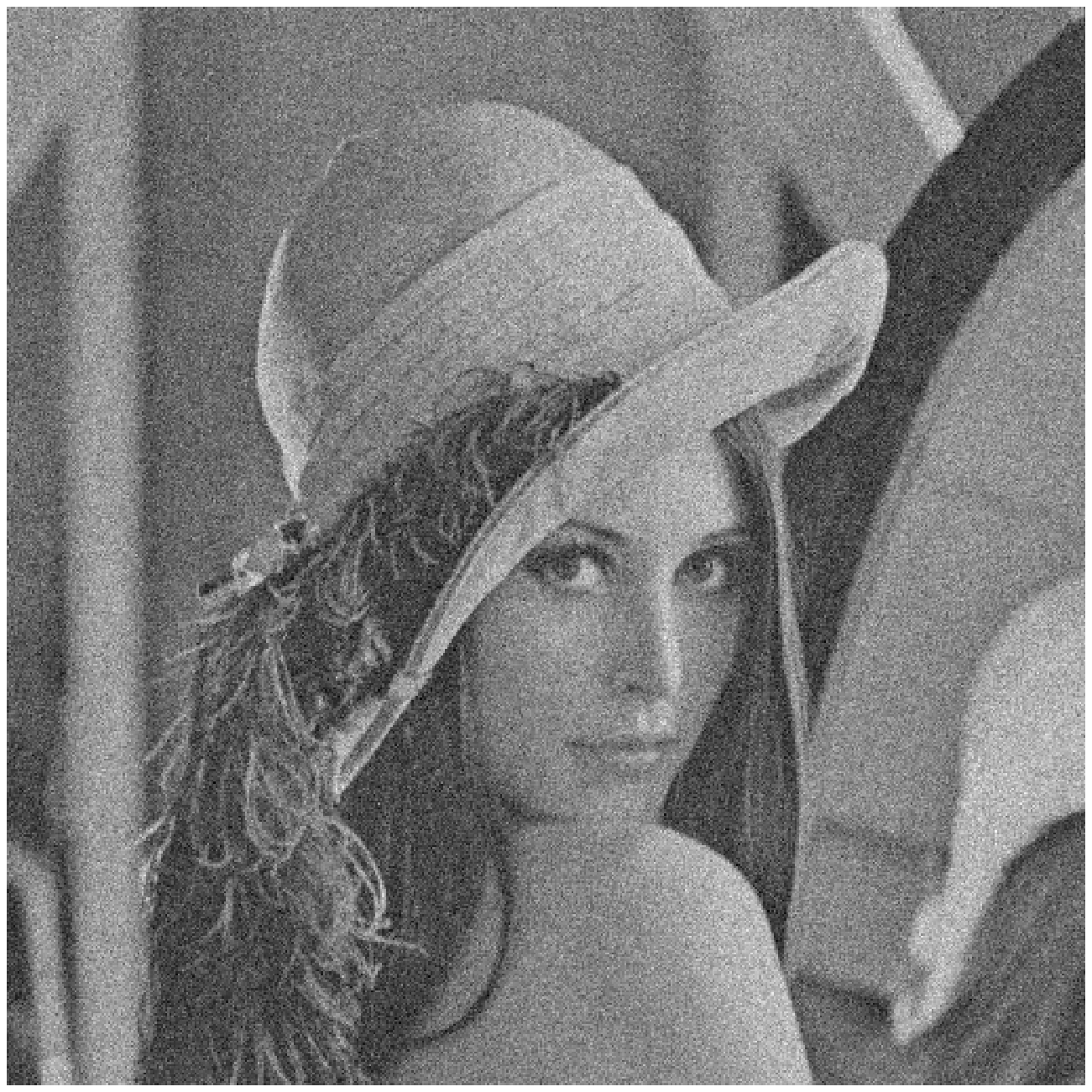}
& \includegraphics[scale=0.3]{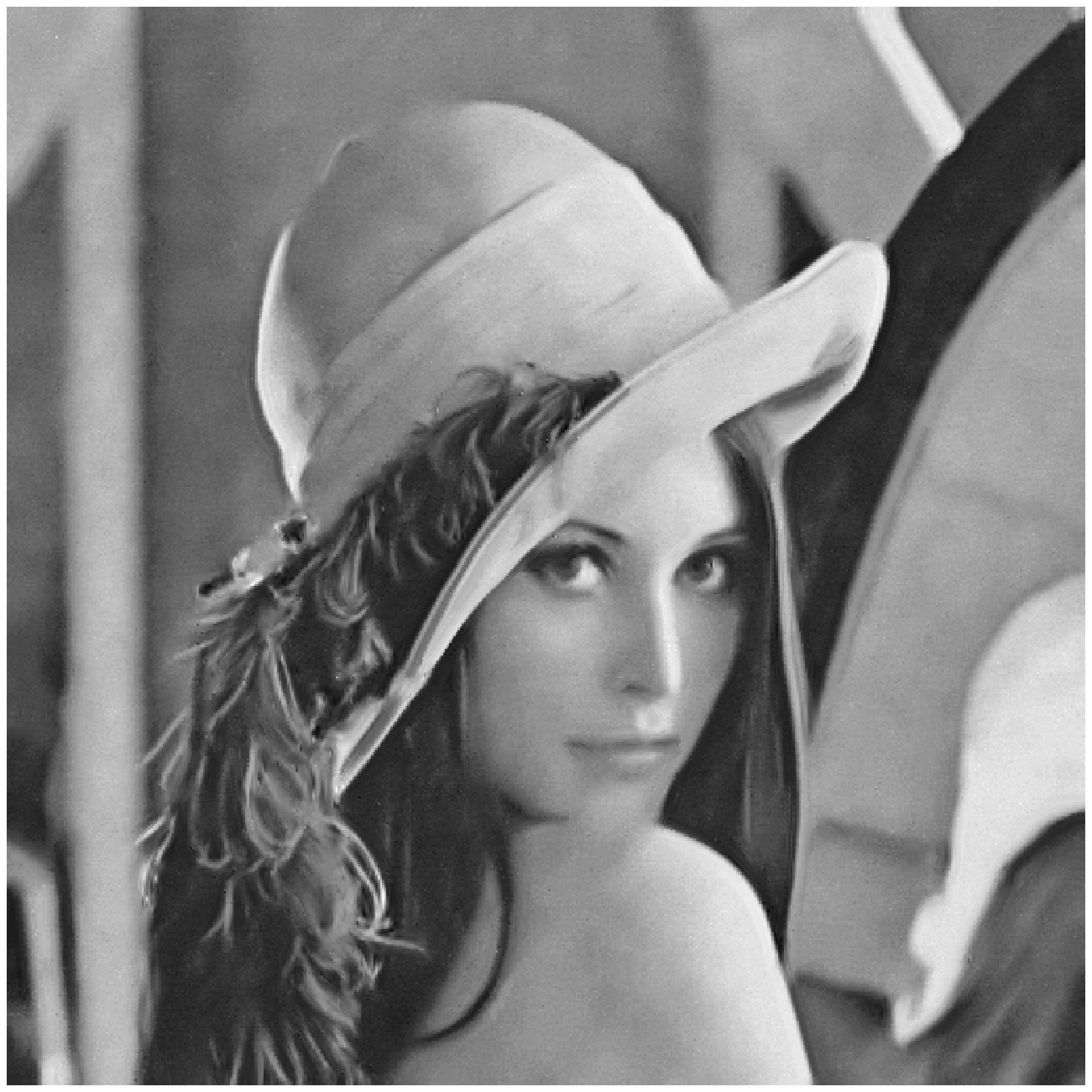} & \includegraphics[scale=0.3]{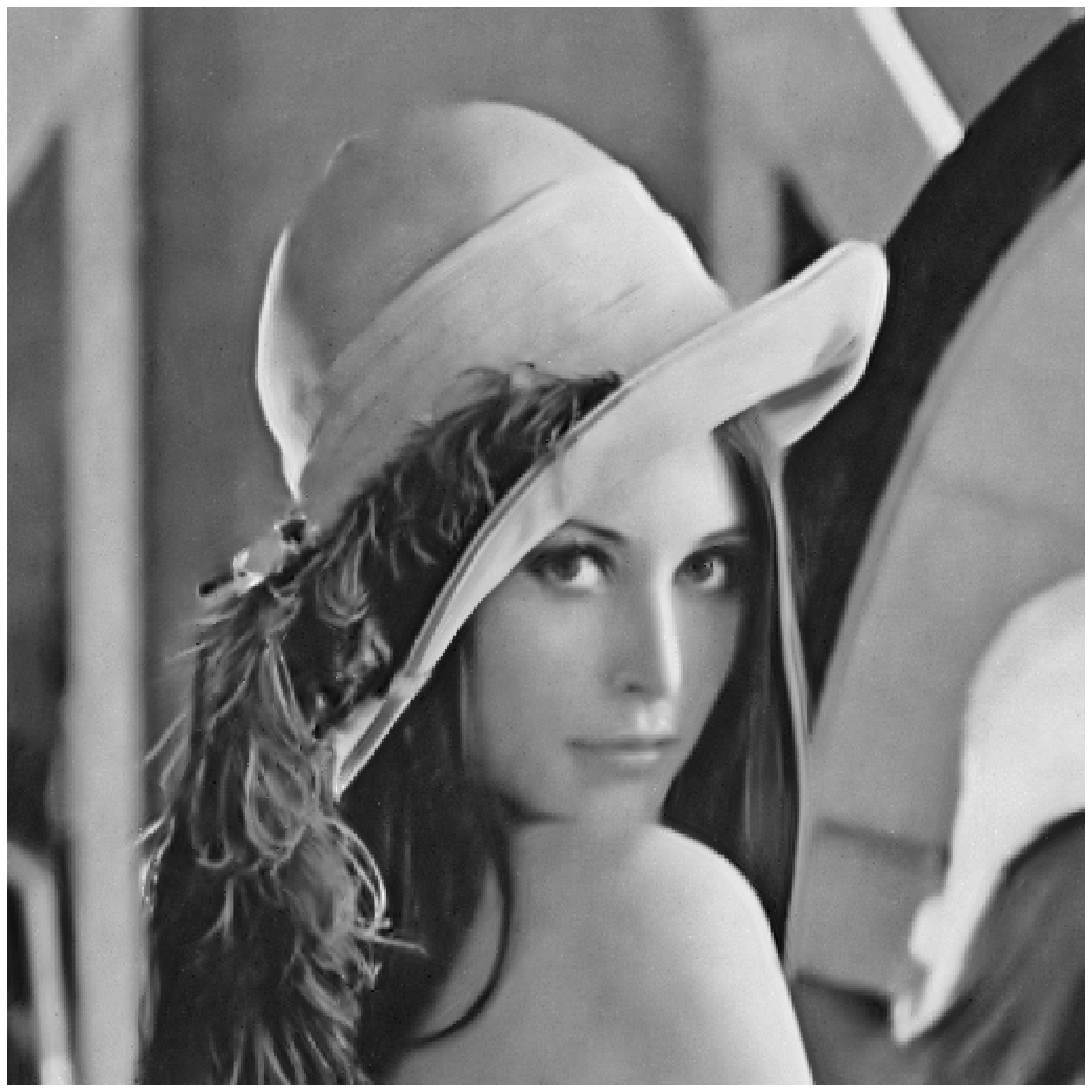} \\
{\small (a)} & {\small (b)} & {\small (c)} & {\small (d)}\\
\includegraphics[scale=0.3]{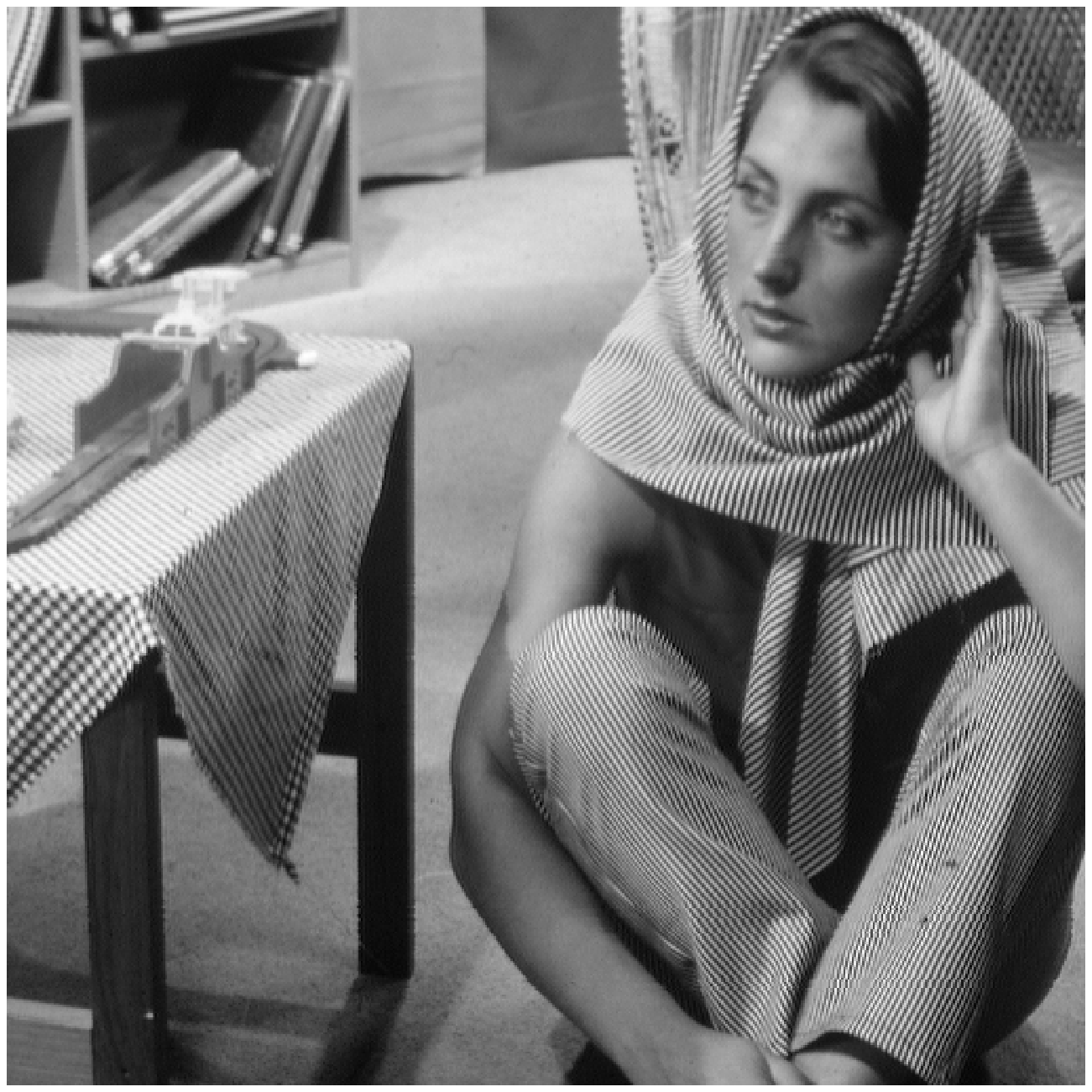} & \includegraphics[scale=0.3]{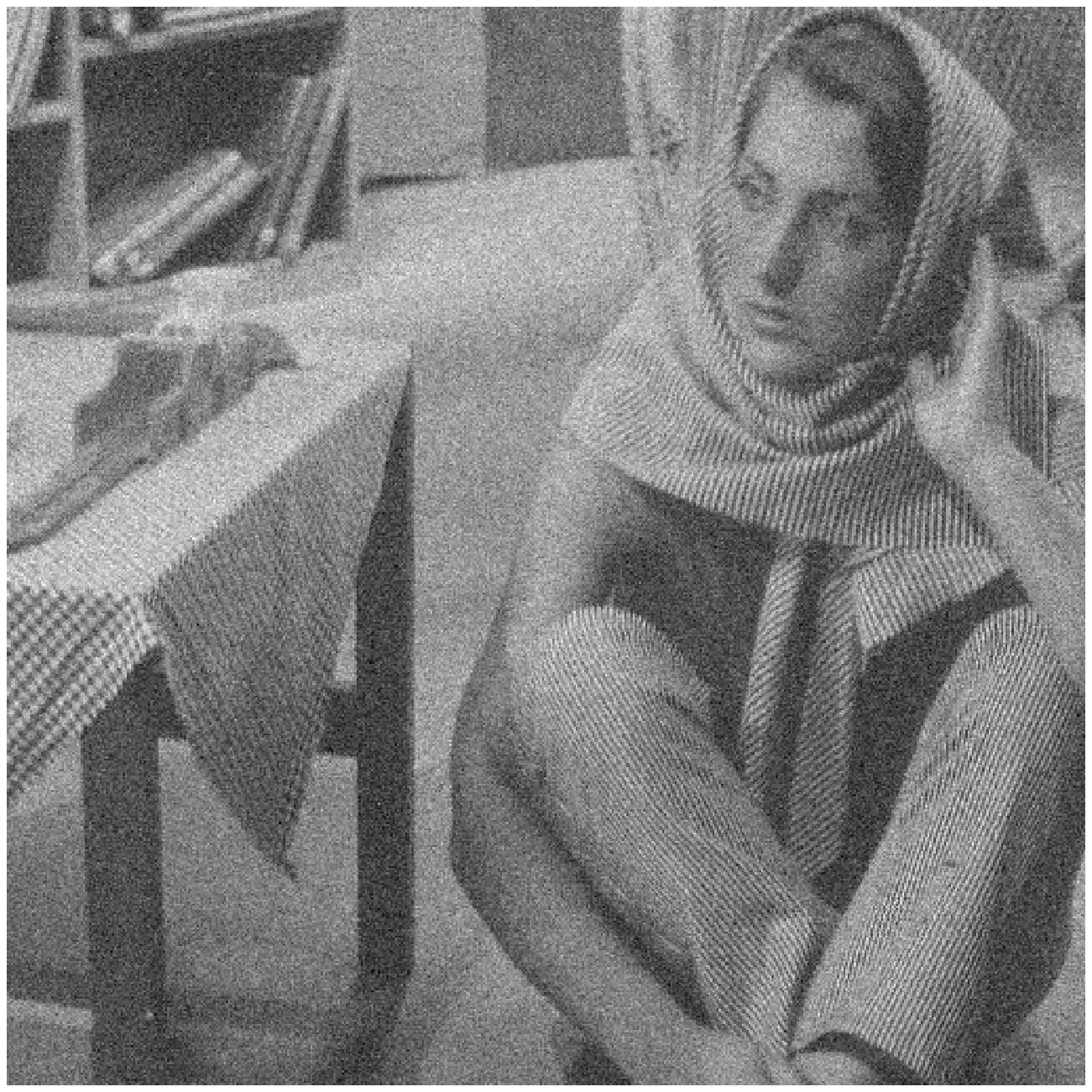}
& \includegraphics[scale=0.3]{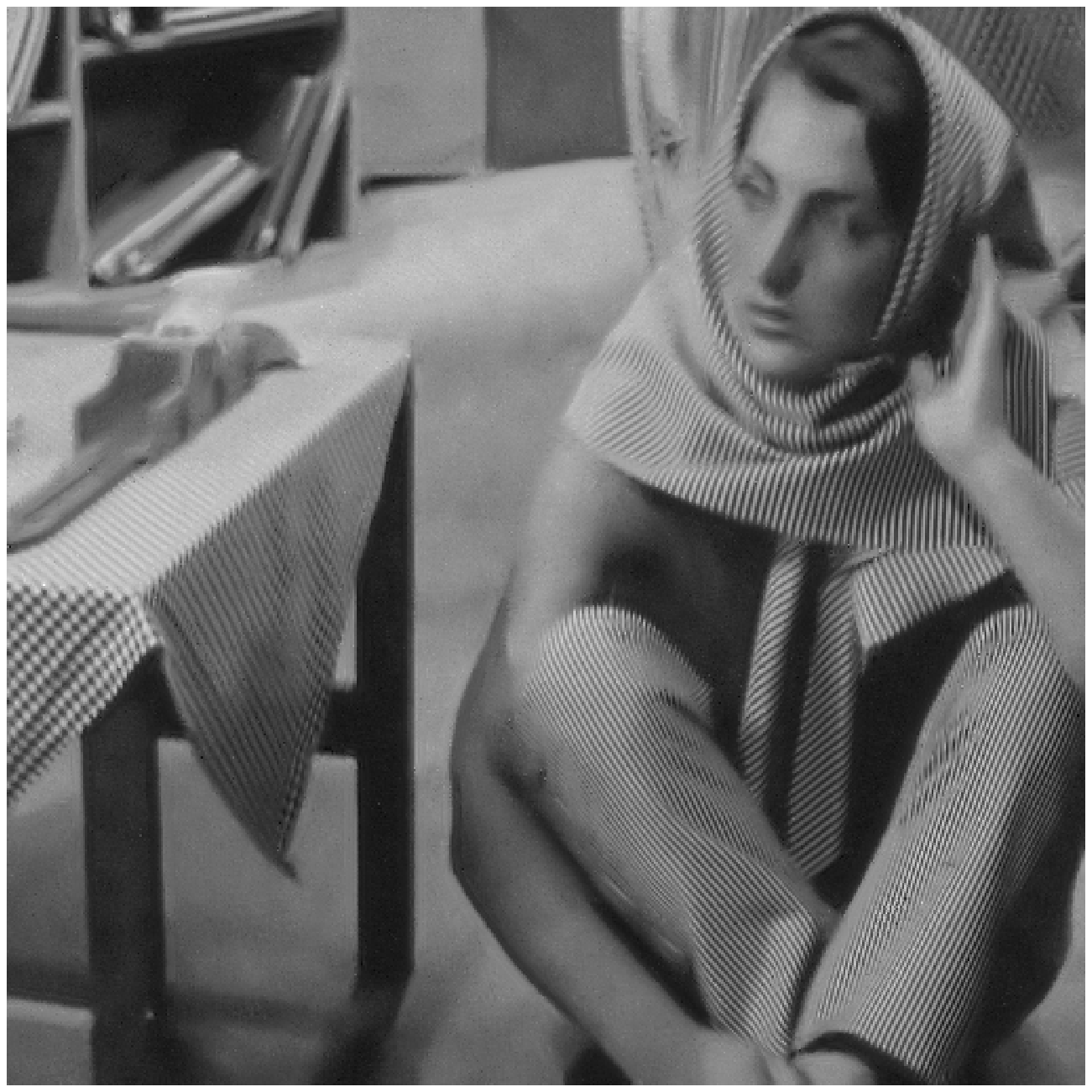} & \includegraphics[scale=0.3]{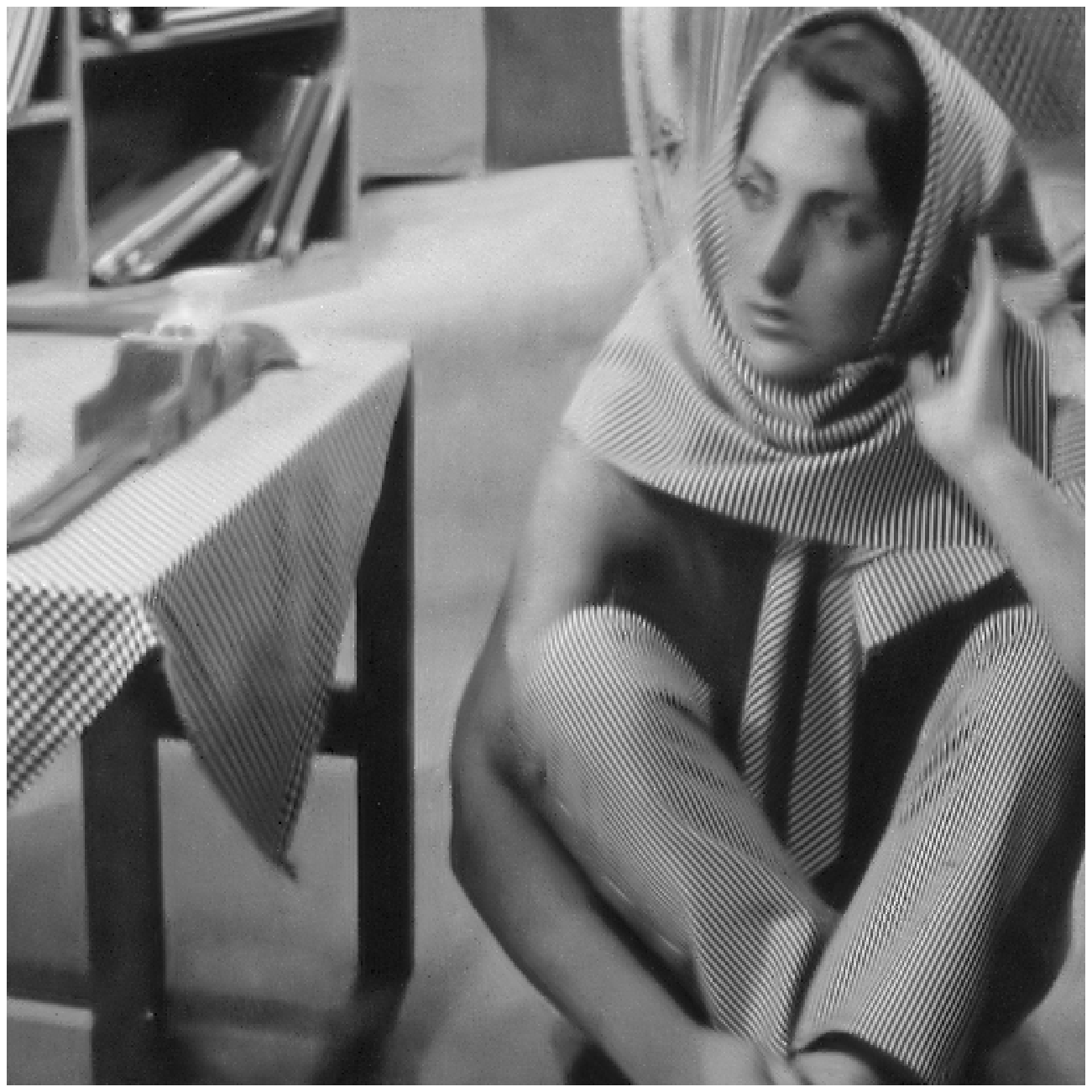} \\
{\small (e)} & {\small (f)} & {\small (g)} & {\small (h)}\\
\end{tabular}
\caption{Denoising results of noisy versions of the images Barbara and Lena ($\sigma=25$, input PSNR=20.17 dB) obtained with GTBWT with a Symmlet 8 filter with subimage averaging (SA) and without it:
(a) Original Lena. (b) Noisy Lena (20.17 dB). (c) Lena denoised  using GTBWT (30.3 dB). (d) Lena denoised using GTBWT with sub image averaging (31.21 dB)
(e) Original Barbara. (f) Noisy Barbara (20.17 dB). (g) Barbara denoised using GTBWT (28.94 dB). (h) Barbara denoised using GTBWT with sub image averaging (29.82 dB). }
\label{Figure: full Images}
\end{figure*}

We also applied the proposed denoising scheme to the noisy Lena and Barbara images.
The reconstructed images are shown in Figs. \ref{Figure: full Images}(d) and (h),
and Table \ref{Table: full_images_PSNR} shows the corresponding PSNR results.
It can be seen that the results obtained with the proposed algorithm for the Lena image are now closer to the ones received by the K-SVD algorithm,
and the results obtained for the Barbara image are better than the ones obtained with the K-SVD algorithm.

We note that only a subset of design parameters has been explored when we applied the proposed image denoising algorithm.
These design parameters include the patch size, the wavelet filter type (we have not explored biorthogonal wavelet filters at all),
and the number of trees averaged by the cycle spinning method.
They also include the number of nearest neighbors to choose from in the construction process of each level of these trees,
and the parameter $\epsilon$ used to determine the probabilities of choosing them.
We believe that better results may be achieved with the adequate design parameters,
and therefore more tests need to performed in order to search for the parameters which optimize the performance of the algorithm.
We also note that the use of the proposed sub-image averaging scheme is not restricted to the generalized tree-based wavelet transform.
This scheme may be used to improve the denoising results obtained with different methods which use
\textcolor{black}{distance} matrices similar to the one employed here.

\begin{table}[t]
\centering
\caption{Denoising results of a noisy version of the image in Fig. \ref{Figure: wavelets}(a) ($\sigma=25$, input PSNR=20.18 dB),
obtained using the GTBWT, with and without subimage averaging (SA), and with different wavelet filters.
Also shown is the average number of coefficients (\#coeffs.) used by each scheme to represent an image.\newline}

\begin{tabular}{|c|c|c|c|c|}\hline
& \multicolumn{2}{|c|}{GTBWT} & \multicolumn{2}{|c|}{with SA} \\\hline
wavelet type & PSNR [dB] & \#coeffs & PSNR [dB] & \#coeffs \\\hline
db1 (Haar) & 28.19  & 303.28 & 29.05  & 349.02 \\\hline
db4 & 28.5 & 307.62 & 29.37 & 341.25 \\\hline
db8 & 28.49 & 384.08 & 29.33 & 403.2 \\\hline
sym2 & 28.45 & 293.84 & 29.25 & 326.48 \\\hline
sym4 & 28.51 & 308.6 & 29.38 & 349.81 \\\hline
sym8 & 28.55 & 342.76 & 29.39 & 392.77 \\\hline
\end{tabular}

\label{Table: lena_face}
\end{table}

\begin{table}[t]
\centering
\caption{Denoising results (PSNR in dB) of noisy versions of the images Barbara and Lena ($\sigma=25$, input PSNR=20.17 dB) obtained with:
1) GTBWT with a Symmlet 8 filter with subimage averaging (SA) and without it. 2) with the K-SVD algorithm.\newline}
\begin{tabular}{|c|c|c|c|}\hline
Image & GTBWT & GTBWT + SA & K-SVD \\\hline
Lena & 30.3  & 31.21 & 31.32  \\\hline
Barbara & 28.94 & 29.82 & 29.62 \\\hline
\end{tabular}
\label{Table: full_images_PSNR}
\end{table}

\subsection{Robustness to noise}
\label{noise robustenss}
\textcolor{black}{We wish to explore the robustness of the generalized tree-based wavelet denoising scheme to the noise in the points $\mathbf{x}_j$.
More specifically we wish to check how using cleaner image patches will affect the denoising results.
Here we obtained "clean" patches from a noisy image by applying our denoising scheme once,
and using the patches from the reconstructed image in a second iteration of the denoising algorithm for defining the permutations.
For comparison reasons we also used clean patches from the original clean image in our denoising scheme,
and regarded the obtained results as oracle estimates.}

\textcolor{black}{We applied the GTBWT denoising schemes, obtained with patches from the clean and reconstructed images,
to the noisy Barbara and Lena images.
The results obtained with the Symmlet 8 filter and with subimage averaging are shown in Table \ref{Table: full_images_oracle}.
It can be seen that large improvements of about $1$ dB for the Lena image and $0.8$ dB for the Barbara image are obtained with clean patches.
Applying $2$ iterations of the denoising algorithm slightly improves the results of the Lena image by about $0.13$ dB.
However, applying $2$ iterations of the denoising algorithm to the Barbara image degrades the obtained results by about $0.11$ dB.
This degradation probably results from oversmoothing of the patches of the reconstructed image, which leads to a loss of details.
A choice of a different threshold in the denoising algorithm applied in first iteration,
or a different method to clean the patches altogether, may lead to improved denoising results.}

\begin{table}[t]
\centering
\caption{\textcolor{black}{Denoising results (PSNR in dB) of noisy versions of the images Barbara and Lena ($\sigma=25$, input PSNR=20.17 dB)
obtained using the GTBWT with subimage averaging.
The trees were constructed using patches obtained from the noisy (1 iter), original (clean) and reconstructed (2 iters) images.}\newline}
\begin{tabular}{|c|c|c|c|}\hline
Image & 1 iter & 2 iters & clean \\\hline
Lena & 31.21 & 31.34 & 32.25 \\\hline
Barbara & 29.62 & 29.71 & 30.61 \\\hline
\end{tabular}
\label{Table: full_images_oracle}
\end{table}

\section{conclusion}

We have proposed a new wavelet transform applicable to graphs and high dimensional data.
This transform is an extension of the multiscale harmonic analysis approach proposed by Gavish et al. \cite{gavish2010mwot}.
We have shown a relation between the transform suggested by Gavish et al. and \ac{1D} Haar wavelet filtering,
and extended the scheme so it will use general wavelet filters.
We demonstrated the ability of the generalized scheme to represent images more efficiently than the common \ac{1D} and separable \ac{2D} wavelet transforms.
We have also shown that our proposed scheme can be used for image denoising,
and that combined with a subimage averaging scheme it achieves denoising results which are close to the state-of-the-art.

In our future work plans, we intend to consider the following issues:
\begin{enumerate}

\item Seek ways to improve the method that reorders the approximation coefficients in each level of the tree, replacing the proposed nearest neighbor method.

\item Extend this work to redundant wavelets.

\item Improve the image denoising results by using two iterations with different threshold settings, and by considering spatial proximity as well in the tree construction.

\end{enumerate}

\section*{Acknowledgements}
The authors thank Ron Rubinstein for the fruitful discussions
and ideas which helped in developing the presented work.
\textcolor{black}{The authors also thank the anonymous reviewers for their helpful comments.}

\bibliographystyle{IEEEtran}
\bibliography{IEEEabrv,treeWave}

\end{document}